\documentclass[letterpaper]{article} 
\usepackage{aaai2026}  
\usepackage{times}  
\usepackage{helvet}  
\usepackage{courier}  
\usepackage[hyphens]{url}  
\usepackage{graphicx} 
\usepackage{natbib}  
\usepackage{caption} 
\usepackage{algorithm}
\usepackage{algorithmic}
\usepackage{booktabs}       
\usepackage{amsfonts}       
\usepackage{nicefrac}       
\usepackage{microtype}      
\usepackage{multirow} 
\usepackage[flushleft]{threeparttable}
\usepackage{amsmath}
\usepackage{amssymb}
\usepackage{mathtools}
\usepackage{bm}
\usepackage{amsthm}
\usepackage{xspace}
\usepackage{tikz}
\usepackage{tabularx}
\newcommand{\method}{FedP\textsuperscript{2}EFT\xspace}
\newcommand{\seen}{{\em seen}\xspace}
\newcommand{\unseen}{{\em unseen}\xspace}
\newcommand{\basemodel}{{\em base model}\xspace}
\newcommand*\circled[1]{\tikz[baseline=(char.base)]{
            \node[shape=circle,fill,inner sep=1.0pt,scale=0.8] (char) {\textcolor{white}{#1}};}}
\DeclareMathOperator*{\argmin}{arg\,min}

\usepackage{newfloat}
\usepackage{listings}
\DeclareCaptionStyle{ruled}{labelfont=normalfont,labelsep=colon,strut=off} 
\floatstyle{ruled}
\newfloat{listing}{tb}{lst}{}
\floatname{listing}{Listing}
\title{FedP$^2$EFT: Federated Learning to Personalize PEFT for Multilingual LLMs}

\author{
    Royson Lee\textsuperscript{\rm 1},
    Minyoung Kim\textsuperscript{\rm 1},
    Fady Rezk\textsuperscript{\rm 2},
    Rui Li\textsuperscript{\rm 1},
    Stylianos I. Venieris\textsuperscript{\rm 1},
    Timothy Hospedales\textsuperscript{\rm 1,2}
}
\affiliations{
    \textsuperscript{\rm 1}Samsung AI Center, Cambridge, UK\\
    \textsuperscript{\rm 2}University of Edinburgh, UK\\
    royson.lee@samsung.com
}

\begin{document}

\maketitle

\begin{abstract}
Federated learning (FL) has enabled training of multilingual large language models (LLMs) on diverse and decentralized multilingual data, especially on low-resource languages. To improve client-specific performance, personalization via the use of parameter-efficient fine-tuning (PEFT) modules such as LoRA is common. 
This involves a {\em personalization strategy} (PS), such as the design of the PEFT adapter structures (\textit{e.g.}, in which layers to add LoRAs and what ranks) and choice of hyperparameters (\textit{e.g.}, learning rates) for fine-tuning. Instead of manual PS configuration, we propose \method{}, a federated {\em learning-to-personalize} method for multilingual LLMs in cross-device FL settings. Unlike most existing PEFT structure selection methods, which are prone to overfitting low-data regimes, \method{} collaboratively learns the optimal personalized PEFT structure for each client via Bayesian sparse rank selection. Evaluations on both simulated and real-world multilingual FL benchmarks demonstrate that \method{} largely outperforms existing personalized fine-tuning methods, while complementing other existing FL methods. \end{abstract}

\begin{links}
    \link{Code}{https://github.com/SamsungLabs/fedp2eft}
\end{links}

\section{Introduction}
\label{sec:intro}

Federated learning (FL) makes it possible to train multilingual large language models (LLMs) across different geographical regions, protecting linguistic diversity for low-resource languages~\cite{zhao2023breaking} while being compliant with privacy regulations~\cite{lim2020federated}, \textit{e.g.},~General Data Protection Regulation (GDPR). Despite the impressive capabilities demonstrated by these models across various languages, their performance varies significantly depending on the language~\cite{rust2021good} and the data volume~\cite{adelani2021masakhaner} per client. 

Moreover, the majority of existing FL-based multilingual LLM approaches have thus far focused on training a single global model~\cite{DEPT, fedllm-bench}, limiting their performance on specific languages. Concretely, scaling a single model to different languages is challenged by issues such as the \textit{Curse of Multilinguality}~\cite{conneau-etal-2020-unsupervised}, where adding more languages often leads to diminishing returns, and \textit{Negative Interference}~\cite{wang2020negative}, where diverse languages compete for limited model capacity. From a personalized FL (pFL) perspective, learning a global model often increases the initial performance at the expense of personalized performance, \textit{e.g.},~fine-tuning from the global model~\cite{jiang2019improving}.

Naturally, pFL approaches can help bridge the gap and improve language personalization. However, existing techniques are either too costly to be applied to LLMs, \textit{e.g.},~the use of meta-learning~\cite{fedmeta} and  hypernetworks~\cite{pfedhn}, or rely on suboptimal hand-crafted personalization strategies, \textit{e.g.},~manual choice of personalized and language-specific layers~\cite{DEPT,zhao2023breaking,fedlora}, parameter-efficient fine-tuning (PEFT) adapter structures~\cite{SA-FedLora, FedDPA}, or clustering based on class labels and/or language commonality~\cite{hypcluster, fedllm-bench}.

Intuitively, optimizing personalization in pFL often necessitates dataset- and task-specific methods. The optimal level of personalization varies significantly depending on the characteristics of the data and the specific FL scenario~\citep{chen2022pfl,fedllm-bench}. For instance, an English-pretrained LLM may require stronger personalization, \textit{e.g.},~higher learning rates, when fine-tuning on German than on English. The optimal personalization strategy (PS) is thus contingent upon the specific task, the client, and the given \basemodel{}~\cite{royson2023fedl2p}. 

\begin{figure}[t]
    \centering
    
    \includegraphics[width=1.0\columnwidth,trim={5cm 6cm 3cm 1cm},clip]{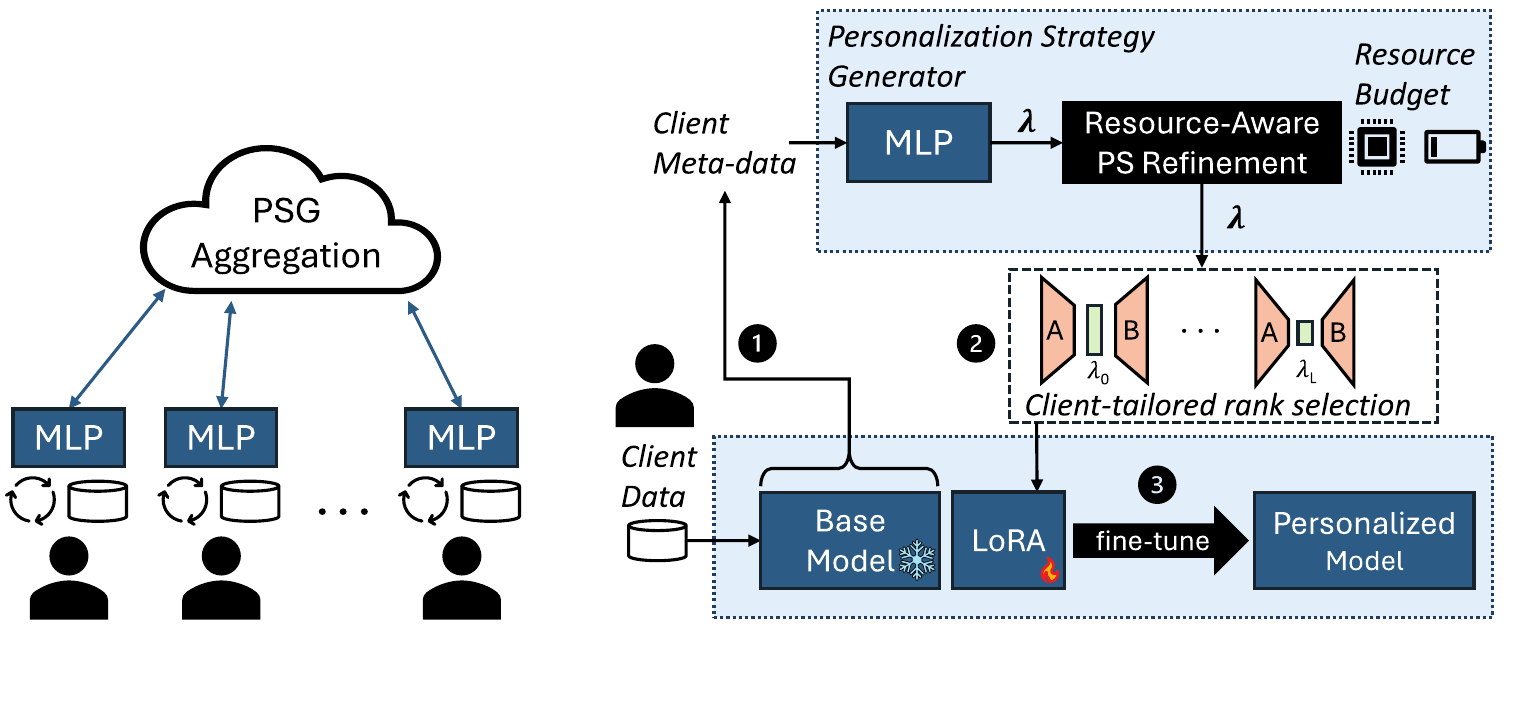} 
    \put (-200,2) {(a)}
    \put (-75,2) {(b)}

    \caption{(a) We train our personalization strategy generator (PSG) using standard FL approaches. (b) \method{}'s inference stage on a single client. 
    \protect\circled{1}~Given the \basemodel{} and the client's train dataset, features are extracted and passed into our PSG to generate a PS, $\bm{\lambda}$, for the client's budget. \protect\circled{2}~$\bm{\lambda}$ is then used to initialize all LoRA modules before \protect\circled{3}~the \basemodel{} is personalized. The resulting personalized model is then used to evaluate on the client's test samples. 
    }
    \label{fig:intro}
\end{figure}

In this paper, we address the issues above by proposing a novel federated hyperoptimization strategy that learns personalized PEFT configurations for each client. To this end, we propose \method{}, a method that enables clients to collaboratively learn language personalization strategies using FL. Specifically, we federatedly train a PS generator (PSG), as depicted in Fig.~\ref{fig:intro}(a), which allows all participating clients to collaboratively learn the optimal mapping between local meta-data to optimal LoRA~\cite{hu2021lora} ranks. Fig.~\ref{fig:intro}(b) illustrates the personalization process of \method{} on a single client during inference, where per-layer LoRA ranks are generated depending on the initial \basemodel{}, the client's dataset, and their resource budget. These personalized LoRA modules are then used to apply PEFT on the \basemodel{} and yield the personalized model. 

As \method{} focuses on improving the PEFT process per-dataset/task/client, it is directly pluggable to any starting \basemodel{}, which may or may not be federatedly trained. This includes off-the-shelf pretrained models, FL approaches that learn a single global model~\cite{fedavg,fedbabu}, and even pFL approaches that deploy personalized layers, \textit{e.g.},~monolingual tokenizer and embeddings~\cite{DEPT}, personalized LoRA adapters~\cite{fedlora, FedDPA}, and language embeddings~\cite{FedPerC}. Through our experiments (Section~\ref{sec:expmts}), we show that our method \textit{1)}~largely outperforms both existing non-FL LoRA rank selection and FL-based learning-to-personalize techniques, and \textit{2)}~complements well with a range of existing FL approaches.

\section{Related Work}\label{sec:related}

\noindent\textbf{Multilingual LLMs (MLLMs).}~Existing efforts in multilingual LLMs often underperform on low-resource languages due to \textit{1)}~data scarcity~\cite{xu2024survey}, \textit{2)}~the model's limited capacity to learn the intricacies of multiple languages~\cite{conneau-etal-2020-unsupervised}, and \textit{3)}~negative transfer learning among languages~\cite{wang2020negative}.
Common ways to counteract these challenges include the use of separate vocabulary and embeddings~\cite{artetxe2020cross}, hand-crafted adapters~\cite{pfeiffer2020mad}, automatic data annotation~\cite{llama3}, clustering and merging languages with similar representations~\cite{chung2020improving}
, among other contributions~\cite{wang2020extending,Conneau2019UnsupervisedCR}. 
Our work is orthogonal to these approaches and builds upon recent FL-based MLLMs~\cite{DEPT,fedllm-bench}, which utilize FL to tap into previously inaccessible low-resource data sources.

\noindent\textbf{Personalized Federated Learning (pFL).}~To obtain personalized client-specific models, various approaches have been proposed, including the use of personalized layers~\cite{fedper}, meta-learning~\cite{fedmeta}, model mixtures~\cite{fedem}, hypernetworks~\cite{pfedhn}, and transfer learning between global and local models~\cite{fml}. Some of these techniques have also been adopted for LLMs, \textit{e.g.},~personalized LoRAs~\cite{FedDPA}, hypernetworks for client embeddings~\cite{FedPerC}, and mixtures of LoRA experts~\cite{FedAMoLE}. Our work complements these approaches as personalized models can benefit from further fine-tuning as shown in Section~\ref{sec:text_class}.

\noindent\textbf{Federated Hyperparameter Optimization (HPO).}~Most federated approaches to HPO do not utilize the client dataset for personalized hyperparameters. Instead, they employ a single set of hyperparameters across all clients based on the local validation loss evaluated before FL~\cite{flora} or sample from federatedly learnt hyperparameter categorical distributions for each client~\cite{fedex}. An exception to this is FedL2P~\cite{royson2023fedl2p} which utilizes a PSG for personalized per-layer learning rates and batch normalization hyperparameters. While FedL2P has been shown to work well in standard small-scale image and speech benchmarks, their applicability to LLMs is unclear: i) LLMs don't use batch normalization (BN), ii) LLMs often use adaptive optimizers, making learning rate a less sensitive hyparam for downstream performance. Learning the learning rates also require FedL2P to adopt expensive 2nd-order optimization methods as the learning rate is not a direct gradient of the loss, a limitation that is further aggravated with LLMs. We compare with FedL2P in our experiments.

\noindent\textbf{PEFT Structure Learning.}~Contrary to the conventional approach of distributing uniform adapter modules across all layers, a recent line of work allows different LoRA ranks to be used across a model's weight matrices. Using fine-grained per-layer rank selection, existing methods include SVD-based LoRA reformulation followed by importance-based rank assignment~\cite{adalora}, trainable rank-gating units~\cite{sora2023emnlp}, selectively employing parallel weight modules~\cite{capaboost2024iclr}, 
meta-learning-based~\cite{autolora} and black-box optimization techniques~\cite{hpolorallm2024aaaiw}, 
specialized training recipes for multi-rank LoRA modules that allow flexible extraction of a range of ranks~\cite{dylora2023eacl}, and coarse-grained dropping of LoRA-enhanced layers~\cite{yao2024layer}. While these methods can be effective in centralized setups, they typically require an excessive number of optimization steps, which is prone to overfitting in FL settings, where clients have limited amount of data.

\section{Our Approach}\label{sec:main}

\subsection{Preliminaries \& Motivation}

In pFL, the goal is to minimize each client's local objective $\mathbb{E}_{(x,y) \sim P^i}\mathcal{L}^i(\Phi^i;x,y)$ where $P^i$ represents the data distribution of the $i$-th client, $x$ and $y$ are the input data and labels, respectively, and $\mathcal{L}^i(\Phi^i;x,y)$ is the loss function for client $i$ given model parameters $\Phi^i$. This is typically achieved via fine-tuning~\cite{chen2022pfl} a \basemodel{}, with parameters $\Phi^i_{BM}$ and a set of hyperparameters, \textit{e.g.}~learning rate. Note that $\Phi^i_{BM}$ may differ across clients if it is already personalized, \textit{e.g.}~if $\Phi^i_{BM}$ is obtained using a pFL algorithm.

Fine-tuning LLMs, however, is unprecedentedly compute and memory intensive, and prone to overfitting. As such, the majority of existing federated LLM works~\cite{zhao2023breaking,fedpeft} rely on PEFT methods, with LoRA~\cite{hu2021lora} being a prevalent choice due to its efficiency and performance. Specifically, for a frozen weight matrix $W \in \mathbb{R}^{d\times e}$, LoRA introduces low-rank matrices $B \in \mathbb{R}^{d \times r}$ and $A \in \mathbb{R}^{r \times e}$ where $r \ll \min(d, e)$. The adapted weights are then expressed as: $W' = W + \frac{\alpha_{\text{lora}}}{r}BA$ where $\alpha_{\text{lora}}$ is a hyperparameter and only $B$ and $A$ are trained during fine-tuning.
Although effective, these FL works rely on a fixed hand-crafted PS, \textit{e.g.},~a manually defined LoRA rank on hand-picked layers, for all clients, leading to suboptimal personalized models. 

\subsection{Personalized PEFT}\label{sec:personalized_peft}

We, instead, propose using a different PS for each client. Common hyperparameter choices from previous federated HPO approaches (Section~\ref{sec:related}) include learning rates and BN hyperparameters. While these hyperparameters have been shown to be effective for handling data heterogeneity in popular vision and speech benchmarks~\cite{fedbn,li2016revisiting,fedper}, they are less consequential or not applicable when fine-tuning LLMs. This stems from the fact that LLMs are often fine-tuned using adaptive optimizers, \textit{e.g.}~Adam, which are more robust to the learning rate~\cite{zhao2025deconstructing}, and BN layers are not typically used. 
A more critical hyperparameter choice shown to be effective, especially for cross-lingual transfer learning~\cite{pfeiffer2020mad}, is the PEFT adapter structure; specifically which layers to introduce LoRAs in and what ranks to utilize~\cite{adalora,autolora}. 

\noindent\textbf{Adapting BayesTune for LoRA Rank Selection.}
We build upon BayesTune~\cite{kim2023bayestune}, a Bayesian sparse model selection approach. Directly using BayesTune in our approach, however, will not only incur an intractable cost but also be difficult to optimize due to the high dimensionality of predicting per-parameter scaling coefficients (Appendix~\ref{app:bayestune}). Hence, we adapt BayesTune by formulating PEFT personalization as a sparse LoRA rank selection problem and propose BayesTune-LoRA (BT-LoRA). 
Concretely, we introduce rank-wise latent variables $\lambda \in \mathbb{R}^r, \; \lambda_i > 0, \; \forall i=1,2,\cdots,r$ for each LoRA matrix: $B\lambda A$. Let $\bm{\lambda}= \{\lambda_{l,\cdot}\}_{l=1}^L$ be the set of all $\lambda$ where $\lambda_{l,\cdot}$ represents the rank-wise scales for layer $l$ in a model with $L$ LoRA modules (similarly for $\bm{A}$ and $\bm{B}$). Using BayesTune, the 
values for $\theta=(\bm{\lambda}$,$\bm{A}$,$\bm{B})$ are optimized as:
%
\begin{align}
\label{eq:bayestune}
\theta^* =& \argmin_{\theta} \mathcal{L}_{\text{CE}}(\theta;D) + \frac{\alpha_{s}}{N} \mathcal{L}_{s}(\bm{\lambda},\bm{B}) + \frac{\alpha_{p}}{N}\mathcal{L}_{p}(\bm{\lambda}) 
\end{align}
where $D$$=$$\{(x_i, y_i)\}_{i=1}^N$ is the train dataset, $N$ the size of $D$, $\mathcal{L}_{\text{CE}}(\theta;D)$ the cross-entropy loss, $\alpha_{p}$ and $\alpha_{s}$ hyperparameters, $\mathcal{L}_s$ the logarithm of the Laplace distribution (prior imposed on $p(B|\lambda)$\footnote{Unlike BayesTune, where every parameter is associated with its own prior scale, we use an ``independent'' Laplace prior where each $\lambda_{l,i}$ applies to all entries of $B_{l,i}$}), $f(\|B_{l,i}\|_1;\mu,b)= \frac{1}{2b} \exp\left(-\frac{|\|B_{l,i}\|_1 - \mu|}{b}\right)
$ with $\mu=0$ ($B$ is initialized to 0 in LoRA) and $b=\lambda_{l,i}$:
\begin{align}
\mathcal{L}_s(\bm{\lambda}, \bm{B}) = \sum^L_l \sum^r_i \left(\log \lambda_{l,i} + \frac{\|B_{l,i}\|_1}{\lambda_{l,i}} + \log2\right)
\end{align}
and $\mathcal{L}_p$ is the logarithm of the Gamma distribution (hyper-prior imposed on $\lambda$), $\mathcal{G}(\lambda_{l,i};\alpha_g,\beta_g)= \frac{\beta_g^{\alpha_g}}{\Gamma(\alpha_g)} \lambda_{l,i}^{\alpha_g-1} e^{-\beta_g \lambda_{l,i}}$ where $\alpha_g=0.01,\beta_g=100$ following the hyperparameters set by the original authors\footnote{$\alpha_g$ and $\beta_g$ do not need to be tuned. Details in  Appendix~\ref{app:bayestune}.}:
\begin{align}
\mathcal{L}_p(\bm{\lambda}) = \sum^L_l \sum^r_i &(0.99\cdot \log \lambda_{l,i} + 100 \cdot \lambda_{l,i} \nonumber \\
&- 0.01\log(100) + \log\Gamma(0.01)) 
\end{align}
In practice, we can save computations by removing all constants and the duplicate term $\log \lambda$, resulting in the following approximated penalty losses:
\begin{align}
\mathcal{L}_s(\bm{\lambda}, \bm{B}) &= \sum^L_l \sum^r_i \frac{\|B_{l,i}\|_1}{\lambda_{l,i}} \\\mathcal{L}_p(\bm{\lambda}) &= \sum^L_l \sum^r_i (\log \lambda_{l,i} + 100 \cdot \lambda_{l,i}) 
\end{align}
\noindent\textbf{Intuition of $\mathcal{L}_p$ and $\mathcal{L}_s$.} $\mathcal{L}_p$ encourages small $\lambda$ while $\mathcal{L}_s$ encourages larger $\lambda$ for larger LoRA $B$ (per column) updates. Hence, minimizing the losses in Eq.~(\ref{eq:bayestune}) encourages larger $\lambda$ in more significant ranks. 

\noindent\textbf{Personalizing PEFT with BT-LoRA.}~For each client, we attach BT-LoRA modules, $\theta$, to all linear layers of its \basemodel{} with rank $r_{\text{init}} = \alpha_{r\_mul} \cdot r_{\text{max target}}$ where $r_{\text{max target}}$ is the maximum inference resource budget and $r_{\text{init}}$ is the initial rank before pruning. $\theta$ is then optimized using Adam~\cite{Kingma_2014} as per Eq.~(\ref{eq:bayestune}).\footnote{BayesTune proposed using SGLD~\cite{welling2011bayesian}, adding Gaussian noise to the gradient updates and sampling from the posterior distribution. Due to the challenges of estimating the full posterior distribution in FL settings, particularly with limited client data, we opt to find a point estimate.}

After training, we freeze the resulting $\bm{\lambda}$ and use it for personalization. Specifically, given a resource budget (total rank budget) of $r \cdot L$, we prune $\bm{\lambda}$ by taking the top-$(r \cdot L)$ largest ranks, along with the corresponding rows of $\bm{A}$ and columns of $\bm{B}$.\footnote{The LoRA module is discarded for layers where $\|\lambda_l\|_1 = 0$} We then reinitialize the pruned $\bm{A}$ and $\bm{B}$ and perform standard fine-tuning on $\mathcal{L}_{\text{CE}}$ with the frozen pruned $\bm{\lambda}$ to obtain the personalized model. Note that we only have to train $\bm{\lambda}$ once for all ranks $\leq r_{\text{max target}}$.

\subsection{\method{}: FL to Personalize PEFT}\label{sec:main_method}

Overfitting often occurs when training an effective per-client PS in isolation due to limited client data. Following FedL2P~\cite{royson2023fedl2p}, we mitigate this by federatedly learning a common PSG that generates client-wise PS. Concretely, we use a one hidden layer multilayer perceptron (MLP) with parameters $\phi$ that takes as input the client meta-data and outputs an estimated PS:
%
\begin{align}
\label{eq:mlp}
\bm{\hat{\lambda}} = \text{MLP}(\phi; \:\: &E(h_0),SD(h_0), E(h_1),SD(h_1), \nonumber\\ &\cdots,E(h_{L-1}),SD(h_{L-1}))
\end{align}
where $h_{l-1}$ is the input feature to the $l$-th layer in the \basemodel{}, and $E(\cdot)$ and $SD(\cdot)$ are the mean and standard deviation (SD), respectively.

In contrast to FedL2P, which adopts a computationally demanding meta-learning approach to train MLP, we take a two-stage strategy for each client: \textit{1)} first, learn $\bm{\lambda}$, followed by \textit{2)} regression learning of MLP to target the learned $\bm{\lambda}$. 

\begin{figure}
    \centering
    \includegraphics[width=0.95\linewidth,trim={4.5cm 2.2cm 12cm 3.8cm},clip]{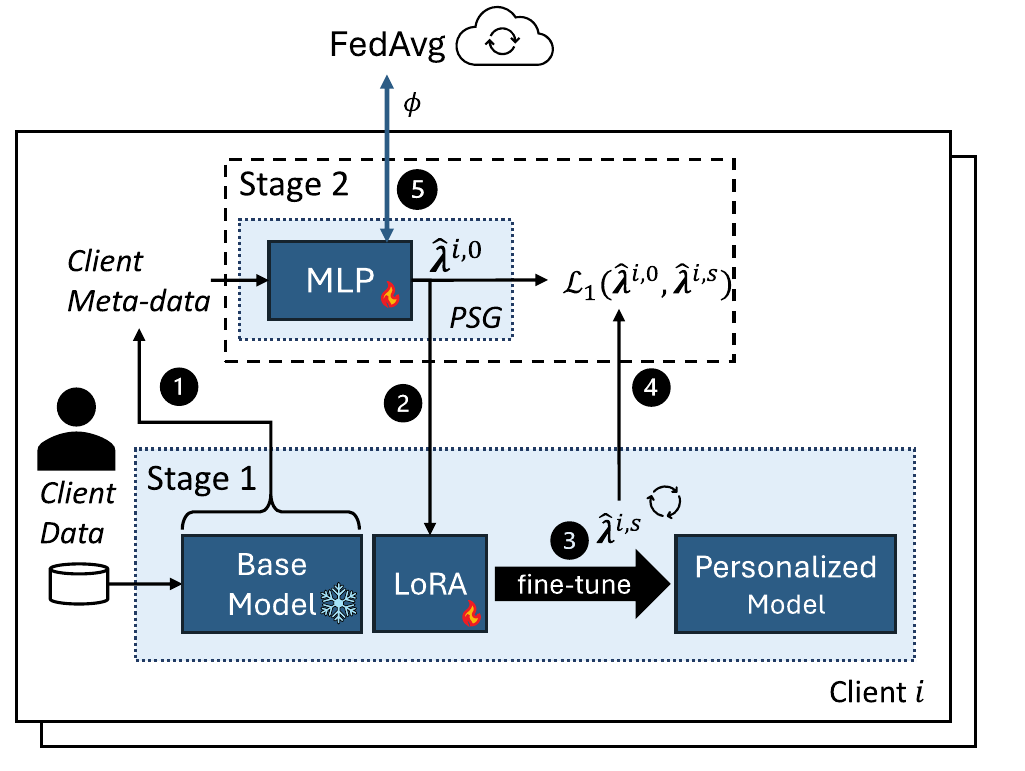} 
    \caption{\method{}'s federated training of PSG for each federated round. Details in Section.~\ref{sec:main_method}.}
    \label{fig:approach}
\end{figure}

\noindent\textbf{Federated Training of \method{}.}~Fig.~\ref{fig:approach} shows the entire \method{} algorithm during federated training. For each federated round, each sampled participating client $i$ receives $\phi$ from the server and loads them into its MLP. They then \protect\circled{1} perform a forward pass of the local train dataset on their \basemodel{} and a forward pass of the MLP with the resulting features as per Eq.~(\ref{eq:mlp}). \protect\circled{2} The estimated $\bm{\hat{\lambda}}^i$ is plugged into our proposed BT-LoRA (Section~\ref{sec:personalized_peft}) and \protect\circled{3} fine-tuning is performed as per Eq.~(\ref{eq:bayestune}) for $s$ steps (Stage 1). \protect\circled{4} The resulting $\hat{\bm{\lambda}}^{i,s}$
is used as an approximated ground-truth for regression learning of MLP to target the learned $\hat{\bm{\lambda}}^{i,s}$, where $\mathcal{L}_1$ is the L1 loss (Stage 2). Finally, \protect\circled{5} $\phi$ is sent back to the server for aggregation. As there is no single aggregation method that outperforms all others in every situation~\cite{matsuda2022empirical,chen2022pfl,fedllm-bench}, we utilize FedAvg~\cite{fedavg}. The aggregated $\phi$ is then sent to clients for the next round. To complement Figure~\ref{fig:approach}, we provide Algorithm~\ref{algo:train} in Appendix~\ref{app:algo}.

The learned $\phi$ is deployable to any client (\seen{} or \unseen{}) after federated training. Note that unlike FedL2P, which requires federated training for every target rank, \method{} inherits the property of BT-LoRA; federated training is a one-time cost for all ranks $\leq r_{\text{max target}}$.

\noindent\textbf{Inference with \method{}.}~Fig.~\ref{fig:intro}(b) shows how \method{} personalizes PEFT for each client upon deployment. Given the learned MLP and the client's \basemodel{}, \protect\circled{1} the client meta-data are retrieved (Eq.~\ref{eq:mlp}) and used to generate the client's PS, $\bm{\lambda}$. \protect\circled{2} Given the client's resource budget of total rank $r \cdot L$, we take the top-$(r \cdot L)$ largest ranks in $\bm{\lambda}$, freeze them, and initialize our proposed BT-LoRA modules for all layers where $\|\lambda_l\|_1$$>$$0$. \protect\circled{3} The personalized LoRA ranks are used for fine-tuning before merging back to the \basemodel{} to obtain the final personalized model. To complement Figure~\ref{fig:intro}(b), we provide Algorithm~\ref{algo:inference} in Appendix~\ref{app:algo}.

\section{Evaluation}\label{sec:expmts}

We conduct experiments on multilingual scenarios, where clients with diverse high- and low-resource languages can collaboratively learn how to personalize a given base model to better cater to their language preferences. In all experiments, we divide clients in two pools, \seen{} and \unseen{}, where only the clients in the \seen{} pool actively participate in federated training. We set the maximum number of communication rounds for training the PSG to $150$, randomly sampling $10\%$ of participating clients every round. We use Adam as the default optimizer for all our experiments. Unless mentioned otherwise, we show results for resource budget $r=2$ where the total rank budget is $r \cdot L$ and leave results for other resource budgets $r=4,8,16$ in the Appendix. 
In the following subsections, we summarize the FL scenarios that we consider in our experiments, leaving comprehensive details in Appendix~\ref{appendix:experiments}.

\subsection{Setup: Tasks, Models, and Datasets}

\noindent\textbf{Text Classification.}~We adopt the pretrained multilingual BERT~\cite{BERT} (mBERT) for all text classification experiments. For datasets, we introduce additional data heterogeneity to the simulated FL setups, XNLI~\cite{XNLI} and MasakhaNEWS~\cite{MasakhaNEWS}, proposed in PE\_FL~\cite{zhao2023breaking}. 

For our XNLI setup, we sample 2k instances for train and $500$ for test in each pool. In contrast to PE\_FL, which had $15$ clients ($1$ client per language), we divide the data equally among $20$ clients per language. We then adopt the latent Dirichlet allocation (LDA) partition method~\cite{hsu2019measuring}, $y \sim Dir(\alpha)$, to simulate non-IID label shifts among these clients, with $\alpha=0.5$. Hence, there is a total of $600$ clients ($15$ languages $\cdot$ $20$ clients $\cdot$ $2$ pools), consisting of both label and feature heterogeneity.

For MasakhaNEWS, we first split the data in each of the $16$ languages by half for each pool. Similar to our XNLI setup, we divide each language's data equally among 10 clients and adopt LDA with $\alpha=0.5$, resulting in $320$ clients in total. Differing from our XNLI setup, each language varies in the amount of samples, adding another layer of data heterogeneity to the setup: quantity skew.

\begin{figure}
    \small
    \centering
    \includegraphics[width=0.9\linewidth,trim={0.58cm 0cm 0cm 0cm},clip]{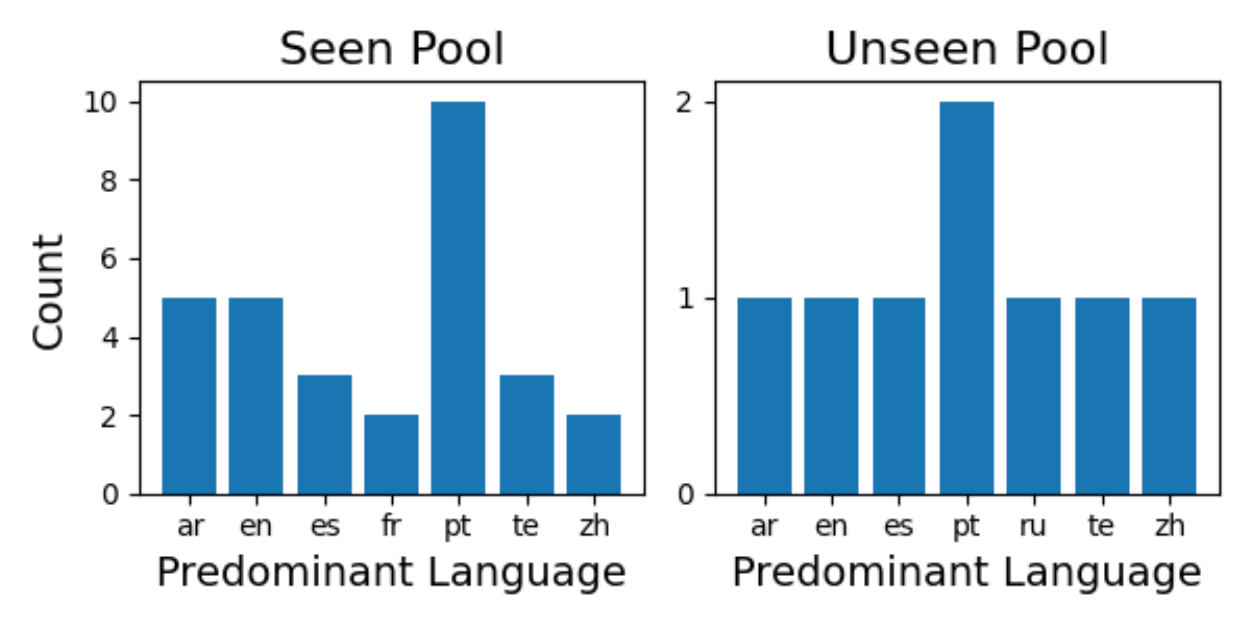}
    \caption{No. of clients by predominant language in our Fed-Aya setup}
    \label{fig:fed-aya}
\end{figure}

\noindent\textbf{Instruction-Tuning Generation.}~We use pretrained MobileLLaMA-1.4B~\cite{mobilellama} and Llama-3.2-3B~\cite{llama3}, which are representative of commonly supported model sizes on high-end edge devices~\cite{2024_mobilequant}\footnote{Refer to Appendix Sec~\ref{app:additional_results} for experiments with Llama-3.1-8B.}, and run experiments with them on the Fed-Aya dataset~\cite{fedllm-bench}. Fed-Aya, a real-world FL dataset naturally partitioned by annotator ID, comprises 38 clients, $8$ of whom we designate as our \unseen{} pool. We split each client's data $80\%/20\%$ for training and testing. Each client has up to $4$ languages and Fig.~\ref{fig:fed-aya} shows the predominant language (client's most frequent language) distribution in our setup. 

\begin{table}[t]
\centering
\small
\begin{tabular}{c|ccccc}
\toprule

\textbf{Lan}     & \multicolumn{1}{c|}{\textbf{LoRA}} & \multicolumn{1}{c|}{\textbf{AdaLoRA}} & \multicolumn{1}{c|}{\textbf{BT-LoRA}} & \multicolumn{1}{c|}{\textbf{FedL2P}} & \textbf{\method{}} \\ \toprule
eng                   & \multicolumn{1}{c|}{90.4±0.1}      & \multicolumn{1}{c|}{89.9±0.0}         & \multicolumn{1}{c|}{89.9±0.0}                & \multicolumn{1}{c|}{90.7±0.6}        & \textbf{92.0±0.0}               \\ \hline
som                   & \multicolumn{1}{c|}{60.1±0.3}      & \multicolumn{1}{c|}{59.4±0.3}         & \multicolumn{1}{c|}{59.6±0.3}                & \multicolumn{1}{c|}{61.0±1.2}        & \textbf{65.5±1.2}               \\ \hline
run                   & \multicolumn{1}{c|}{81.4±0.5}      & \multicolumn{1}{c|}{81.2±0.3}         & \multicolumn{1}{c|}{81.4±0.0}                & \multicolumn{1}{c|}{82.0±0.9}        & \textbf{87.8±0.3}               \\ \hline
fra                   & \multicolumn{1}{c|}{88.6±0.0}      & \multicolumn{1}{c|}{88.6±0.0}         & \multicolumn{1}{c|}{88.6±0.0}                & \multicolumn{1}{c|}{89.1±0.7}        & \textbf{93.5±0.2}               \\ \hline
lin                   & \multicolumn{1}{c|}{83.5±0.5}      & \multicolumn{1}{c|}{82.8±0.0}         & \multicolumn{1}{c|}{83.1±0.5}                & \multicolumn{1}{c|}{83.9±0.0}        & \textbf{88.5±0.9}               \\ \hline
ibo                   & \multicolumn{1}{c|}{79.8±0.2}      & \multicolumn{1}{c|}{79.0±0.0}         & \multicolumn{1}{c|}{79.0±0.0}                & \multicolumn{1}{c|}{79.7±0.2}        & \textbf{82.6±0.0}               \\ \hline
amh                   & \multicolumn{1}{c|}{45.7±0.0}      & \multicolumn{1}{c|}{45.2±0.0}         & \multicolumn{1}{c|}{45.2±0.0}                & \multicolumn{1}{c|}{45.7±0.0}        & \textbf{52.0±0.2}               \\ \hline
hau                   & \multicolumn{1}{c|}{75.8±0.0}      & \multicolumn{1}{c|}{75.0±0.1}         & \multicolumn{1}{c|}{75.2±0.0}                & \multicolumn{1}{c|}{76.7±1.1}        & \textbf{79.7±0.3}               \\ \hline
pcm                   & \multicolumn{1}{c|}{96.0±0.0}      & \multicolumn{1}{c|}{96.0±0.0}         & \multicolumn{1}{c|}{96.0±0.0}                & \multicolumn{1}{c|}{96.0±0.0}        & \textbf{97.6±0.3}               \\ \hline
swa                   & \multicolumn{1}{c|}{79.0±0.0}      & \multicolumn{1}{c|}{78.6±0.0}         & \multicolumn{1}{c|}{78.6±0.0}                & \multicolumn{1}{c|}{79.7±0.4}        & \textbf{84.7±0.5}               \\ \hline
orm                   & \multicolumn{1}{c|}{64.2±0.0}      & \multicolumn{1}{c|}{64.0±0.3}         & \multicolumn{1}{c|}{64.0±0.3}                & \multicolumn{1}{c|}{64.4±0.3}        & \textbf{72.2±1.0}               \\ \hline
xho                   & \multicolumn{1}{c|}{69.1±0.3}      & \multicolumn{1}{c|}{69.6±0.0}         & \multicolumn{1}{c|}{69.6±0.0}                & \multicolumn{1}{c|}{69.1±0.3}        & \textbf{76.3±0.6}               \\ \hline
yor                   & \multicolumn{1}{c|}{79.2±0.2}      & \multicolumn{1}{c|}{78.9±0.2}         & \multicolumn{1}{c|}{79.0±0.0}                & \multicolumn{1}{c|}{79.7±0.6}        & \textbf{82.1±0.2}               \\ \hline
sna                   & \multicolumn{1}{c|}{78.8±0.0}      & \multicolumn{1}{c|}{78.8±0.0}         & \multicolumn{1}{c|}{78.8±0.0}                & \multicolumn{1}{c|}{78.8±0.0}        & \textbf{81.3±0.7}               \\ \hline
lug                   & \multicolumn{1}{c|}{67.6±0.0}      & \multicolumn{1}{c|}{67.6±0.0}         & \multicolumn{1}{c|}{67.6±0.0}                & \multicolumn{1}{c|}{67.9±0.4}        & \textbf{69.1±0.4}               \\ \hline
tir                   & \multicolumn{1}{c|}{44.9±0.0}      & \multicolumn{1}{c|}{44.9±0.0}         & \multicolumn{1}{c|}{44.9±0.0}                & \multicolumn{1}{c|}{45.3±0.7}        & \textbf{63.5±0.3}               \\ \bottomrule
\end{tabular}
\caption{Mean±SD Accuracy of each language across 3 different seeds for \seen{} clients of our MasakhaNEWS setup ($r=2$, see Appendix Table~\ref{tab:masakha_seen_full} for other $r$ values). The pretrained model is trained using \textbf{Standard FL with full fine-tuning} and the resulting \basemodel{} is personalized to each client given a baseline approach.}
\label{tab:masakha_seen}
\end{table}

\begin{table}[t]
\centering
{
\small
\begin{tabular}{c|c|c|c|c|c}
\toprule
\textbf{Lan} & \textbf{LoRA}     & \textbf{AdaLoRA} & \textbf{BT-LoRA} & \textbf{FedL2P} & \textbf{\method{}} \\ \toprule
eng            & \textbf{90.7±0.0} & 90.3±0.0         & 90.3±0.0                & 90.6±0.1        & \textbf{90.7±0.1}                        \\ \hline
som            & 68.5±0.3          & 67.3±0.0         & 67.3±0.0                & 68.5±0.3        & \textbf{72.1±0.6}               \\ \hline
run            & 82.0±0.5          & 80.8±0.0         & 80.8±0.0                & 82.6±0.0        & \textbf{88.4±0.3}               \\ \hline
fra            & 84.4±0.0          & 84.4±0.0         & 84.4±0.0                & 84.4±0.0        & \textbf{88.2±0.4}               \\ \hline
lin            & 79.5±0.0          & 79.5±0.0         & 79.5±0.0                & 79.5±0.0        & \textbf{86.4±0.0}               \\ \hline
ibo            & 76.8±0.2          & 76.9±0.0         & 76.9±0.0                & 76.4±0.4        & \textbf{81.5±0.4}               \\ \hline
amh            & 46.3±0.0          & 46.3±0.0         & 45.9±0.2                & 46.8±0.8        & \textbf{51.1±0.0}               \\ \hline
hau            & 75.5±0.1          & 74.8±0.1         & 74.6±0.0                & 76.0±0.6        & \textbf{79.5±0.3}               \\ \hline
pcm            & 90.2±0.0          & 90.2±0.0         & 90.2±0.0                & 90.2±0.0        & \textbf{93.7±0.3}               \\ \hline
swa            & 75.6±0.3          & 75.9±0.2         & 75.5±0.2                & 75.9±0.4        & \textbf{78.0±0.4}               \\ \hline
orm            & 62.0±0.0          & 61.5±0.3         & 61.5±0.3                & 62.2±0.3        & \textbf{73.0±0.5}               \\ \hline
xho            & 64.2±0.6          & 64.2±0.3         & 63.8±0.0                & 64.4±0.9        & \textbf{78.5±1.1}               \\ \hline
yor            & 80.1±0.0          & 79.6±0.0         & 79.5±0.2                & 80.3±0.2        & \textbf{83.7±0.2}               \\ \hline
sna            & 74.6±0.0          & 74.6±0.0         & 74.4±0.2                & 74.8±0.3        & \textbf{80.0±0.4}               \\ \hline
lug            & 65.2±0.0          & 65.2±0.0         & 65.2±0.0                & 66.1±0.0        & \textbf{69.6±0.0}               \\ \hline
tir            & 41.9±0.0          & 41.9±0.0         & 42.6±0.0                & 41.9±0.0        & \textbf{58.3±0.3}               \\ \bottomrule
\end{tabular}
}
\caption{Mean±SD Accuracy of each language for \unseen{} clients of our MasakhaNEWS setup ($r=2$, see Appendix Table~\ref{tab:masakha_unseen_full} for other $r$ values). The pretrained model is trained using \textbf{Standard FL with full fine-tuning} and the \basemodel{} is personalized to each client given a baseline approach.}
\label{tab:masakha_unseen}
\end{table}

\subsection{Setup: Complementary Approaches}\label{sec:complementary}

\method{} is compatible with off-the-shelf models and models trained with existing FL approaches. Concretely, given a pretrained model, we obtain a \basemodel{} using one of the following approaches: 

\noindent\textbf{Standard FL.} We further train the pretrained model federatedly on the \seen{} pool, either using existing PEFT methods or full fine-tuning~\cite{fedllm-bench, fedpeft},  

\noindent\textbf{Personalized FL (pFL).} We adopt two recent pFL works: \textit{i)} FedDPA-T~\cite{FedDPA}, which learns per-client personalized LoRA modules in addition to global LoRA modules, and \textit{ii)} DEPT (SPEC)~\cite{DEPT}, which learns per-client personalized token and positional embeddings while keeping the rest of the model shared. The \basemodel{} hence differs for each client. 

\noindent\textbf{Off-the-shelf.} We use the pretrained model as the \basemodel{} without additional training.

\subsection{Setup: Baselines}\label{sec:baselines}

Given a \basemodel{}, we compare \method{} with existing fine-tuning and {\em learning to personalize} approaches. For each baseline, we either follow best practices recommended by the corresponding authors or employ a simple grid search and pick the best performing hyperparameters (Appendix~\ref{appendix:experiments}).

\noindent\textbf{LoRA PEFT.}~We deploy LoRA~\cite{hu2021lora} on all linear layers of the model with a fixed rank $r$. 

\noindent\textbf{Non-FL Rank Selection.}~We compare with AdaLoRA~\cite{adalora} and our proposed LoRA-variant of BayesTune~\cite{kim2023bayestune}, BT-LoRA (Section~\ref{sec:personalized_peft}), which optimizes $\bm{\lambda}$ separately for each client. 

\noindent\textbf{FL to Personalize.}~We compare with FedL2P~\cite{royson2023fedl2p} which trains a MLP federatedly to output per-client learning rates for each LoRA module.

\subsection{Results on Text Classification}\label{sec:text_class}

We evaluate our approach in a typical FL setup, where the pretrained model is first trained using Standard FL with full fine-tuning and the resulting \basemodel{} is then personalized to each client. Tables~\ref{tab:masakha_seen} \& \ref{tab:masakha_unseen} show the mean and standard deviation (SD) of the accuracy for each language in our MasakhaNEWS setup for \seen{} and \unseen{} pool respectively (similarly for XNLI in Appendix Tables~\ref{tab:xnli_seen} \& \ref{tab:xnli_unseen}). 

The results in all four tables show that federated {\em learning to personalize} methods (FedL2P and \method{}) outperform the other baselines in most cases. Non-FL rank selection approaches (AdaLoRA and BT-LoRA), on the other hand, tend to overfit and/or struggle to learn an optimal rank structure given the limited number of samples in each client. Comparing FedL2P and \method{}, \method{} largely surpass FedL2P with a few exceptions, indicating that learning to personalize LoRA rank structure is the better hyperparameter choice than personalizing learning rates; this finding is also aligned with recent LLM-based optimizer findings~\cite{zhao2025deconstructing}, which shows that Adam's performance is robust with respect to its learning rate.

\noindent\textbf{\method{}'s pFL Compatibility.}
Apart from Standard FL, we show that \method{} can be plugged into existing pFL works that trains both a subset of the pretrained model and personalized layers for each client. Table~\ref{tab:xnli_seen_feddpa} and Appendix Table~\ref{tab:xnli_seen_dept} show that \method{} outperforms baselines in almost all cases in our XNLI setup given a \basemodel{} trained using FedDPA-T~\cite{FedDPA} and DEPT(SPEC)~\cite{DEPT} respectively. In short, \method{} can be integrated into a larger family of existing pFL approaches, listed in Section~\ref{sec:related}, to further improve personalization performance. 


\begin{table}[t]
\centering
{
\small
\begin{tabular}{c|c|c|c|c|c}
\toprule
\textbf{Lan} & \textbf{LoRA} & \textbf{AdaLoRA} & \textbf{BT-LoRA} & \textbf{FedL2P} & \textbf{\method{}} \\ \toprule
bg             & 45.8±0.3      & 44.1±0.1         & 43.8±0.0                & 47.5±2.8        & \textbf{64.4±0.2}               \\ \hline
hi             & 42.8±0.2      & 41.2±0.0         & 40.6±0.0                & 44.5±3.2        & \textbf{57.8±1.3}               \\ \hline
es             & 48.7±0.2      & 47.5±0.1         & 47.2±0.0                & 50.3±2.9        & \textbf{58.5±1.3}               \\ \hline
el             & 50.9±0.1      & 50.0±0.0         & 50.0±0.0                & 51.5±1.4        & \textbf{59.7±2.6}               \\ \hline
vi             & 53.0±0.0      & 52.4±0.0         & 52.4±0.0                & 53.6±0.7        & \textbf{60.8±0.7}               \\ \hline
tr             & 48.0±0.3      & 46.5±0.1         & 46.2±0.0                & 50.1±2.8        & \textbf{58.9±2.9}               \\ \hline
de             & 49.9±0.1      & 48.0±0.0         & 47.4±0.2                & 50.9±2.6        & \textbf{55.0±0.2}               \\ \hline
ur             & 41.9±0.1      & 38.8±0.0         & 38.8±0.0                & 44.8±4.5        & \textbf{63.7±0.2}               \\ \hline
en             & 46.5±0.3      & 44.7±0.1         & 44.3±0.1                & 49.1±4.3        & \textbf{55.9±0.2}               \\ \hline
zh             & 44.4±0.2      & 42.2±0.0         & 41.6±0.0                & 46.5±4.0        & \textbf{56.3±0.3}               \\ \hline
th             & 42.5±0.2      & 40.7±0.1         & 40.5±0.1                & 44.4±3.7        & \textbf{58.3±0.2}               \\ \hline
sw             & 51.8±0.2      & 50.4±0.0         & 49.8±0.0                & 52.3±1.3        & \textbf{59.5±0.5}               \\ \hline
ar             & 46.9±0.2      & 45.0±0.0         & 44.7±0.1                & 48.4±3.0        & \textbf{55.2±0.2}               \\ \hline
fr             & 48.1±0.2      & 46.0±0.0         & 46.0±0.0                & 49.5±3.0        & \textbf{57.5±0.2}               \\ \hline
ru             & 50.5±0.2      & 48.8±0.0         & 48.1±0.2                & 51.6±2.3        & \textbf{55.2±0.4}               \\ \bottomrule
\end{tabular}
}
\caption{Mean±SD Accuracy of each language across 3 seeds for clients in the \seen{} pool of our XNLI setup ($r=2$, see Appendix Table~\ref{tab:xnli_seen_feddpa_full} for other $r$ values). The pretrained model is trained using \textbf{FedDPA-T} and the resulting \basemodel{} is personalized to each client given a baseline approach. See Appendix Table~\ref{tab:xnli_seen_dept} for results with \textbf{DEPT (SPEC)}.}
\label{tab:xnli_seen_feddpa}
\end{table}

\subsection{Results on Instruction-Tuning Generation}\label{sec:ift_gen}

We evaluate our approach on the more challenging real-world multilingual benchmark, Fed-Aya. Tables~\ref{tab:lama_fedaya_seen} and \ref{tab:lama_fedaya_unseen} show the average METEOR~\cite{meteor}/ROUGE-L~\cite{ROUGE}\footnote{Due to limited space, we include ROUGE-1 in the Appendix.} of each language given the off-the-shelf instruction finetuned Llama-3.2-3B (Llama-3.2-3B-Instruct) for \seen{} and \unseen{} clients respectively. Similarly, in Appendix Tables~\ref{tab:mobilellama_fedaya_seen} and \ref{tab:mobilellama_fedaya_unseen}, we show the same tables given a pretrained MobileLLaMA-1.4B model trained using Standard FL with LoRA following the training recipe from FedLLM-Bench~\cite{fedllm-bench}. These two models represent scenarios where the \basemodel{} may or may not be trained using FL. 

In all four tables, \method{} outperforms baselines in most scenarios. We also observe that FedL2P mostly underperforms standard baselines, a phenomenon also observed for our XNLI setup when the \basemodel{} is trained with FedDPA-T (Appendix Table~\ref{tab:xnli_seen_feddpa_full}). We hypothesize that the inner-loop optimization in FedL2P fail to reach a stationary point\footnote{FedL2P relies on the implicit function theorem for hypergradient computation.} due to the inherent task difficulty (Fed-Aya) or a less-performant \basemodel{}, resulting in a sub-optimal hypergradient and downstream performance.

\noindent\textbf{\method{} Limitations.}\label{sec:limitations} In some cases, \method{} falls short, especially in the recall performance (ROUGE), such as Russian (\textit{ru}) and French (\textit{fr}) for \unseen{} clients for both {\em base models} in most scenarios. These cases highlight two limitations of our approach: 

\textit{i)} Challenge with zero-shot transfer. \textit{ru} is not seen by PSG during federated training; there are no \textit{ru} samples, nor other languages similar to \textit{ru}, in the \seen{} pool, resulting in worse \textit{ru} performance. Hence, we do not expect a similar outcome in datasets with a more diverse pool of clients. 

\textit{ii)} Predominant language dependence. None of the clients in the \unseen{} pool have \textit{fr} as a predominant language (Fig.~\ref{fig:fed-aya}). Because the number of predominant language samples heavily influences $\bm{\lambda}$, the performance for \textit{fr} is worse. This predominant language bias is quantified and future potential directions are discussed in Appendix Section~\ref{app:additional_results}.

\begin{table}[t]
{
\small
\begin{tabular}{c|c|c|c|c|c}
\toprule
\textbf{Lan} & \textbf{LoRA}  & \textbf{AdaLoRA}                                                   & \textbf{BT-LoRA}                         & \textbf{FedL2P} & \textbf{\method{}}          \\ \toprule
te                & \textbf{0.24}/0.13 & \textbf{0.24}/\textbf{0.14}                                                    & \textbf{0.24}/\textbf{0.14} & 0.23/0.13  & \textbf{0.24}/\textbf{0.14} \\ \hline
ar                & 0.34/0.07 & 0.34/0.07                                                     & 0.32/0.06                           & 0.33/0.07  & \textbf{0.37/0.08}                  \\ \hline
es                & \textbf{0.39}/0.39 & \textbf{0.39/0.40}                                            & 0.38/0.38                           & 0.38/0.38  & \textbf{0.39}/0.39                           \\ \hline
en                & 0.33/0.31 & 0.35/0.33                                                     & 0.29/0.23                           & 0.33/0.30  & \textbf{0.37/0.36}                  \\ \hline
fr                & 0.29/0.30 & 0.29/0.29                                                     & 0.29/0.29                           & 0.29/0.30  & \textbf{0.34/0.33}                  \\ \hline
zh                & \textbf{0.11}/\textbf{0.12} & \textbf{0.11}/\textbf{0.12} & 0.09/0.11                           & 0.09/\textbf{0.12}  & \textbf{0.11}/\textbf{0.12} \\ \hline
pt                & 0.38/0.41 & 0.38/0.41                                                     & 0.37/0.39                           & 0.38/0.40  & \textbf{0.40/0.42}                  \\ \bottomrule
\end{tabular}
}
\caption{Avg. METEOR/ROUGE-L for \seen{} clients in our Fed-Aya setup. ($r=2$, see Appendix Table~\ref{tab:lama_fedaya_seen_full} for other $r$ values) {\em Base model} is off-the-shelf Llama-3.2-3B-Instruct.}
\label{tab:lama_fedaya_seen}
\end{table}
\begin{table}[t]
{
\small
\begin{tabular}{c|ccccc}
\toprule
\textbf{Lan} & \multicolumn{1}{c|}{\textbf{LoRA}}                            & \multicolumn{1}{c|}{\textbf{AdaLoRA}}                         & \multicolumn{1}{c|}{\textbf{BT-LoRA}}                         & \multicolumn{1}{c|}{\textbf{FedL2P}}                          & \textbf{\method{}}          \\ \toprule
te            & \multicolumn{1}{c|}{0.16/0.08}                           & \multicolumn{1}{c|}{0.16/\textbf{0.11}}                           & \multicolumn{1}{c|}{\textbf{0.17}/0.09} & \multicolumn{1}{c|}{0.16/\textbf{0.11}} & \textbf{0.17}/0.07                           \\ \hline
ar            & \multicolumn{1}{c|}{0.24/0.04}                           & \multicolumn{1}{c|}{0.23/\textbf{0.07}} & \multicolumn{1}{c|}{0.22/0.05}                           & \multicolumn{1}{c|}{0.24/0.05}                           & \textbf{0.26}/0.05 \\ \hline
es            & \multicolumn{1}{c|}{0.43/0.44}                           & \multicolumn{1}{c|}{0.43/0.44}                           & \multicolumn{1}{c|}{0.38/0.41}                           & \multicolumn{1}{c|}{0.41/0.43}                           & \textbf{0.44/0.48}                  \\ \hline
en            & \multicolumn{1}{c|}{\textbf{0.32}/0.25} & \multicolumn{1}{c|}{\textbf{0.32}/0.25}                           & \multicolumn{1}{c|}{\textbf{0.32}/0.24}                           & \multicolumn{1}{c|}{0.31/0.24}                           & 0.31/\textbf{0.27} \\ \hline
fr            & \multicolumn{1}{c|}{\textbf{0.55/0.67}}                           & \multicolumn{1}{c|}{\textbf{0.55/0.67}}                           & \multicolumn{1}{c|}{0.40/0.33}                           & \multicolumn{1}{c|}{0.40/0.33}                           & 0.22/0.22                           \\ \hline
zh            & \multicolumn{1}{c|}{0.25/0.00}                           & \multicolumn{1}{c|}{0.25/0.00}                           & \multicolumn{1}{c|}{0.23/0.00}                           & \multicolumn{1}{c|}{0.23/0.00}                           & \textbf{0.28/0.01}                  \\ \hline
pt            & \multicolumn{1}{c|}{\textbf{0.36}/0.40} & \multicolumn{1}{c|}{0.33/0.40}                           & \multicolumn{1}{c|}{0.31/0.36}                           & \multicolumn{1}{c|}{0.34/0.39}                           & 0.35/\textbf{0.41} \\ \hline
ru            & \multicolumn{1}{c|}{0.22/0.17}                           & \multicolumn{1}{c|}{0.23/0.17}                           & \multicolumn{1}{c|}{0.25/0.18}                           & \multicolumn{1}{c|}{0.25/0.20}                           & \textbf{0.34/0.27}                  \\\bottomrule

\end{tabular}
}
\caption{Avg. METEOR/ROUGE-L for \unseen{} clients in our Fed-Aya setup ($r=2$, see Appendix Table~\ref{tab:lama_fedaya_unseen_full} for other $r$ values). {\em Base model} is off-the-shelf Llama-3.2-3B-Instruct.}
\label{tab:lama_fedaya_unseen}
\end{table}

\subsection{Cost of \method{}}\label{sec:cost}

Table~\ref{tab:cost} shows the mean latency and the peak memory usage across 100 runs on the first client in the \seen{} pool for $r=16$ using an Nvidia A100 GPU. Non-FL baselines do not incur a federated training cost while FL approaches requires training the PSG. Comparing FedL2P and \method{}, \method{} does not require second-order optimization, resulting in better efficiency. Note that FedL2P needs to be run for every $r$ while \method{} runs once for all targeted ranks.

For communication costs, not shown in the table, \method{} is more costly as it predicts per LoRA rank while FedL2P predicts per layer. Nonetheless, these costs are negligible compared to running FL on the \basemodel{}; FedL2P uses 0.02\% and 0.002\% and \method{} uses 0.2\% and 0.16\% of the parameters of mBERT and Llama-3.2-3B respectively.

During inference, FL-based approaches incur an additional forward pass of \basemodel{} and the PSG compared to non-FL approaches. Memory-wise, \method{} results in the smallest memory footprint for autoregressive generation as the PSG learns not to attach LoRA modules $\lambda_l=0$ on some layers, skipping {\em matmul} operations entirely. More details discussions on cost can be found in Appendix~\ref{app:cost_device}.

\begin{table}[t]
\centering
\begin{threeparttable}
{
\small
\begin{tabular}{cccc}
\toprule
\multicolumn{4}{c}{\textbf{Federated Training}} \\ \midrule
\multicolumn{1}{c}{\textbf{\begin{tabular}[c]{@{}c@{}}Dataset \\ (Model)\end{tabular}}}                 & \multicolumn{1}{c}{\textbf{Approach}}      & \multicolumn{1}{c}{\textbf{\begin{tabular}[c]{@{}c@{}}Mean\\ Latency (s)\end{tabular}}} & \textbf{\begin{tabular}[c]{@{}c@{}}Peak\\ Memory (GB)\end{tabular}} \\ \hline

\multicolumn{1}{c}{\multirow{2}{*}{\begin{tabular}[c]{@{}c@{}}XNLI \\ (mBERT)\end{tabular}}}          & \multicolumn{1}{c}{FedL2P}                 & \multicolumn{1}{c}{28.2}                                                                & 3.5                                                                 \\ \cline{2-4} 
\multicolumn{1}{c}{}                                                                                  & \multicolumn{1}{c}\method{} & \multicolumn{1}{c}{3.4}                                                                 & 3.1                                                                 \\ \hline
\multicolumn{1}{c}{\multirow{2}{*}{\begin{tabular}[c]{@{}c@{}}Fed-Aya\\ (Llama-3 3B)\end{tabular}}} & \multicolumn{1}{c}{FedL2P}                 & \multicolumn{1}{c}{226.5}                                                               & 32.2                                                                \\ \cline{2-4} 
\multicolumn{1}{c}{}                                                                                  & \multicolumn{1}{c}\method{} & \multicolumn{1}{c}{80.3}                                                                & 25.2                                                                \\ \bottomrule\toprule
\multicolumn{4}{c}{\textbf{Inference} }                                                                                                                  \\ \midrule
\multicolumn{1}{c}{\textbf{\begin{tabular}[c]{@{}c@{}}Dataset \\ (Model)\end{tabular}}}                 & \multicolumn{1}{c}{\textbf{Approach}}      & \multicolumn{1}{c}{\textbf{\begin{tabular}[c]{@{}c@{}}Mean\\ Latency (s)\end{tabular}}} & \textbf{\begin{tabular}[c]{@{}c@{}}Peak\\ Memory (GB)\end{tabular}} \\ \hline
\multicolumn{1}{c}{\multirow{5}{*}{\begin{tabular}[c]{@{}c@{}}XNLI\\ (mBERT)\end{tabular}}}           & \multicolumn{1}{c}{LoRA}                   & \multicolumn{1}{c}{3.1}                                                                 & 2.6                                                                 \\ \cline{2-4} 
\multicolumn{1}{c}{}                                                                                  & \multicolumn{1}{c}{AdaLoRA}                & \multicolumn{1}{c}{3.4}                                                                 & 2.7                                                                 \\ \cline{2-4} 
\multicolumn{1}{c}{}                                                                                  & \multicolumn{1}{c}{BT-LoRA}                & \multicolumn{1}{c}{2.4\tnote{1} + 3.1}                                 & 3.1\tnote{1} + 2.7                                 \\ \cline{2-4} 
\multicolumn{1}{c}{}                                                                                  & \multicolumn{1}{c}{FedL2P}                 & \multicolumn{1}{c}{4.5}                                                                 & 3.0                                                                 \\ \cline{2-4} 
\multicolumn{1}{c}{}                                                                                  & \multicolumn{1}{c}\method{} & \multicolumn{1}{c}{4.4}                                                                 & 2.7                                                                 \\ \hline
\multicolumn{1}{c}{\multirow{5}{*}{\begin{tabular}[c]{@{}c@{}}Fed-Aya\\ (Llama-3 3B)\end{tabular}}} & \multicolumn{1}{c}{LoRA}                   & \multicolumn{1}{c}{347.3}                                                               & 17.9                                                                \\ \cline{2-4} 
\multicolumn{1}{c}{}                                                                                  & \multicolumn{1}{c}{AdaLoRA}                & \multicolumn{1}{c}{423.9}                                                               & 18.8                                                                \\ \cline{2-4} 
\multicolumn{1}{c}{}                                                                                  & \multicolumn{1}{c}{BT-LoRA}                & \multicolumn{1}{c}{72.3\tnote{1} + 357.6}                              & 25.2\tnote{1} + 18.3                               \\ \cline{2-4} 
\multicolumn{1}{c}{}                                                                                  & \multicolumn{1}{c}{FedL2P}                 & \multicolumn{1}{c}{380.0}                                                               & 19.8                                                                \\ \cline{2-4} 
\multicolumn{1}{c}{}                                                                                  & \multicolumn{1}{c}\method{} & \multicolumn{1}{c}{400.6}                                                               & 15.7                                                                \\ \bottomrule
\end{tabular}
\begin{tablenotes}
\item[1] One time cost per client for all targeted ranks
\end{tablenotes}
}
\end{threeparttable}
\caption{Mean latency and memory costs across 100 runs of the first client in the \seen{} pool using an NVIDIA A100 GPU. }
\label{tab:cost}
\end{table}

\begin{figure}
    \small
    \centering
    \includegraphics[width=0.88\linewidth,trim={0cm 3.4cm 0cm 5.0cm},clip]{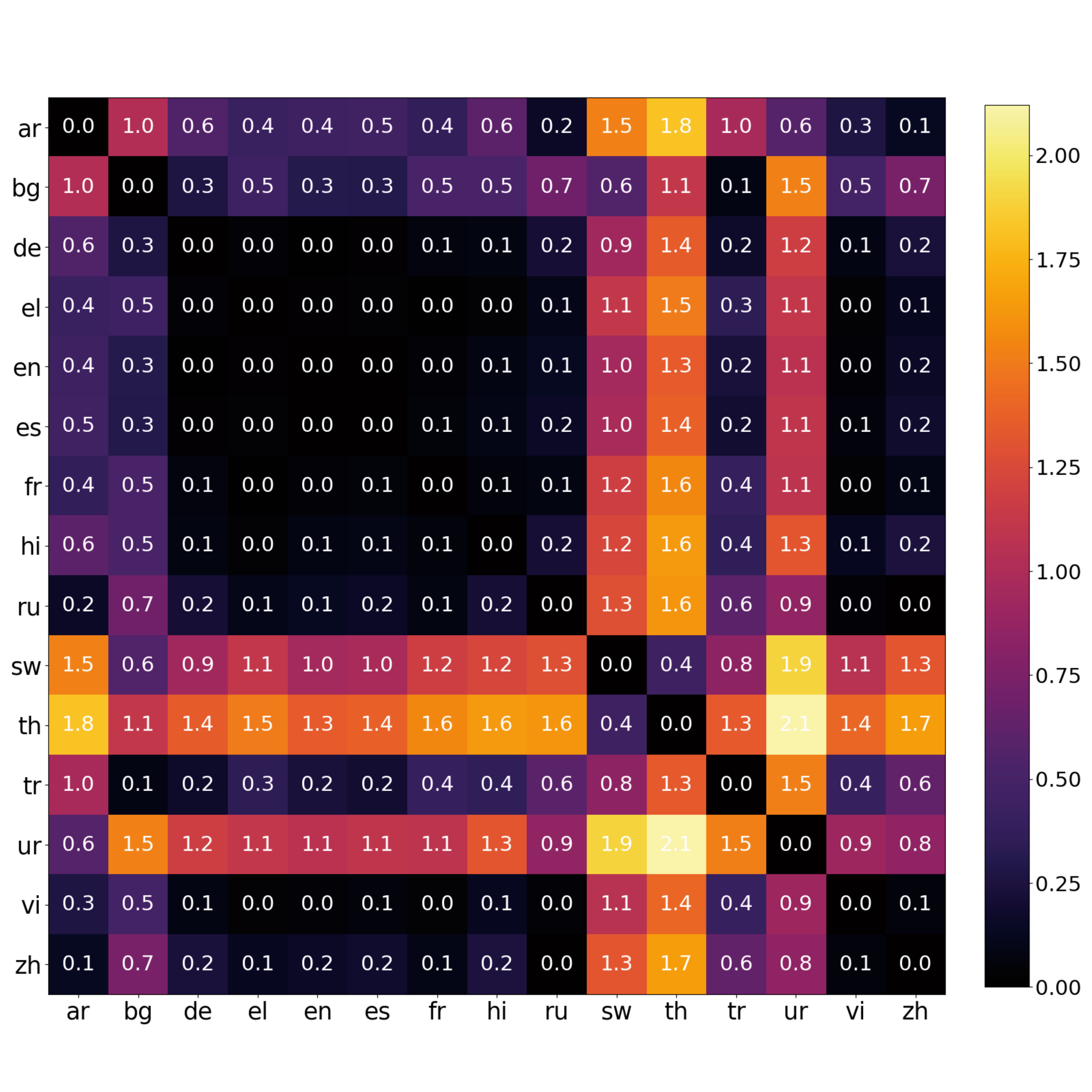}
    \caption{Cross-lingual $\bm{\lambda}$ distance in our XNLI setup (See Appendix Fig.~\ref{fig:masakha_out} for our MasakhaNEWS setup). Each block shows the log-scale normalized average Euclidean distances between all pairs of clients' $\bm{\lambda}$ in their respective languages. The smaller the distance, the more similar $\bm{\lambda}$ is. }
    \label{fig:xnli_out}
\end{figure}

\subsection{Further Analysis}\label{sec:analysis}

We further analyze $\bm{\lambda}$ and how they differ across languages. Surprisingly, we find that \method{} learns language-agnostic rank structures. In other words, depending on the task and the \basemodel{}, the rank structure of $\bm{\lambda}$ is fixed across languages. For instance, in the case where $r=2$, \method{} allocate ranks to dense layers instead of attention blocks. With more budget, \textit{e.g.},~$r=16$, \method{} allocates more rank to either the query attention layer or the value attention layer depending on the setup. We show these rank structures across all setups for $r=2$ and $r=16$ in Appendix Fig.~\ref{fig:xnli_fedavg_out_r16}-\ref{fig:llama3_fedavg_out_r2}.

While the rank structure is the same across languages, the rank-wise scales (absolute values of $\bm{\lambda}$) differ. Following FedL2P, we visualize the difference in $\bm{\lambda}$ for different languages using the normalized mean distance, $d(j,k)$, between all clients pairs holding data for languages $j$ and $k$. Fig.~\ref{fig:xnli_out} and Appendix Fig.~\ref{fig:masakha_out} show these distances for XNLI and MasakhaNEWS setup respectively. Specifically, the value of each block in each figure is computed as follows: $\log(\frac{d(j,k)}{\sqrt{d(j,j)}\sqrt{d(k,k)}})$. Hence, the smaller the distance, the more similar $\bm{\lambda}$ is between languages. The results are aligned with our intuition that similar languages have similar $\bm{\lambda}$. For instance, the closest language to Urdu (\textit{ur}) is Arabic (\textit{ar}), both of which have the closest $\bm{\lambda}$ similarity (Fig.~\ref{fig:xnli_out}); likewise, for Tigrinya (\textit{tir}) and Amharic (\textit{amh}) in Appendix Fig.~\ref{fig:masakha_out}. We also observe that unrelated languages have similar $\bm{\lambda}$, \textit{e.g.},~Mandarin (\textit{zh}) and Vietnamese (\textit{vi}) share similar $\bm{\lambda}$ with the Indo-European languages (Fig.~\ref{fig:xnli_out}). This finding adds to existing evidence that leveraging dissimilar languages can sometimes benefit particular languages~\cite{fedllm-bench}.

\section{Conclusion}\label{sec:conclusion}

In this work, we tackle language personalization through \method{}, a federated learning method that learns how to perform PEFT on heterogeneous data. We show that our proposed federated {\em learning-to-personalize} approach is easily pluggable to off-the-shelf LLMs and standard and personalized FL methods alike, surpassing other personalized fine-tuning baselines in most cases. Our results show that \method{} automatically learns model- and task-specific language-agnostic LoRA rank structures as well as effective cross-lingual transfers, where both diverse low- and high-resource languages can share similar LoRA rank magnitudes. Despite clear advantages, our approach falls short in personalizing for each client's minority languages, as the personalized solution is skewed towards their predominant language. Nonetheless, our work is a significant step towards successfully merging the benefits of multilingual learning and personalized FL.

\bibliography{aaai2026}


\appendix
\setcounter{secnumdepth}{2}
\renewcommand{\thesubsection}{\Alph{section}.\arabic{subsection}}
\clearpage
\newpage
\lstset{language=Python,basicstyle=\small\ttfamily,columns=fullflexible}
\section{Detailed Experimental Setup}\label{appendix:experiments}

For reproducibility and completeness, we provide code, along with comprehensive details of all setups, datasets, tasks, models, baselines, and hyperparameters.

\subsection{Tasks, Datasets, and Data Partitioning}

\noindent\textbf{XNLI~\cite{XNLI}.} A natural language inference benchmark dataset for evaluating cross-lingual understanding covering 15 diverse languages including both high- and low-resources languages: English, French, Spanish, German, Greek, Bulgarian, Russian, Turkish, Arabic, Vietnamese, Thai, Chinese, Hindi, Swahili and Urdu. XNLI consists of premise-hypothesis pairs, labeled as entailment, contradiction, or neutral across different languages. We sample 2k instances from the XNLI train split and 500 instances from the test split for each pool. The data is then divided equally among 20 clients for each language using the latent Dirichlet allocation (LDA) partition with $\alpha=0.5$. Hence, the total number of clients is 600 (15 languages $\cdot$ 20 clients per language $\cdot$ 2 pools).

\noindent\textbf{MasakhaNEWS~\cite{MasakhaNEWS}.} A news topic classification benchmark designed to address the lack of resources for African languages. It covers 2 high-resource languages, English and French, and 14 low-resource languages, namely Amharic, Hausa, Igbo, Lingala, Luganda, Naija, Oromo, Rundi, chiShona, Somali, Kiswahili, Tigrinya, isiXhosa, and Yorùbá. Each sample contains a headline, the body text, and one of the 7 labels: business, entertainment, health, politics, religion, sports, or technology. We first combine all samples from the MasakhaNEWS train and validation split to form our train set, and use the MasakhaNEWS test split as our test set. We then split both train and test in each of the 16 languages by half for each pool. Following our XNLI setup, we adopt LDA $\alpha=0.5$ and divide each language's data equally into 10 clients. Hence, the total number of clients is 320 (16 languages $\cdot$ 10 clients per language $\cdot$ 2 pools). Note that unlike XNLI, the number of samples for each language differs, hence there is quantity skew across clients. 

\noindent\textbf{Fed-Aya~\cite{fedllm-bench}.} A federated instruction tuning benchmark, based on Aya~\cite{singh2024aya}, where the data is annotated by contributors and partitioned by annotator ID. Following FedLLM-Bench~\cite{fedllm-bench}, we focus on 6 high-resource languages, English, Spanish, French, Russian, Portuguese, Chinese, and 2 low-resource languages, Arabic and Telugu. Additionally, we filter out the other languages from the dataset. Out of 38 clients, we select 8 for our \unseen{} pool, $\text{client\_ids}=[21, 22, 23, 24, 25, 26, 27, 34]$ and the rest goes into our \seen{} pool. Each client can have up to 4 languages where the number of data samples can range from a hundred to over a thousand samples per client.

\subsection{Models, Tokenizers, and Data Preprocessing}

\noindent\textbf{mBERT~\cite{BERT}.} We use the pretrained multilingual BERT with its WordPiece tokenizer for all sequence classification experiments, namely all XNLI and MasakhaNEWS setups with various {\em base models}. For both datasets, we use a training batch size of 32 and pad input tokens on the right to a max token length of 128 and 256 respectively.

\noindent\textbf{MobileLLaMA-1.4B~\cite{mobilellama}.} We train a \basemodel{} with a pretrained MobileLLaMA-1.4B with Standard FL using LoRA in our Fed-Aya setup. We use the default LLaMA tokenizer which is a BPE model based on sentencepiece~\cite{Kudo2018SentencePieceAS} and adopt the UNK token as the PAD token. During training, we use an effective batch size of $16$ and pad right to the longest token in the batch up to a max token length of 1024. For evaluation, we use a batch size of 8, padding left instead, with greedy sampling up to a max new token length of 1024. We use the Alpaca template to format each prompt:

\begin{lstlisting}[linewidth=\columnwidth,breaklines=true]
alpaca_template = """Below is an instruction that describes a task. Write a response that appropriately completes the request.

### Instruction:
{} 

### Response: {}{}"""
\end{lstlisting}

\noindent\textbf{Llama-3.2-3B~\cite{llama3}.} We use the off-the-shelf Llama-3.2-3B-Instruct model as our \basemodel{} and its default tokenizer which is a BPE model based on tiktoken\footnote{https://github.com/openai/tiktoken}. Training and evaluation hyperparameters are the same as the ones we use for MobileLLaMA. The only two differences are \textit{1)} we add a PAD token `\verb|<pad>|', and \textit{2)} we use the Llama 3 instruction template instead:

\begin{lstlisting}[linewidth=\columnwidth,breaklines=true]
llama3_instruct_template = """<|begin_of_text|><|start_header_id|>user<|end_header_id|>

{}<|eot_id|><|start_header_id|>assistant<|end_header_id|>

{}{}
"""
\end{lstlisting}

\subsection{Complementary Approaches and Base Models}

In this work, we experiment with different {\em base models} to show that \method{} is complementary to a range of off-the-shelf models and models trained using existing FL approaches. In this section, we detail the different approaches we used to obtain these {\em base models}.

\noindent\textbf{Standard FL.} Standard FL involves training a single global model. Given a pretrained LLM, we run FedAvg on the \seen{} pool of clients, where $10\%$ of participating clients are sampled every round to train the model before sending the weights back for aggregation. In our XNLI and MasakhaNEWS setup, we do full fine-tuning of mBERT, setting each client's learning rate to $5e-5$ and running FedAvg for 100 rounds. In our Fed-Aya setup, we adopt the training recipe from FedLLM-Bench~\cite{fedllm-bench} for MobileLLaMA-1.4B, where we do PEFT with LoRA applied to query and value attention weights ($r=16$, $\alpha_{lora}=32$, dropout$=0.05$) for 200 rounds. We use the cosine learning rate decay over 200 rounds with initial learning rate $2e^{-5}$ and minimum learning rate $1e^{-6}$.

\noindent\textbf{Personalized FL.} We train personalized {\em base models} using FedDPA-T~\cite{FedDPA} and DEPT(SPEC)~\cite{DEPT} in our XNLI setup. FedDPA-T proposed having two separate LoRA adapters, one of which is shared (global) and the other is kept locally for each client (local). 
We adopted their training recipe for sequence classification, where the classifier is shared together with the global LoRA modules and the local LoRA modules stay local. LoRA modules are only applied to query and value attention weights (r=8, $\alpha_{lora}=8$, dropout=$0.05$). We set the learning rate to $5e^{-5}$.

DEPT (SPEC), on the other hand, proposed having personalized token and positional embeddings for each client. As DEPT was proposed for cross-silo FL, while we target cross-device FL, they assumed that each data source has an abundance of data to retrain the newly initialized embeddings. Hence, to adapt to the cross-device FL setting, we did not reinitialize the embeddings; each client fine-tunes their own embeddings starting from the pretrained mBERT embeddings. In other words, for each FL round, each client does full fine-tuning, sending weights of all layers except their own embeddings back to the server for aggregation. As with FedDPA-T, the learning rate is set to $5e^{-5}$.

\noindent\textbf{Off-the-shelf.} In our Fed-Aya setup, we use an off-the-shelf instruction finetuned Llama-3.2-3B-Instruct as our \basemodel{}.

\subsection{Baselines and \method{}}

To avoid an exponentially big search space, all hyperparameter tuning is done using simple grid search on our XNLI setup, with mBERT, and Fed-Aya setup, with MobileLLaMA as the \basemodel{}. The best hyperparameters found are then used for MasakhaNEWS and Fed-Aya with Llama3 respectively.

\noindent\textbf{LoRA PEFT~\cite{hu2021lora}.} We search for the learning rate $[1e^{-5},1e^{-4},1e^{-3}]$ and the number of epochs $[1,2,3]$ and find that the learning rate $1e^{-4}$ with $2$ epochs had the best performance on the train set. We fixed $\alpha_{lora}=2r$. To ensure a similar inference budget across baselines, we set the number of epochs to $2$ for all our experiments.

\noindent\textbf{AdaLoRA~\cite{adalora}.} Similarly to LoRA, we search for the learning rate $[1e^{-5},1e^{-4},1e^{-3},1e^{-2}]$, the time interval between two budget allocations, $\Delta_T$, $[1,10,100]$ and the coefficient of the orthogonal regularization, $\gamma$, $[0.1,0.5]$. Within our search space, we find learning rate$=1e^{-3}$, $\Delta_T=1.0$, and $\gamma=0.1$ to be the best performing one. We run AdaLoRA once per resource budget $r$, setting the initial rank to be $1.5\times$ of $r$, as recommended. We set the initial fine-tuning warmup steps and final fine-tuning steps to be $10\%$ and $30\%$ of the total steps respectively. 

\noindent\textbf{BT-LoRA (Section\ref{sec:personalized_peft}).} For fair comparison with \method{}, we use the same hyperparameters as \method{}. This baseline, hence, is an ablation study of how much performance collaboratively learning a PSG adds.

\noindent\textbf{FedL2P~\cite{royson2023fedl2p}.} As FedL2P requires a validation set for outer-loop bi-level optimization and federated early stopping, we split the train set of every client $80\%$ train and $20\%$ validation. Following FedL2P, we set the federated early stopping patience to 50 rounds, MLP hidden dimension is set to 100, the inner-loop learning rate to be the same as finetuning, $1e^{-4}$, and the hypergradient hyperparameters, $Q=3, \psi=0.1$ with hypergradient clamping of $[-1,1]$. 
We use Adam for both the inner-loop and outer-loop optimizers and search for the learning rate for the MLP (LRNet) $[1e^{-5},1e^{-4},1e^{-3}]$ and the learnable post-multiplier learning rate $\tilde{\eta}$ $[1e^{-4},1e^{-5},1e^{-6}]$, picking $1e^{-4}$ and $1e^{-6}$ to be the best respectively. Finally, we use $3$ outer-loop steps with an effective outer-loop batch size of $16$.

\noindent\textbf{\method{} (Section~\ref{sec:main_method}).} We set $\alpha_{r\_mul}=2$, our resulting $r_{init}$ is, hence, $32$ since our $r_{\text{max target}}=16$ for all experiments. Following our standard LoRA fine-tuning baseline, we adopt the same learning rate and $\alpha_{lora}$, $1e-4$ and $2r_{init}$ respectively. The learning rate of $\bm{\lambda}$, on the other hand, is searched $[1e^{-1},1e^{-2},1e^{-3},1e^{-4}]$, and we pick $1e^{-2}$ for all experiments.
All $\lambda$ values are initialized to $1e^{-4}$. The MLP hidden dimension is set to $2\times$ the input dimension, which is model dependent. We clamp the output of the MLP as well as $\lambda$ with a minimum value of $1e-4$ in the forward pass during training. Following FedL2P, we use a straight-through estimator after clamping to propagate gradients. We initialize the weights of MLP with Xavier initialization using the normal distribution with a gain value of $1e-6$ and the bias with a constant $1e-4$. Lastly, we set $\alpha_s=1e^{+2}$ and $\alpha_p=1e^{-2}$ for all experiments.

\subsection{Experiments Compute Resource}~\label{app:compute_res}
\noindent We used Flower~\cite{beutel2020flower}\footnote{https://github.com/adap/flower}, which uses Ray~\cite{moritz2018ray}\footnote{https://github.com/ray-project/ray} for distributed training and inference, for all our experiments.
Each experiment in our XNLI and MasakhaNEWS setup was run with 4 to 6 NVIDIA A40s GPUs, with each client running on 1 GPU at a time. Each \method{} federated training run took less than an hour in total while FedL2P training run took a few hours for each rank budget. As accuracy is cheap to compute, each inference run only took a few minutes to evaluate on all clients. Our Fed-Aya setup with Llama-3.2-3B, on the other hand, took under a day to federatedly train \method{} and 1-3 days to train FedL2P for each rank budget on 8 NVIDIA A100s 80GB. Due to the higher cost of auto-regressive generative runs compared to classification, each inference run took roughly 1-3 hours.

\section{LoRA Rank Cost and Adaptability to Heterogeneous Devices}\label{app:cost_device}

\noindent\textbf{LoRA Rank Cost.} In our experiments, we consider LoRA rank budgets of $r=2,4,8,16$ where the total rank budget is $r \cdot L$. Our empirical results on an NVIDIA A40 show a relatively small increase of 0.3GB in memory and 2.4 seconds in runtime between r=2 and r=16. This seemingly minor overhead can have major practical implications on devices with limited resources, such as laptops. Specifically, a 0.3GB memory footprint can be prohibitive on GPUs like QC Adreno 618 with typically only around 4GB of VRAM. We also predict a significant latency increase on mobile devices, as they lack the processing power and specialized hardware of larger systems and commonly are forced to offload parameters from RAM to the excessively slow-access flash memory, underscoring the need for efficient LoRA rank selection in real-world deployments. In scenarios where resources become more abundant, the server can host multiple models of different sizes and simply switch to a larger \basemodel{} instead of using a higher rank. Nonetheless, we emphasize that our focus is to generate personalized fine-tuning hyperparameters for any \basemodel{} chosen by the user.

\noindent\textbf{Heterogeneous Devices Compatibility.} During federated training, we assume that all clients that partake in training (\seen{} clients) have sufficient resources to train the full model and the max LoRA rank. Clients with fewer resources are assumed to be in the \unseen{} pool. Extending the work to train the PSG over heterogeneous resources is interesting as it allows clients with fewer resources to partake in the training stage. However, it is out of scope and we leave it for future work. During inference, we show that our approach mostly outperforms existing baselines with varying LoRA rank budgets across different model sizes.

\section{On BayesTune}\label{app:bayestune}
\noindent BayesTune~\cite{kim2023bayestune} is a sparse fine-tuning approach that leverages a Bayesian framework to identify and update only the most important parameters of the model. It introduces a per-parameter scaling coefficient, where the magnitude of this coefficient governs the degree to which the corresponding parameter is updated during fine-tuning. These scaling coefficients, along with the model parameters, are explored and sampled from their posterior distribution using SGLD. To encourage sparsity, the Laplace and Gamma priors are imposed on the model parameters and their scaling coefficients respectively as discussed in Section~\ref{sec:personalized_peft}.

\noindent\textbf{Motivation for Adapting BayesTune.} We adapt BayesTune for 2 reasons: 1) Most existing LoRA rank selection methods require separate runs for each rank resource budget while adapting BayesTune enables us to train only once for all targeted rank budgets. This is particularly useful as we only need to run the costly process of federated training once per base model. 2) The original proposed parameter-wise BayesTune is shown to be highly effective across a wide range of tasks, outperforming many existing pruning approaches and PEFT methods.

However, as mentioned in Section~\ref{sec:personalized_peft}, directly using BayesTune in our approach will not only incur an intractable cost but also be difficult to optimize as the output dimension of our MLP will be too big and the likelihood of getting stuck in a suboptimal local minimum is high. This limitation can also be observed in existing FL hypernetwork approaches~\cite{pfedhn} where the experiments were constrained to learning the personalized parameters of a tiny LeNet-5. Similarly, the authors of FedL2P~\cite{royson2023fedl2p} found that extending these hypernetwork approaches to ResNet-18 is costly and often struggle to converge. To make federated hyperparameter optimization feasible, we scale down this output space by learning optimal LoRA rank coefficients instead. While doing so improves convergence stability, the output dimension scales linearly with the number of LoRA ranks, thereby also susceptible to the aforementioned limitation at higher ranks.


Besides limiting the output dimensionality through learning LoRA rank coefficients, we also opt to find a point estimate rather than a full characterization of the posterior distribution due to the limited number of data samples often found in cross-device FL. Doing so allows us to reduce the number of hyperparameters, namely SGLD's noise schedule and level, to tune and simplify the optimization process, focusing valuable on-device computational resources on effectively learning the client-specific personalization strategy with the available limited data.

\noindent\textbf{Choice of $\alpha_g$ and $\beta_g$.} Following BayesTune, we aim to yield sparse models by putting high density on small $\lambda$ close to 0, penalizing $\lambda$ that deviates from 0. To achieve this, the Gamma prior imposed on $\lambda$ needs to meet several constraints: 1) it has mode close to 0 ($\alpha_g < 1$  for mode=0) and, 2) it decays rapidly as $\lambda$ grows (mean of Gamma, $\frac{\alpha_g}{\beta_g} \approx 0$ and variance of Gamma, $\frac{\alpha_g}{\beta_g^2} \approx 0$). The choice $\alpha_g=0.01$ and $\beta_g=100$ is good enough to meet these constraints. In the original BayesTune work, the authors found there isn’t substantial difference in the final performance as long as $\alpha_g$, and $\beta_g$ satisfies the above constraints.

\clearpage
\section{\method{} Algorithms}\label{app:algo}

Complementing the federated training and inference pipeline (Figure~\ref{fig:approach} and \ref{fig:intro}, respectively), this section details Algorithm~\ref{algo:train} (federated training) and Algorithm~\ref{algo:inference} (inference).

\begin{figure}[H]
\begin{minipage}[t]{\textwidth}
\begin{algorithm}[H]
    \small
    \caption{\small{\method: Federated Training}} \label{algo:train}
    \algsetup{indent=0.5em}
    \textbf{Input:} \footnotesize{A \basemodel{} $\Phi^i_{BM}$ with $L$ layers for each client $i$. }
    \begin{algorithmic}[1]
        \STATE initialize MLP $\phi$
        \FOR{each federated round}
            \STATE Random sample clients $\tilde{C}$
            \FOR{client $i \in \tilde{C}$}
                \STATE send $\phi$ to client
                \STATE Forward pass of local dataset to compute $E(h_l),SD(h_l)$ for $1 \leq l \leq L-1$
                \STATE Compute $\bm{\hat{\lambda}^{i,0}}$ using $\phi$ (Eq.~\ref{eq:mlp})
                \STATE Initialize BayesTune-LoRA: Randomly initialize LoRA modules $\bm{A^i}\bm{B^i}$ and plug $\bm{\hat{\lambda}^{i,0}}$ in them
                \STATE $\bm{\hat{\lambda}^{i,s}} \leftarrow$ Finetune $\bm{A^i},\bm{\hat{\lambda}^{i,0}},\bm{B^i}$ for $s$ steps, minimizing the losses in Eq.~\ref{eq:bayestune}
                \STATE $\phi^i \leftarrow$ Finetune $\phi$, minimizing L1 loss $\mathcal{L}_1(\bm{\hat{\lambda}^{i,0}}, \bm{\hat{\lambda}^{i,s}})$
                \STATE send $\phi^i$ and num of data samples $N$ to server
            \ENDFOR
            \STATE $\phi \leftarrow \sum^{\tilde{C}}_{i} \frac{N_i}{\sum N_i} \phi^i$ 
        \ENDFOR
    \end{algorithmic}
    \textbf{Output:} $\phi$  
\end{algorithm}
\vspace{-2em}
\end{minipage}
\begin{minipage}[t]{\textwidth}
\begin{algorithm}[H]
    \small
    \caption{\small{\method: Inference}} \label{algo:inference}
    \algsetup{indent=0.5em}
    \vspace{-0.4em}
    \begin{flushleft}
    \textbf{Input:} \footnotesize{A \basemodel{} $\Phi^i_{BM}$ with $L$ layers for each client $i$. Learned MLP $\phi$}.
    \end{flushleft}
    \begin{algorithmic}[1]
        \FOR{each client $i$}
            \STATE send $\phi$ to client
            \STATE Forward pass of local dataset to compute $E(h_l),SD(h_l)$ for $1 \leq l \leq L-1$
            \STATE Compute $\bm{\hat{\lambda}^{i}}$ using $\phi$ (Eq.~\ref{eq:mlp})
            \STATE $\bm{\hat{\lambda}^{i}} \leftarrow$ Take top-($r \cdot L$) largest ranks of $\bm{\hat{\lambda}^{i}}$ and freeze them given total rank budget $r \cdot L$ 
            \STATE Initialize BayesTune-LoRA for all layers $||\lambda_l||_1 > 0$
            \STATE Finetune BayesTune-LoRA layers $\bm{A^i},\bm{B^i}$ with frozen $\bm{\hat{\lambda}^{i}}$ and obtain personalized model $\Phi^i$
        \ENDFOR
    \end{algorithmic}
    
    \textbf{Output:} $\Phi^i$
\end{algorithm}
\end{minipage}
\end{figure}

\clearpage
\section{Additional Results}\label{app:additional_results}

This section contains supplementary results and analyses, omitted from the main paper due to space limitations, that complement the presented findings.  

\noindent\textbf{Popularity Bias.} In our work, we simply adopt FedAvg for aggregation. A potential known limitation of FedAvg-based approaches is popularity bias, where clients with larger datasets may disproportionately influence the model. To verify this, we measure the Pearson correlation coefficient between the number of train samples of each client and their personalized accuracy in our MasakhaNEWS setup and got a weak correlation of 0.23. Although quantity skew can bias clients with larger datasets, we also observe that some languages are “easier to learn” than others, hence requiring fewer samples and resulting in the observed weak correlation between dataset size and performance.

\noindent\textbf{Predominant Language Bias.} A limitation of \method{}, as discussed in the main paper, is that $\bm{\lambda}$ is heavily influenced by the client's predominant language. To quantify this bias, we correlated each client's predominant language proportion with the disparity in performance gain between predominant and minority languages in our Llama-3 3B setup, finding moderate positive Pearson coefficients of 0.6 (ROUGE-1), 0.56 (ROUGE-L), and 0.52 (METEOR). Potential future directions to address this include extracting per-language (rather than per-client) metadata as input to the PSG, or exploring multi-adapter approaches where each language receives a dedicated adapter.

\begin{table*}[h]
\begin{tabular}{c|ccccc|ccccc}
\toprule
\textbf{$\mathbf{r}$} & \multicolumn{5}{c|}{2}                                                                                                                                                                                & \multicolumn{5}{c}{4}                                                                                                                                                                                    \\ \midrule
\textbf{Lang.} & \multicolumn{1}{c|}{\textbf{LoRA}}     & \multicolumn{1}{c|}{\textbf{AdaLoRA}} & \multicolumn{1}{c|}{\textbf{BT-LoRA}} & \multicolumn{1}{c|}{\textbf{FedL2P}}   & \textbf{\method{}} & \multicolumn{1}{c|}{\textbf{LoRA}}     & \multicolumn{1}{c|}{\textbf{AdaLoRA}} & \multicolumn{1}{c|}{\textbf{BT-LoRA}} & \multicolumn{1}{c|}{\textbf{FedL2P}}   & \textbf{\method{}} \\ \hline
eng                   & \multicolumn{1}{c|}{90.4±0.1}      & \multicolumn{1}{c|}{89.9±0.0}         & \multicolumn{1}{c|}{89.9±0.0}                & \multicolumn{1}{c|}{90.7±0.6}          & \textbf{92.0±0.0}                & \multicolumn{1}{c|}{91.3±0.0}          & \multicolumn{1}{c|}{89.9±0.0}         & \multicolumn{1}{c|}{89.9±0.0}                & \multicolumn{1}{c|}{91.7±0.4}          & \textbf{92.0±0.4}                \\ \hline
som                   & \multicolumn{1}{c|}{60.1±0.3}      & \multicolumn{1}{c|}{59.4±0.3}         & \multicolumn{1}{c|}{59.6±0.3}                & \multicolumn{1}{c|}{61.0±1.2}          & \textbf{65.5±1.2}                & \multicolumn{1}{c|}{61.9±0.0}          & \multicolumn{1}{c|}{59.6±0.3}         & \multicolumn{1}{c|}{59.4±0.3}                & \multicolumn{1}{c|}{63.5±1.2}          & \textbf{67.1±0.6}                \\
\bottomrule
\end{tabular}
\caption{Full Mean±SD Accuracy of each language across 3 different seeds for \seen{} clients of our MasakhaNEWS setup with ranks \{2, 4, 8, 16\}. The pretrained model is trained using \textbf{Standard FL with full fine-tuning} and the resulting \basemodel{} is personalized to each client given a baseline approach.}
\label{tab:masakha_seen_full}
\end{table*}
\begin{table*}[t]
\centering

\begin{tabular}{c|ccccc|ccccc}
\toprule
$\mathbf{r}$   & \multicolumn{5}{c|}{2}                                                                                                                                                                                   & \multicolumn{5}{c}{4}                                                                                                                                                                                   \\ \midrule
\textbf{Lang.} & \multicolumn{1}{c|}{\textbf{LoRA}}     & \multicolumn{1}{c|}{\textbf{AdaLoRA}} & \multicolumn{1}{c|}{\textbf{BT-LoRA}} & \multicolumn{1}{c|}{\textbf{FedL2P}}   & \textbf{\method{}} & \multicolumn{1}{c|}{\textbf{LoRA}}     & \multicolumn{1}{c|}{\textbf{AdaLoRA}} & \multicolumn{1}{c|}{\textbf{BT-LoRA}} & \multicolumn{1}{c|}{\textbf{FedL2P}}   & \textbf{\method{}} \\ \hline
eng            & \multicolumn{1}{c|}{\textbf{90.7±0.0}} & \multicolumn{1}{c|}{90.3±0.0}         & \multicolumn{1}{c|}{90.3±0.0}                & \multicolumn{1}{c|}{90.6±0.1}          & 90.7±0.1                        & \multicolumn{1}{c|}{90.8±0.1}          & \multicolumn{1}{c|}{90.2±0.1}         & \multicolumn{1}{c|}{90.2±0.1}                & \multicolumn{1}{c|}{90.8±0.1}          & \textbf{90.9±0.4}               \\ \hline
som            & \multicolumn{1}{c|}{68.5±0.3}          & \multicolumn{1}{c|}{67.3±0.0}         & \multicolumn{1}{c|}{67.3±0.0}                & \multicolumn{1}{c|}{68.5±0.3}          & \textbf{72.1±0.6}               & \multicolumn{1}{c|}{68.7±0.0}          & \multicolumn{1}{c|}{67.3±0.0}         & \multicolumn{1}{c|}{67.1±0.3}                & \multicolumn{1}{c|}{68.4±0.3}          & \textbf{72.6±0.3}               \\
\bottomrule
\end{tabular}

\caption{Mean±SD Accuracy of each language for \unseen{} clients of our MasakhaNEWS setup. The pretrained model is trained using \textbf{Standard FL with full fine-tuning} and the resulting \basemodel{} is personalized to each client given a baseline approach.}
\label{tab:masakha_unseen_full}
\end{table*}

\caption{Mean±SD Accuracy of each language across 3 different seeds for clients in the \seen{} pool of our XNLI setup. The pretrained model is trained using \textbf{FedDPA-T} and the resulting \basemodel{} is personalized to each client given a baseline approach.}
\label{tab:xnli_seen_feddpa_full}
\end{table*}
\begin{table*}[t]
\centering

\begin{tabular}{c|ccccc}
\toprule
$\mathbf{r}$      & \multicolumn{5}{c}{2}                                                                                                                                                                                                                                                                                                                                       \\ \midrule
\textbf{Lang.} & \multicolumn{1}{c|}{\textbf{LoRA}}                            & \multicolumn{1}{c|}{\textbf{AdaLoRA}}                                                   & \multicolumn{1}{c|}{\textbf{BT-LoRA}}                  & \multicolumn{1}{c|}{\textbf{FedL2P}}                          & \textbf{\method{}}                                    \\ \hline
te                & \multicolumn{1}{c|}{\textbf{0.24}/\textbf{0.14}/0.13}                           & \multicolumn{1}{c|}{\textbf{0.24/0.14/0.14}}                                                     & \multicolumn{1}{c|}{\textbf{0.24}/\textbf{0.14/0.14}} & \multicolumn{1}{c|}{0.23/\textbf{0.14}/0.13}                           & \textbf{0.24}/\textbf{0.14/0.14}                           \\ \hline
ar                & \multicolumn{1}{c|}{0.34/0.07/0.07}                           & \multicolumn{1}{c|}{0.34/0.07/0.07}                                                     & \multicolumn{1}{c|}{0.32/0.06/0.06}                           & \multicolumn{1}{c|}{0.33/0.07/0.07}                           & \textbf{0.37/0.08/0.08}                                            \\ \hline
es                & \multicolumn{1}{c|}{\textbf{0.39}/0.41/0.39}                           & \multicolumn{1}{c|}{\textbf{0.39/0.43/0.40}}                                            & \multicolumn{1}{c|}{0.38/0.41/0.38}                           & \multicolumn{1}{c|}{0.38/0.40/0.38}                           & \textbf{0.39}/0.42/0.39                                                     \\
\bottomrule
\end{tabular}

\caption{Avg. METEOR/ROUGE-1/ROUGE-L for \seen{} clients in our Fed-Aya setup. {\em Base model} is off-the-shelf Llama-3.2-3B-Instruct.}
\label{tab:lama_fedaya_seen_full}
\end{table*}
\begin{table*}[t]
\centering

\begin{tabular}{c|ccccc}
\toprule
$\mathbf{r}$  & \multicolumn{5}{c}{2}                                                                                                                                                                                                                                                                                   \\ \midrule
\textbf{Lang.} & \multicolumn{1}{c|}{\textbf{LoRA}}                            & \multicolumn{1}{c|}{\textbf{AdaLoRA}}                         & \multicolumn{1}{c|}{\textbf{BT-LoRA}}                         & \multicolumn{1}{c|}{\textbf{FedL2P}}                          & \textbf{\method{}}          \\ \hline
te            & \multicolumn{1}{c|}{0.16/0.09/0.08}                           & \multicolumn{1}{c|}{0.16/0.11/\textbf{0.11}}                           & \multicolumn{1}{c|}{\textbf{0.17}/0.09/0.09} & \multicolumn{1}{c|}{0.16/\textbf{0.12/0.11}} & \textbf{0.17}/0.07/0.07                           \\ \hline
ar            & \multicolumn{1}{c|}{0.24/0.05/0.04}                           & \multicolumn{1}{c|}{0.23/\textbf{0.07/0.07}} & \multicolumn{1}{c|}{0.22/0.05/0.05}                           & \multicolumn{1}{c|}{0.24/0.05/0.05}                           & \textbf{0.26}/0.05/0.05 \\ \hline
es            & \multicolumn{1}{c|}{0.43/0.49/0.44}                           & \multicolumn{1}{c|}{0.43/0.50/0.44}                           & \multicolumn{1}{c|}{0.38/0.46/0.41}                           & \multicolumn{1}{c|}{0.41/0.48/0.43}                           & \textbf{0.44/0.53/0.48}                  \\ \hline
en            & \multicolumn{1}{c|}{\textbf{0.32}/0.31/0.25} & \multicolumn{1}{c|}{\textbf{0.32}/0.31/0.25}                           & \multicolumn{1}{c|}{\textbf{0.32}/0.31/0.24}                           & \multicolumn{1}{c|}{0.31/0.30/0.24}                           & 0.31/\textbf{0.33/0.27} \\
\bottomrule
\end{tabular}

\caption{Avg. METEOR/ROUGE-1/ROUGE-L for \unseen{} clients in our Fed-Aya setup. {\em Base model} is off-the-shelf Llama-3.2-3B-Instruct.}
\label{tab:lama_fedaya_unseen_full}
\end{table*}
\begin{table*}[t]
\centering

\begin{tabular}{c|cccc}
\toprule
$\mathbf{r}$   & \multicolumn{4}{c}{2}                                                                                                                                    \\ \midrule
\textbf{Lang.} & \multicolumn{1}{c|}{\textbf{LoRA}}  & \multicolumn{1}{c|}{\textbf{AdaLoRA}} & \multicolumn{1}{c|}{\textbf{BT-LoRA}} & \textbf{\method{}} \\ \hline
te             & \multicolumn{1}{c|}{0.24/0.15/0.14} & \multicolumn{1}{c|}{0.24/0.15/\textbf{0.15}}   & \multicolumn{1}{c|}{0.24/0.14/0.14}   & \textbf{0.25/0.16/0.15}                      \\ \hline
ar             & \multicolumn{1}{c|}{\textbf{0.34/0.07/0.07}} & \multicolumn{1}{c|}{0.33/\textbf{0.07/0.07}}   & \multicolumn{1}{c|}{0.32/0.06/0.06}   & \textbf{0.34/0.07/0.07}                      \\ \hline
es             & \multicolumn{1}{c|}{0.38/0.42/0.39} & \multicolumn{1}{c|}{0.39/0.42/\textbf{0.40}}   & \multicolumn{1}{c|}{0.37/0.41/0.38}   & \textbf{0.40/0.43/0.40 }                     \\ \hline
en             & \multicolumn{1}{c|}{0.36/0.41/0.34} & \multicolumn{1}{c|}{0.37/0.43/0.35}   & \multicolumn{1}{c|}{0.32/0.33/0.27}   & \textbf{0.40/0.45/0.38 }                     \\ \hline
fr             & \multicolumn{1}{c|}{0.30/0.37/0.31} & \multicolumn{1}{c|}{0.32/0.39/0.32}   & \multicolumn{1}{c|}{0.29/0.36/0.30}   & \textbf{0.33/0.40/0.33}                      \\ \hline
zh             & \multicolumn{1}{c|}{0.10/\textbf{0.13/0.13}} & \multicolumn{1}{c|}{\textbf{0.11/0.13/0.13}}   & \multicolumn{1}{c|}{0.10/\textbf{0.13}/0.12}   & 0.10\textbf{/0.13}/0.12                      \\
\bottomrule
\end{tabular}

\caption{Avg. METEOR/ROUGE-1/ROUGE-L for \seen{} clients in our Fed-Aya setup. {\em Base model} is off-the-shelf Llama-3.1-8B-Instruct. FedL2P leads to out-of-memory exceptions due to its expensive second-order optimization.}
\label{tab:lama_fedaya_seen_full_8B}
\end{table*}
\begin{table*}[t]
\centering

\begin{tabular}{c|cccc}
\toprule
$\mathbf{r}$   & \multicolumn{4}{c}{2}                                                                                                                                     \\ \midrule
\textbf{Lang.} & \multicolumn{1}{c|}{\textbf{LoRA}}  & \multicolumn{1}{c|}{\textbf{AdaLoRA}} & \multicolumn{1}{c|}{\textbf{BT-LoRA}} & \textbf{\method{}} \\ \hline
te             & \multicolumn{1}{c|}{\textbf{0.17/0.12}/0.11} & \multicolumn{1}{c|}{\textbf{0.17/0.12}/0.11}   & \multicolumn{1}{c|}{0.15/\textbf{0.12/0.12}}   & 0.16/0.11/0.10                      \\ \hline
ar             & \multicolumn{1}{c|}{0.25/0.03/0.03} & \multicolumn{1}{c|}{\textbf{0.27}/0.05/0.05}   & \multicolumn{1}{c|}{0.23/0.05/0.05}   & 0.26/\textbf{0.06/0.06}                      \\ \hline
es             & \multicolumn{1}{c|}{0.43/0.53/0.48} & \multicolumn{1}{c|}{0.42/0.53/0.48}   & \multicolumn{1}{c|}{0.39/0.52/0.47}   & \textbf{0.45/0.57/0.51}                      \\ \hline
en             & \multicolumn{1}{c|}{\textbf{0.33}/0.31/0.25} & \multicolumn{1}{c|}{0.32/0.31/0.25}   & \multicolumn{1}{c|}{0.31/0.29/0.23}   & \textbf{0.33/0.35/0.29}                      \\ \hline
fr             & \multicolumn{1}{c|}{0.44/0.50/0.50} & \multicolumn{1}{c|}{0.21/0.56/0.56}   & \multicolumn{1}{c|}{\textbf{0.53/0.61/0.61}}   & \textbf{0.53}/0.60/0.60                      \\ \hline
zh             & \multicolumn{1}{c|}{0.22/0.00/0.00} & \multicolumn{1}{c|}{\textbf{0.24/0.01/0.01}}   & \multicolumn{1}{c|}{0.23/0.00/0.00}   & \textbf{0.24}/0.00/0.00                      \\
\bottomrule
\end{tabular}

\caption{Avg. METEOR/ROUGE-1/ROUGE-L for \unseen{} clients in our Fed-Aya setup. {\em Base model} is off-the-shelf Llama-3.1-8B-Instruct. FedL2P leads to out-of-memory exceptions due to its expensive second-order optimization.}
\label{tab:lama_fedaya_unseen_full_8B}
\end{table*}

\begin{figure*}
\centering
\begin{minipage}{.23\linewidth}
  \includegraphics[width=\linewidth]{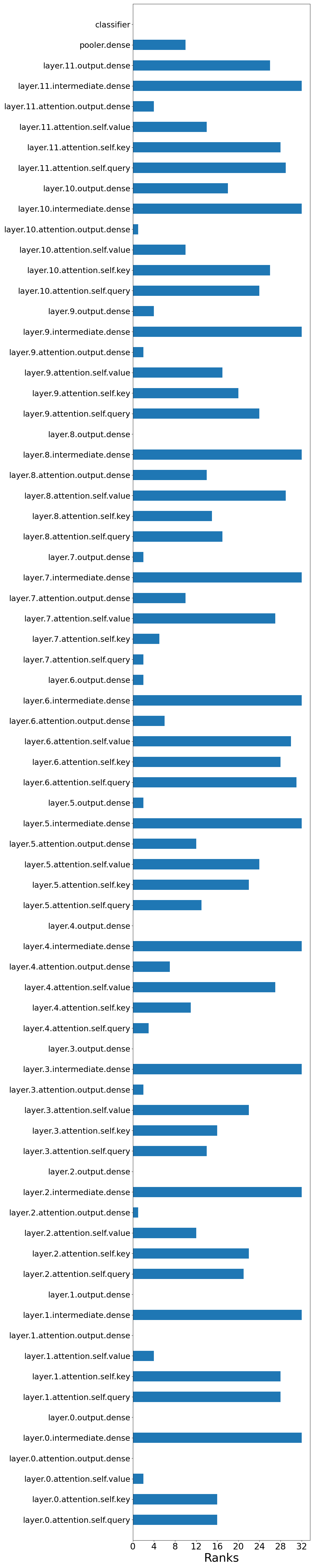}
  \captionof{figure}{Language agnostic rank structure of mBERT in our XNLI setup where the \basemodel{} is trained with Standard FL full-finetuning ($r=16$).}
  \label{fig:xnli_fedavg_out_r16}
\end{minipage}
\hspace{.01\linewidth}
\begin{minipage}{.23\linewidth}
  \includegraphics[width=\linewidth]{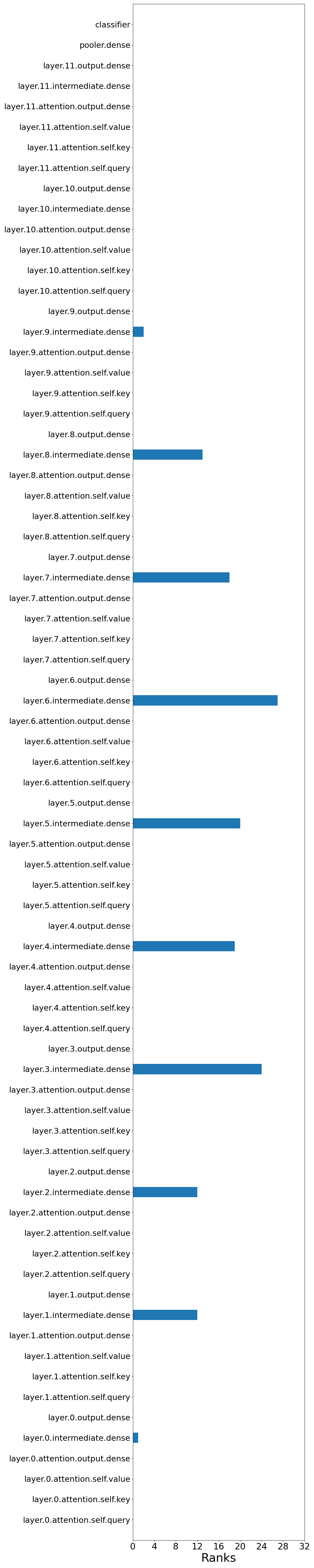}
  \captionof{figure}{Language agnostic rank structure of mBERT in our XNLI setup where the \basemodel{} is trained with Standard FL full-finetuning ($r=2$).}
  \label{fig:xnli_fedavg_out_r2}
\end{minipage}
\hspace{.01\linewidth}
\begin{minipage}{.23\linewidth}
  \includegraphics[width=\linewidth]{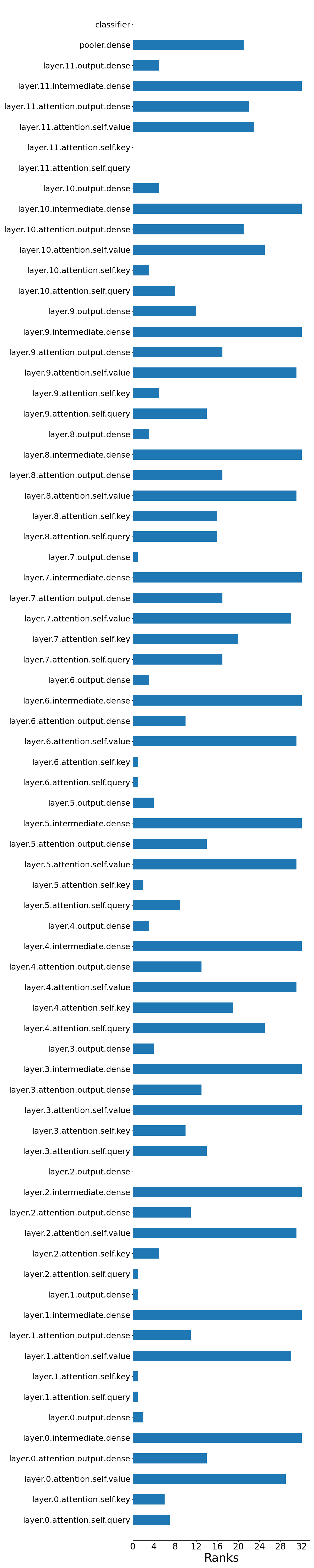}
  \captionof{figure}{Language agnostic rank structure of mBERT in our XNLI setup where the \basemodel{} is trained with DEPT(SPEC) ($r=16$).}
  \label{fig:xnli_dept_out_r16}
\end{minipage}
\hspace{.01\linewidth}
\begin{minipage}{.23\linewidth}
  \includegraphics[width=\linewidth]{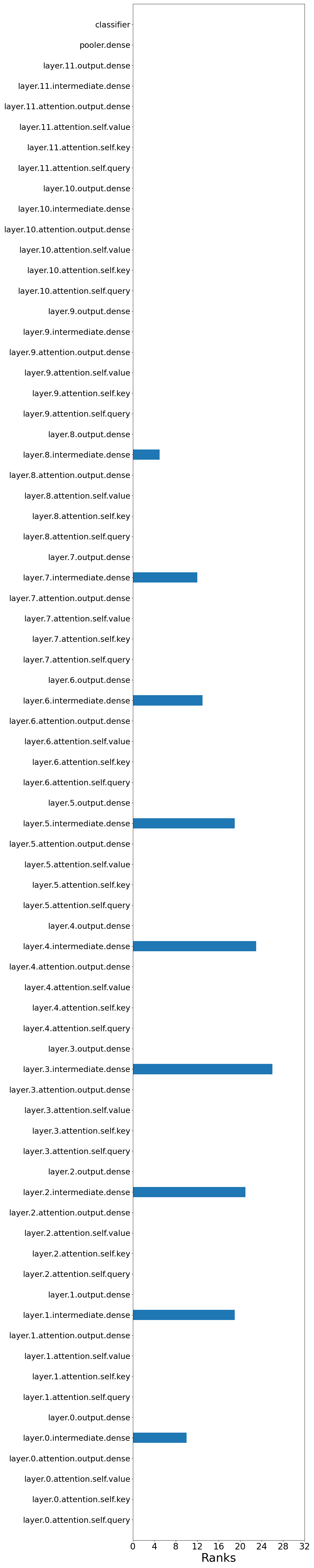}
  \captionof{figure}{Language agnostic rank structure of mBERT in our XNLI setup where the \basemodel{} is trained with DEPT(SPEC) ($r=2$).}
  \label{fig:xnli_dept_out_r2}
\end{minipage}
\end{figure*}

\begin{figure*}
\centering
\begin{minipage}{.24\linewidth}
  \includegraphics[width=\linewidth]{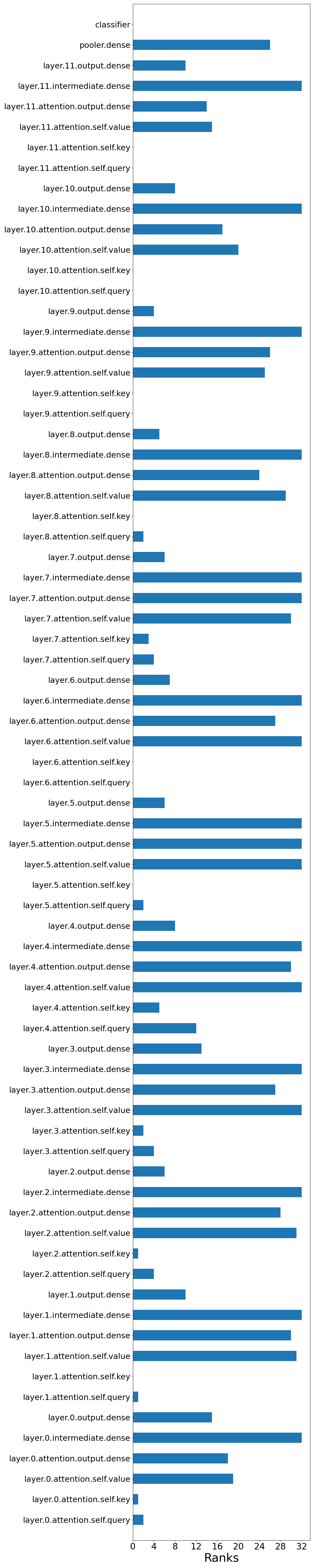}
  \captionof{figure}{Language agnostic rank structure of mBERT in our MasakhaNEWS setup where the \basemodel{} is trained with Standard FL full-finetuning ($r=16$).}
  \label{fig:masakha_fedavg_out_r16}
\end{minipage}
\hspace{.2\linewidth}
\begin{minipage}{.24\linewidth}
  \includegraphics[width=\linewidth]{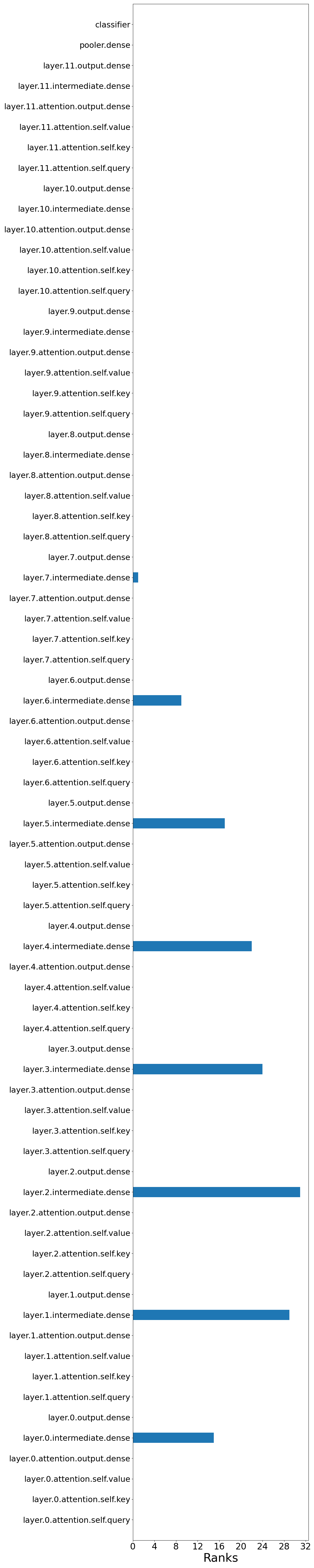}
  \captionof{figure}{Language agnostic rank structure of mBERT in our MasakhaNEWS setup where the \basemodel{} is trained with Standard FL full-finetuning ($r=2$).}
  \label{fig:masakha_fedavg_out_r2}
\end{minipage}
\end{figure*}

\begin{figure*}
\centering
\begin{minipage}{.23\linewidth}
  \includegraphics[width=\linewidth]{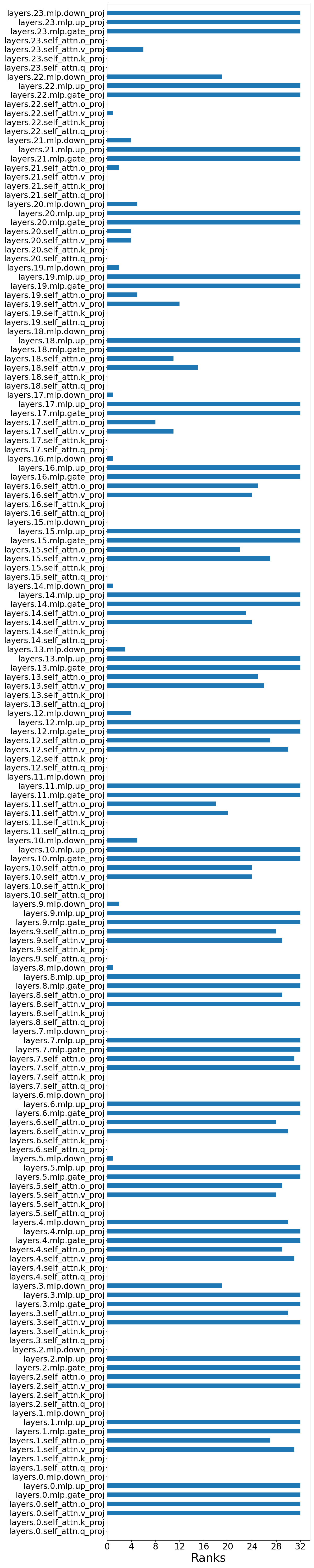}
  \captionof{figure}{Language agnostic rank structure of MobileLLaMA-1.4B in our Fed-Aya setup where the \basemodel{} is trained with Standard FL LoRA ($r=16$). Zoom in for best results.}
  \label{fig:mobilellama_fedavg_out_r16}
\end{minipage}
\hspace{.01\linewidth}
\begin{minipage}{.23\linewidth}
  \includegraphics[width=\linewidth]{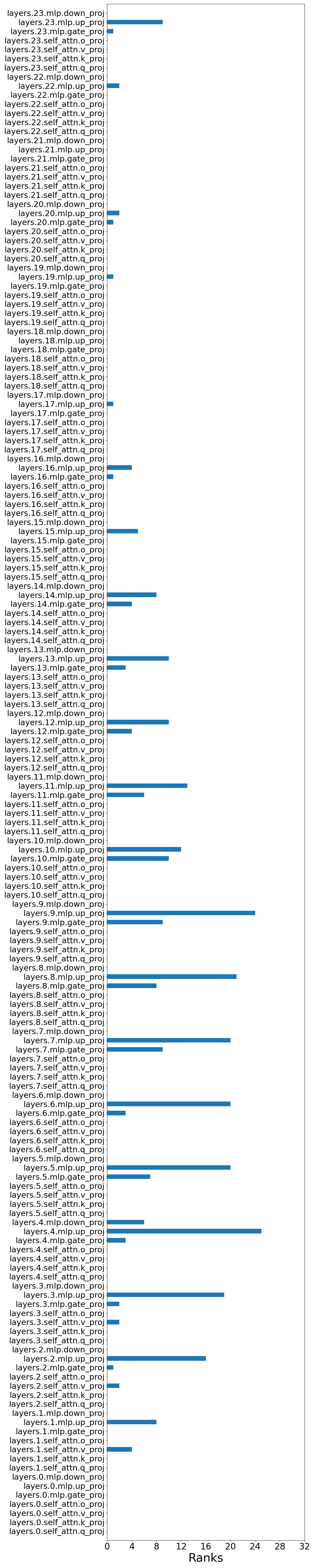}
  \captionof{figure}{Language agnostic rank structure of MobileLLaMA-1.4B in our Fed-Aya setup where the \basemodel{} is trained with Standard FL LoRA ($r=2$). Zoom in for best results.}
  \label{fig:mobilellama_fedavg_out_r2}
\end{minipage}
\hspace{.01\linewidth}
\begin{minipage}{.23\linewidth}
  \includegraphics[width=\linewidth]{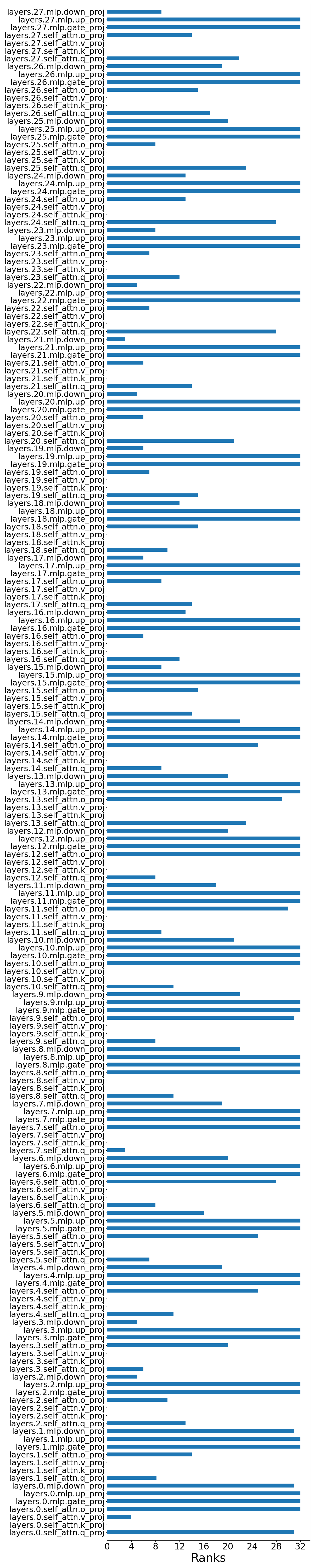}
  \captionof{figure}{Language agnostic rank structure of Llama-3.2-3B in our Fed-Aya setup where the \basemodel{} is an off-the-shelf instruction tuned Llama-3.2-3B-Instruct ($r=16$). Zoom in for best results.}
  \label{fig:llama3_fedavg_out_r16}
\end{minipage}
\hspace{.01\linewidth}
\begin{minipage}{.23\linewidth}
  \includegraphics[width=\linewidth]{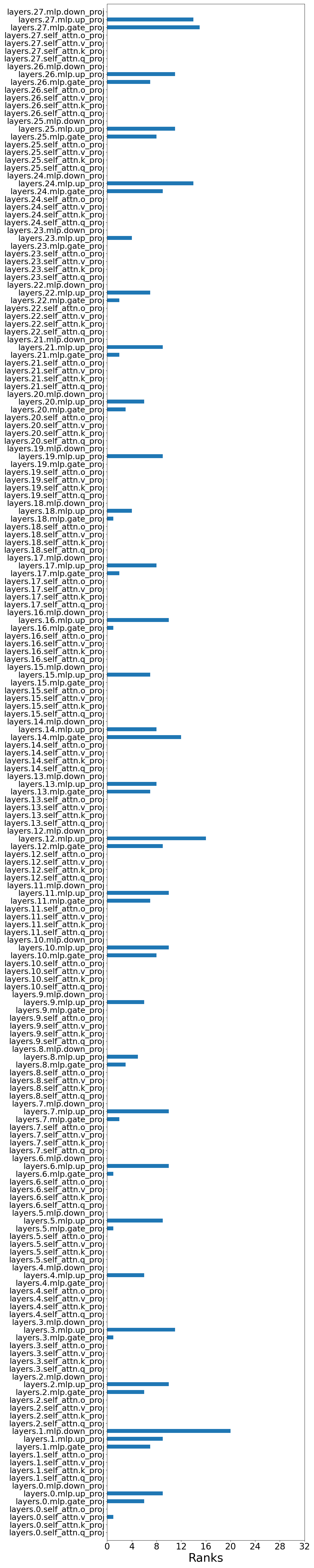}
  \captionof{figure}{Language agnostic rank structure of Llama-3.2-3B in our Fed-Aya setup where the \basemodel{} is an off-the-shelf instruction tuned Llama-3.2-3B-Instruct ($r=2$). Zoom in for best results.}
  \label{fig:llama3_fedavg_out_r2}
\end{minipage}
\end{figure*}

\noindent\textbf{Learned Rank Structures.} Fig.~\ref{fig:xnli_dept_out_r16}-\ref{fig:llama3_fedavg_out_r2} show the language-agnostic rank structures under different budgets ($r=2$ and $r=16$) learnt by \method{} for different setups as mentioned in Section~\ref{sec:analysis}. These plots illustrate the prioritization of layers for LoRA fine-tuning. 

\begin{figure*}[t]
    \small
    \centering
    \includegraphics[width=1.0\columnwidth]{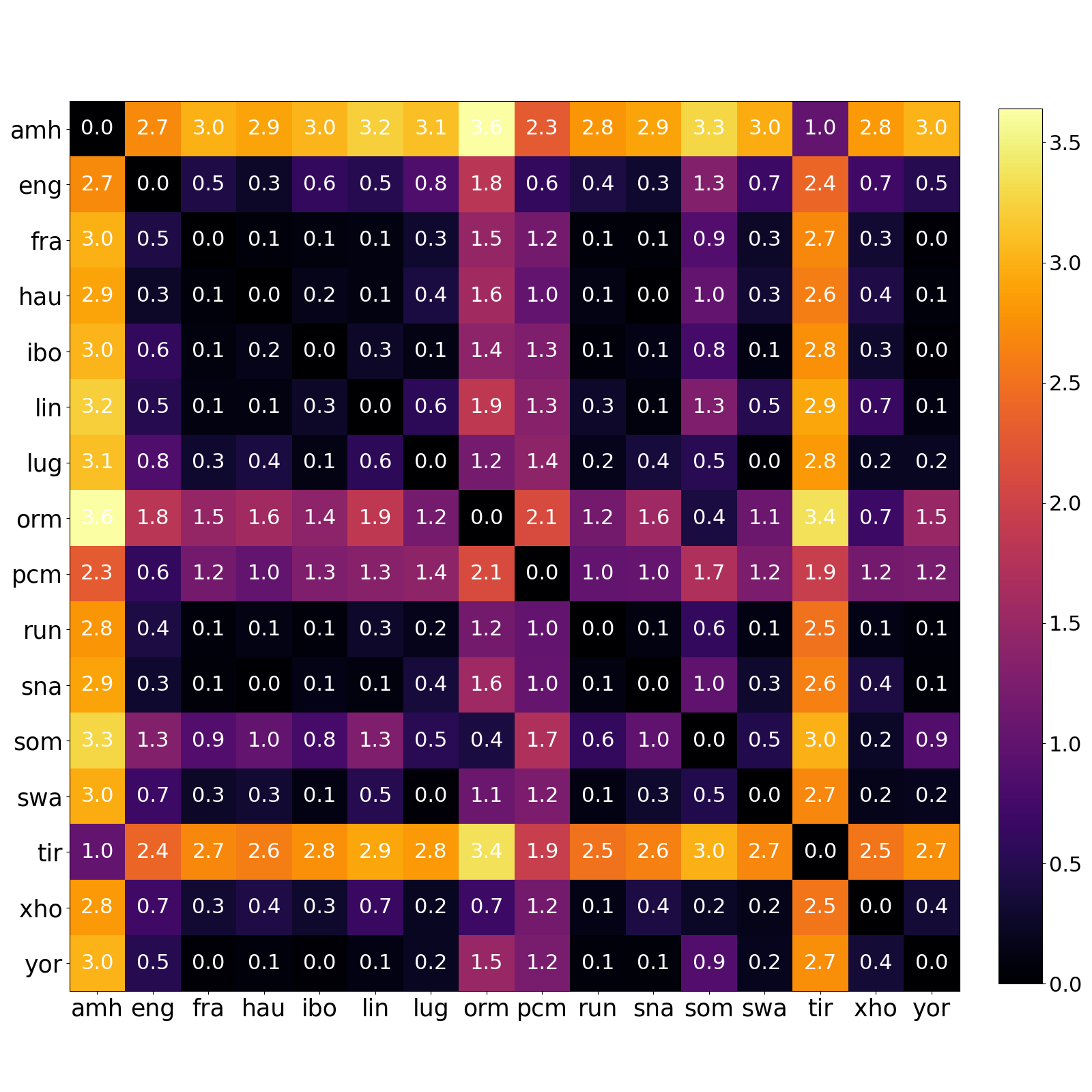}
    \caption{Cross-lingual $\bm{\lambda}$ distance among languages in our MasakhaNEWS setup. Each block shows the log-scale normalized average Euclidean distances between all pairs of clients' $\bm{\lambda}$ in their respective languages. The smaller the distance, the more similar $\bm{\lambda}$ is. }
    \label{fig:masakha_out}
\end{figure*}

Note that while the rank structure is the same across languages, the strength of personalization, absolute value of $\bm{\lambda}$, differs, as shown in Fig.~\ref{fig:masakha_out} in this Section and Fig.~\ref{fig:xnli_out} in the main paper. These two figures show the difference in $\bm{\lambda}$ across languages as described in Section~\ref{sec:analysis}. To sum up, the smaller the distance between two languages, represented as a block in the figure, the more similar their generated $\bm{\lambda}$ are. The results show that while similar languages sometimes exhibit similar $\bm{\lambda}$ values, unrelated languages also occasionally share similar $\bm{\lambda}$, consistent with findings in the literature that leveraging dissimilar languages can be beneficial.

\begin{table*}[]
\centering
\begin{tabular}{c|ccccc|ccccc}
\toprule
$\mathbf{r}$   & \multicolumn{5}{c|}{2}                                                                                                                                                                             & \multicolumn{5}{c}{4}                                                                                                                                                                               \\ \midrule
\textbf{Lang.} & \multicolumn{1}{c|}{\textbf{LoRA}} & \multicolumn{1}{c|}{\textbf{AdaLoRA}} & \multicolumn{1}{c|}{\textbf{BT-LoRA}} & \multicolumn{1}{c|}{\textbf{FedL2P}} & \textbf{\method{}} & \multicolumn{1}{c|}{\textbf{LoRA}} & \multicolumn{1}{c|}{\textbf{AdaLoRA}} & \multicolumn{1}{c|}{\textbf{BT-LoRA}} & \multicolumn{1}{c|}{\textbf{FedL2P}}   & \textbf{\method{}} \\ \hline
bg             & \multicolumn{1}{c|}{54.1±0.1}      & \multicolumn{1}{c|}{53.3±0.1}         & \multicolumn{1}{c|}{53.4±0.0}                & \multicolumn{1}{c|}{64.7±1.1}        & \textbf{70.7±0.8}               & \multicolumn{1}{c|}{56.7±0.3}      & \multicolumn{1}{c|}{53.3±0.1}         & \multicolumn{1}{c|}{53.3±0.1}                & \multicolumn{1}{c|}{66.5±1.1}          & \textbf{71.3±0.3}               \\ \hline
hi             & \multicolumn{1}{c|}{55.8±0.0}      & \multicolumn{1}{c|}{54.5±0.1}         & \multicolumn{1}{c|}{54.4±0.0}                & \multicolumn{1}{c|}{64.2±2.8}        & \textbf{70.1±1.0}               & \multicolumn{1}{c|}{59.1±0.2}      & \multicolumn{1}{c|}{54.4±0.2}         & \multicolumn{1}{c|}{54.5±0.2}                & \multicolumn{1}{c|}{65.4±1.8}          & \textbf{70.1±1.1}               \\ \hline
es             & \multicolumn{1}{c|}{57.1±0.1}      & \multicolumn{1}{c|}{55.3±0.1}         & \multicolumn{1}{c|}{55.4±0.0}                & \multicolumn{1}{c|}{67.9±1.3}        & \textbf{73.9±1.5}               & \multicolumn{1}{c|}{59.1±0.4}      & \multicolumn{1}{c|}{55.3±0.1}         & \multicolumn{1}{c|}{55.4±0.2}                & \multicolumn{1}{c|}{69.9±1.9}          & \textbf{73.3±2.3}               \\ \hline
el             & \multicolumn{1}{c|}{55.1±0.3}      & \multicolumn{1}{c|}{52.8±0.3}         & \multicolumn{1}{c|}{53.1±0.1}                & \multicolumn{1}{c|}{68.5±1.1}        & \textbf{73.5±0.2}               & \multicolumn{1}{c|}{57.3±0.2}      & \multicolumn{1}{c|}{53.2±0.0}         & \multicolumn{1}{c|}{53.0±0.2}                & \multicolumn{1}{c|}{70.5±0.9}          & \textbf{73.3±0.7}               \\ \hline
vi             & \multicolumn{1}{c|}{56.3±0.1}      & \multicolumn{1}{c|}{54.1±0.1}         & \multicolumn{1}{c|}{54.0±0.2}                & \multicolumn{1}{c|}{68.6±0.2}        & \textbf{73.4±0.6}               & \multicolumn{1}{c|}{59.0±0.3}      & \multicolumn{1}{c|}{54.1±0.2}         & \multicolumn{1}{c|}{54.5±0.2}                & \multicolumn{1}{c|}{70.1±0.3}          & \textbf{72.6±0.3}               \\ \hline
tr             & \multicolumn{1}{c|}{54.3±0.1}      & \multicolumn{1}{c|}{52.9±0.1}         & \multicolumn{1}{c|}{52.9±0.1}                & \multicolumn{1}{c|}{65.2±4.0}        & \textbf{71.9±2.1}               & \multicolumn{1}{c|}{56.9±0.2}      & \multicolumn{1}{c|}{52.9±0.1}         & \multicolumn{1}{c|}{52.9±0.1}                & \multicolumn{1}{c|}{66.9±4.1}          & \textbf{71.9±2.3}               \\ \hline
de             & \multicolumn{1}{c|}{55.1±0.1}      & \multicolumn{1}{c|}{53.9±0.1}         & \multicolumn{1}{c|}{54.1±0.1}                & \multicolumn{1}{c|}{67.3±0.7}        & \textbf{73.3±0.6}               & \multicolumn{1}{c|}{57.4±0.6}      & \multicolumn{1}{c|}{54.0±0.0}         & \multicolumn{1}{c|}{54.0±0.0}                & \multicolumn{1}{c|}{69.7±0.9}          & \textbf{74.6±0.2}               \\ \hline
ur             & \multicolumn{1}{c|}{53.6±0.2}      & \multicolumn{1}{c|}{52.9±0.2}         & \multicolumn{1}{c|}{53.2±0.0}                & \multicolumn{1}{c|}{67.0±0.4}        & \textbf{71.9±0.6}               & \multicolumn{1}{c|}{55.9±0.1}      & \multicolumn{1}{c|}{52.9±0.1}         & \multicolumn{1}{c|}{53.1±0.1}                & \multicolumn{1}{c|}{68.2±0.6}          & \textbf{72.9±0.4}               \\ \hline
en             & \multicolumn{1}{c|}{61.5±0.1}      & \multicolumn{1}{c|}{60.4±0.2}         & \multicolumn{1}{c|}{60.0±0.0}                & \multicolumn{1}{c|}{71.5±0.3}        & \textbf{74.5±0.5}               & \multicolumn{1}{c|}{63.5±0.2}      & \multicolumn{1}{c|}{60.3±0.2}         & \multicolumn{1}{c|}{60.6±0.2}                & \multicolumn{1}{c|}{72.8±0.2}          & \textbf{75.0±0.3}               \\ \hline
zh             & \multicolumn{1}{c|}{54.8±0.0}      & \multicolumn{1}{c|}{53.1±0.1}         & \multicolumn{1}{c|}{53.2±0.0}                & \multicolumn{1}{c|}{64.9±0.5}        & \textbf{73.8±0.4}               & \multicolumn{1}{c|}{57.3±0.1}      & \multicolumn{1}{c|}{53.1±0.1}         & \multicolumn{1}{c|}{53.1±0.1}                & \multicolumn{1}{c|}{67.4±1.2}          & \textbf{74.3±1.2}               \\ \hline
th             & \multicolumn{1}{c|}{50.7±0.1}      & \multicolumn{1}{c|}{49.4±0.2}         & \multicolumn{1}{c|}{49.3±0.1}                & \multicolumn{1}{c|}{61.7±1.7}        & \textbf{65.6±3.0}               & \multicolumn{1}{c|}{52.5±0.3}      & \multicolumn{1}{c|}{49.2±0.2}         & \multicolumn{1}{c|}{49.3±0.2}                & \multicolumn{1}{c|}{62.6±2.0}          & \textbf{66.1±3.2}               \\ \hline
sw             & \multicolumn{1}{c|}{52.5±0.1}      & \multicolumn{1}{c|}{50.7±0.1}         & \multicolumn{1}{c|}{50.5±0.1}                & \multicolumn{1}{c|}{64.6±2.8}        & \textbf{69.9±1.2}               & \multicolumn{1}{c|}{54.5±0.2}      & \multicolumn{1}{c|}{50.6±0.2}         & \multicolumn{1}{c|}{51.1±0.1}                & \multicolumn{1}{c|}{65.7±2.9}          & \textbf{68.1±0.3}               \\ \hline
ar             & \multicolumn{1}{c|}{53.4±0.2}      & \multicolumn{1}{c|}{51.4±0.2}         & \multicolumn{1}{c|}{51.5±0.1}                & \multicolumn{1}{c|}{65.8±2.0}        & \textbf{74.4±0.5}               & \multicolumn{1}{c|}{56.4±0.3}      & \multicolumn{1}{c|}{51.3±0.1}         & \multicolumn{1}{c|}{51.5±0.1}                & \multicolumn{1}{c|}{67.7±1.3}          & \textbf{75.1±1.6}               \\ \hline
fr             & \multicolumn{1}{c|}{53.1±0.1}      & \multicolumn{1}{c|}{52.8±0.2}         & \multicolumn{1}{c|}{52.6±0.2}                & \multicolumn{1}{c|}{63.4±0.0}        & \textbf{71.3±1.5}               & \multicolumn{1}{c|}{55.9±0.1}      & \multicolumn{1}{c|}{52.6±0.2}         & \multicolumn{1}{c|}{52.6±0.0}                & \multicolumn{1}{c|}{66.5±1.1}          & \textbf{72.2±0.3}               \\ \hline
ru             & \multicolumn{1}{c|}{55.3±0.1}      & \multicolumn{1}{c|}{54.2±0.2}         & \multicolumn{1}{c|}{54.0±0.0}                & \multicolumn{1}{c|}{67.4±0.4}        & \textbf{72.7±1.9}               & \multicolumn{1}{c|}{58.4±0.3}      & \multicolumn{1}{c|}{54.0±0.0}         & \multicolumn{1}{c|}{54.5±0.2}                & \multicolumn{1}{c|}{69.0±0.4}          & \textbf{73.5±1.6}               \\ 

\bottomrule \toprule
$\mathbf{r}$   & \multicolumn{5}{c|}{8} & \multicolumn{5}{c}{16}
    \\ \midrule
\textbf{Lang.} & \multicolumn{1}{c|}{\textbf{LoRA}} & \multicolumn{1}{c|}{\textbf{AdaLoRA}} & \multicolumn{1}{c|}{\textbf{BT-LoRA}} & \multicolumn{1}{c|}{\textbf{FedL2P}} & \textbf{\method{}} & \multicolumn{1}{c|}{\textbf{LoRA}} & \multicolumn{1}{c|}{\textbf{AdaLoRA}} & \multicolumn{1}{c|}{\textbf{BT-LoRA}} & \multicolumn{1}{c|}{\textbf{FedL2P}}   & \textbf{\method{}} \\ \hline
bg             & \multicolumn{1}{c|}{60.6±0.3}      & \multicolumn{1}{c|}{53.3±0.1}         & \multicolumn{1}{c|}{53.3±0.1}                & \multicolumn{1}{c|}{66.4±0.6}        & \textbf{70.5±0.2}               & \multicolumn{1}{c|}{67.1±0.3}      & \multicolumn{1}{c|}{53.4±0.0}         & \multicolumn{1}{c|}{53.5±0.1}                & \multicolumn{1}{c|}{68.0±1.0}          & \textbf{69.8±0.2}               \\ \hline
hi             & \multicolumn{1}{c|}{62.5±0.1}      & \multicolumn{1}{c|}{54.5±0.2}         & \multicolumn{1}{c|}{54.4±0.0}                & \multicolumn{1}{c|}{65.5±1.1}        & \textbf{69.3±0.8}               & \multicolumn{1}{c|}{67.8±0.0}      & \multicolumn{1}{c|}{54.4±0.0}         & \multicolumn{1}{c|}{54.7±0.1}                & \multicolumn{1}{c|}{65.9±1.3}          & \textbf{69.3±1.0}               \\ \hline
es             & \multicolumn{1}{c|}{63.7±0.1}      & \multicolumn{1}{c|}{55.3±0.1}         & \multicolumn{1}{c|}{55.5±0.2}                & \multicolumn{1}{c|}{70.5±3.5}        & \textbf{73.3±1.8}               & \multicolumn{1}{c|}{69.5±0.1}      & \multicolumn{1}{c|}{55.4±0.0}         & \multicolumn{1}{c|}{55.8±0.0}                & \multicolumn{1}{c|}{70.2±3.1}          & \textbf{73.5±1.5}               \\ \hline
el             & \multicolumn{1}{c|}{62.6±0.3}      & \multicolumn{1}{c|}{52.8±0.3}         & \multicolumn{1}{c|}{53.2±0.0}                & \multicolumn{1}{c|}{70.1±0.5}        & \textbf{71.3±1.4}               & \multicolumn{1}{c|}{71.3±0.2}      & \multicolumn{1}{c|}{52.6±0.2}         & \multicolumn{1}{c|}{53.4±0.0}                & \multicolumn{1}{c|}{\textbf{71.7±0.8}} & 70.4±1.8                        \\ \hline
vi             & \multicolumn{1}{c|}{64.7±0.1}      & \multicolumn{1}{c|}{54.3±0.2}         & \multicolumn{1}{c|}{54.7±0.1}                & \multicolumn{1}{c|}{70.5±1.7}        & \textbf{71.3±0.5}               & \multicolumn{1}{c|}{69.2±0.0}      & \multicolumn{1}{c|}{54.0±0.0}         & \multicolumn{1}{c|}{54.7±0.1}                & \multicolumn{1}{c|}{71.2±1.4}          & \textbf{71.3±0.6}               \\ \hline
tr             & \multicolumn{1}{c|}{61.9±0.4}      & \multicolumn{1}{c|}{53.0±0.0}         & \multicolumn{1}{c|}{53.0±0.0}                & \multicolumn{1}{c|}{68.2±2.2}        & \textbf{70.5±2.1}               & \multicolumn{1}{c|}{68.1±0.4}      & \multicolumn{1}{c|}{52.9±0.1}         & \multicolumn{1}{c|}{53.1±0.1}                & \multicolumn{1}{c|}{69.6±1.7}          & \textbf{70.7±2.9}               \\ \hline
de             & \multicolumn{1}{c|}{62.5±0.1}      & \multicolumn{1}{c|}{53.9±0.1}         & \multicolumn{1}{c|}{54.1±0.1}                & \multicolumn{1}{c|}{69.6±0.4}        & \textbf{75.3±0.4}               & \multicolumn{1}{c|}{69.0±0.3}      & \multicolumn{1}{c|}{53.9±0.1}         & \multicolumn{1}{c|}{54.5±0.2}                & \multicolumn{1}{c|}{71.2±1.0}          & \textbf{74.8±0.2}               \\ \hline
ur             & \multicolumn{1}{c|}{60.3±0.2}      & \multicolumn{1}{c|}{52.9±0.1}         & \multicolumn{1}{c|}{53.1±0.2}                & \multicolumn{1}{c|}{67.5±1.3}        & \textbf{73.1±0.8}               & \multicolumn{1}{c|}{68.7±0.2}      & \multicolumn{1}{c|}{52.9±0.1}         & \multicolumn{1}{c|}{53.3±0.1}                & \multicolumn{1}{c|}{69.0±1.0}          & \textbf{73.2±1.1}               \\ \hline
en             & \multicolumn{1}{c|}{67.3±0.2}      & \multicolumn{1}{c|}{60.1±0.3}         & \multicolumn{1}{c|}{60.6±0.0}                & \multicolumn{1}{c|}{72.6±0.6}        & \textbf{74.4±0.8}               & \multicolumn{1}{c|}{71.5±0.2}      & \multicolumn{1}{c|}{60.3±0.4}         & \multicolumn{1}{c|}{60.7±0.1}                & \multicolumn{1}{c|}{73.8±0.8}          & \textbf{74.5±1.2}               \\ \hline
zh             & \multicolumn{1}{c|}{61.4±0.2}      & \multicolumn{1}{c|}{53.0±0.0}         & \multicolumn{1}{c|}{53.5±0.1}                & \multicolumn{1}{c|}{67.1±1.3}        & \textbf{73.9±1.1}               & \multicolumn{1}{c|}{68.0±0.2}      & \multicolumn{1}{c|}{53.0±0.0}         & \multicolumn{1}{c|}{53.4±0.2}                & \multicolumn{1}{c|}{69.1±1.5}          & \textbf{73.5±0.8}               \\ \hline
th             & \multicolumn{1}{c|}{56.5±0.1}      & \multicolumn{1}{c|}{49.2±0.2}         & \multicolumn{1}{c|}{49.5±0.1}                & \multicolumn{1}{c|}{61.8±1.0}        & \textbf{66.7±3.8}               & \multicolumn{1}{c|}{62.8±0.0}      & \multicolumn{1}{c|}{49.3±0.1}         & \multicolumn{1}{c|}{49.4±0.0}                & \multicolumn{1}{c|}{64.0±1.1}          & \textbf{65.9±2.6}               \\ \hline
sw             & \multicolumn{1}{c|}{59.3±0.1}      & \multicolumn{1}{c|}{50.3±0.2}         & \multicolumn{1}{c|}{51.3±0.3}                & \multicolumn{1}{c|}{65.6±1.8}        & \textbf{68.1±0.9}               & \multicolumn{1}{c|}{67.3±0.2}      & \multicolumn{1}{c|}{50.3±0.1}         & \multicolumn{1}{c|}{51.7±0.1}                & \multicolumn{1}{c|}{67.8±1.8}          & \textbf{68.6±1.7}               \\ \hline
ar             & \multicolumn{1}{c|}{61.3±0.2}      & \multicolumn{1}{c|}{51.2±0.2}         & \multicolumn{1}{c|}{51.7±0.2}                & \multicolumn{1}{c|}{68.2±1.0}        & \textbf{74.7±1.2}               & \multicolumn{1}{c|}{66.8±0.2}      & \multicolumn{1}{c|}{51.2±0.3}         & \multicolumn{1}{c|}{52.1±0.2}                & \multicolumn{1}{c|}{69.3±1.2}          & \textbf{73.7±1.8}               \\ \hline
fr             & \multicolumn{1}{c|}{58.9±0.1}      & \multicolumn{1}{c|}{52.7±0.4}         & \multicolumn{1}{c|}{52.9±0.1}                & \multicolumn{1}{c|}{67.2±1.4}        & \textbf{71.9±0.2}               & \multicolumn{1}{c|}{65.1±0.2}      & \multicolumn{1}{c|}{52.8±0.3}         & \multicolumn{1}{c|}{53.1±0.1}                & \multicolumn{1}{c|}{68.5±1.3}          & \textbf{70.7±0.2}               \\ \hline
ru             & \multicolumn{1}{c|}{63.1±0.2}      & \multicolumn{1}{c|}{53.9±0.1}         & \multicolumn{1}{c|}{54.4±0.2}                & \multicolumn{1}{c|}{69.1±1.7}        & \textbf{72.9±1.0}               & \multicolumn{1}{c|}{67.3±0.1}      & \multicolumn{1}{c|}{53.8±0.2}         & \multicolumn{1}{c|}{54.8±0.0}                & \multicolumn{1}{c|}{69.9±1.9}          & \textbf{72.8±0.7}               \\ 
\bottomrule
\end{tabular}

\caption{Mean±SD Accuracy of each language across 3 different seeds for clients in the \seen{} pool of our XNLI setup. The pretrained model is trained using \textbf{DEPT(SPEC)} and the resulting \basemodel{} is personalized to each client given a baseline approach.}
\label{tab:xnli_seen_dept}
\end{table*}

\noindent\textbf{Additional Quantitative Results.} We show \textbf{extended} results for 1) our MasakhaNEWS setup in Tables~\ref{tab:masakha_seen_full} and \ref{tab:masakha_unseen_full}, 2) our XNLI setup with FedDPA-T used to pretrain the \basemodel{} in Table~\ref{tab:xnli_seen_feddpa_full}, and 3) our Fed-Aya setup with off-the-shelf Llama3.2 3B in Tables~\ref{tab:lama_fedaya_seen_full} and \ref{tab:lama_fedaya_unseen_full} and off-the-shelf Llama3.1 8B in Tables~\ref{tab:lama_fedaya_seen_full_8B} and \ref{tab:lama_fedaya_unseen_full_8B} for \seen{} and \unseen{} clients respectively.

We also present additional results to supplement the results in the main paper, namely 1) our XNLI setup with Standard FL (full fine-tuning) used to pretrain the \basemodel{} in Tables~\ref{tab:xnli_seen} and \ref{tab:xnli_unseen}, 2) our XNLI setup to show that \method{} is complementary with other pFL methods in Table~\ref{tab:xnli_seen_dept}, and 3) our Fed-Aya setup with MobileLLaMA as the \basemodel{} in Tables~\ref{tab:mobilellama_fedaya_seen} and \ref{tab:mobilellama_fedaya_unseen}. 

\clearpage
\begin{table*}[t]
\centering

\begin{tabular}{c|ccccc|ccccc}
\toprule
$\mathbf{r}$      & \multicolumn{5}{c|}{2}                                                                                                                                                                             & \multicolumn{5}{c}{4}                                                                                                                                                                             \\ \midrule
\textbf{Lang.} & \multicolumn{1}{c|}{\textbf{LoRA}} & \multicolumn{1}{c|}{\textbf{AdaLoRA}} & \multicolumn{1}{c|}{\textbf{BT-LoRA}} & \multicolumn{1}{c|}{\textbf{FedL2P}} & \textbf{\method{}} & \multicolumn{1}{c|}{\textbf{LoRA}} & \multicolumn{1}{c|}{\textbf{AdaLoRA}} & \multicolumn{1}{c|}{\textbf{BT-LoRA}} & \multicolumn{1}{c|}{\textbf{FedL2P}} & \textbf{\method{}} \\ \hline
bg                & \multicolumn{1}{c|}{47.5±0.2}      & \multicolumn{1}{c|}{45.7±0.1}         & \multicolumn{1}{c|}{45.7±0.2}                & \multicolumn{1}{c|}{66.7±0.9}        & \textbf{71.7±0.4}               & \multicolumn{1}{c|}{49.4±0.2}      & \multicolumn{1}{c|}{45.9±0.2}         & \multicolumn{1}{c|}{45.6±0.0}                & \multicolumn{1}{c|}{67.4±1.4}        & \textbf{72.8±0.3}               \\ \hline
hi                & \multicolumn{1}{c|}{45.9±0.3}      & \multicolumn{1}{c|}{44.4±0.0}         & \multicolumn{1}{c|}{44.4±0.0}                & \multicolumn{1}{c|}{69.5±0.8}        & \textbf{72.3±0.2}               & \multicolumn{1}{c|}{50.0±0.9}              & \multicolumn{1}{c|}{44.3±0.1}         & \multicolumn{1}{c|}{44.3±0.1}                & \multicolumn{1}{c|}{69.7±1.4}        & \textbf{72.1±0.4}               \\ \hline
es                & \multicolumn{1}{c|}{50.8±0.2}      & \multicolumn{1}{c|}{49.0±0.3}         & \multicolumn{1}{c|}{48.3±0.2}                & \multicolumn{1}{c|}{74.3±0.9}        & \textbf{75.4±0.6}               & \multicolumn{1}{c|}{54.5±0.6}      & \multicolumn{1}{c|}{48.6±0.2}         & \multicolumn{1}{c|}{48.9±0.2}                & \multicolumn{1}{c|}{74.5±0.3}        & \textbf{75.4±0.3}               \\ \hline
el                & \multicolumn{1}{c|}{50.8±0.2}      & \multicolumn{1}{c|}{49.1±0.2}         & \multicolumn{1}{c|}{48.8±0.2}                & \multicolumn{1}{c|}{70.7±1.0}        & \textbf{73.7±0.8}               & \multicolumn{1}{c|}{53.3±0.4}      & \multicolumn{1}{c|}{48.9±0.1}         & \multicolumn{1}{c|}{49.1±0.1}                & \multicolumn{1}{c|}{70.2±0.8}        & \textbf{74.3±0.3}               \\ \hline
vi                & \multicolumn{1}{c|}{50.8±0.3}      & \multicolumn{1}{c|}{48.0±0.2}         & \multicolumn{1}{c|}{48.1±0.2}                & \multicolumn{1}{c|}{71.3±0.8}        & \textbf{74.8±0.4}               & \multicolumn{1}{c|}{53.0±0.3}      & \multicolumn{1}{c|}{48.3±0.3}         & \multicolumn{1}{c|}{48.0±0.3}                & \multicolumn{1}{c|}{71.5±0.9}        & \textbf{74.9±0.1}               \\ \hline
tr                & \multicolumn{1}{c|}{48.8±0.6}      & \multicolumn{1}{c|}{47.1±0.3}         & \multicolumn{1}{c|}{46.8±0.3}                & \multicolumn{1}{c|}{71.3±0.6}        & \textbf{74.7±0.4}               & \multicolumn{1}{c|}{51.5±0.7}      & \multicolumn{1}{c|}{46.9±0.1}         & \multicolumn{1}{c|}{47.0±0.3}                & \multicolumn{1}{c|}{71.3±0.9}        & \textbf{75.2±0.3}               \\ \hline
de                & \multicolumn{1}{c|}{50.1±0.7}      & \multicolumn{1}{c|}{48.3±0.1}         & \multicolumn{1}{c|}{48.1±0.1}                & \multicolumn{1}{c|}{72.3±1.3}        & \textbf{75.0±0.7}               & \multicolumn{1}{c|}{53.3±0.6}      & \multicolumn{1}{c|}{48.5±0.1}         & \multicolumn{1}{c|}{48.5±0.1}                & \multicolumn{1}{c|}{72.7±0.9}        & \textbf{75.8±0.2}               \\ \hline
ur                & \multicolumn{1}{c|}{49.7±0.5}      & \multicolumn{1}{c|}{47.9±0.1}         & \multicolumn{1}{c|}{47.8±0.0}                & \multicolumn{1}{c|}{73.2±0.3}        & \textbf{74.3±0.6}               & \multicolumn{1}{c|}{52.9±1.1}      & \multicolumn{1}{c|}{47.8±0.0}         & \multicolumn{1}{c|}{47.8±0.2}                & \multicolumn{1}{c|}{73.1±0.3}        & \textbf{75.1±0.2}               \\ \hline
en                & \multicolumn{1}{c|}{53.1±0.4}      & \multicolumn{1}{c|}{50.9±0.1}         & \multicolumn{1}{c|}{50.9±0.2}                & \multicolumn{1}{c|}{75.3±0.8}        & \textbf{75.5±0.2}               & \multicolumn{1}{c|}{56.3±0.5}      & \multicolumn{1}{c|}{51.3±0.1}         & \multicolumn{1}{c|}{51.1±0.1}                & \multicolumn{1}{c|}{75.3±0.5}        & \textbf{75.9±0.2}               \\ \hline
zh                & \multicolumn{1}{c|}{50.0±0.3}      & \multicolumn{1}{c|}{48.2±0.2}         & \multicolumn{1}{c|}{48.0±0.0}                & \multicolumn{1}{c|}{68.3±1.3}        & \textbf{70.5±0.5}               & \multicolumn{1}{c|}{52.2±0.4}      & \multicolumn{1}{c|}{48.1±0.1}         & \multicolumn{1}{c|}{48.1±0.1}                & \multicolumn{1}{c|}{68.7±1.2}        & \textbf{71.6±0.2}               \\ \hline
th                & \multicolumn{1}{c|}{45.5±0.4}      & \multicolumn{1}{c|}{43.8±0.2}         & \multicolumn{1}{c|}{43.5±0.2}                & \multicolumn{1}{c|}{67.9±0.8}        & \textbf{71.3±0.1}               & \multicolumn{1}{c|}{49.1±0.6}      & \multicolumn{1}{c|}{43.7±0.1}         & \multicolumn{1}{c|}{43.7±0.2}                & \multicolumn{1}{c|}{68.3±0.5}        & \textbf{70.8±0.3}               \\ \hline
sw                & \multicolumn{1}{c|}{44.3±0.1}      & \multicolumn{1}{c|}{43.0±0.0}         & \multicolumn{1}{c|}{42.1±0.2}                & \multicolumn{1}{c|}{73.5±1.3}        & \textbf{73.7±0.4}               & \multicolumn{1}{c|}{48.8±0.0}      & \multicolumn{1}{c|}{42.7±0.2}         & \multicolumn{1}{c|}{42.4±0.2}                & \multicolumn{1}{c|}{73.7±0.7}        & \textbf{73.1±0.2}               \\ \hline
ar                & \multicolumn{1}{c|}{45.3±0.2}      & \multicolumn{1}{c|}{44.2±0.2}         & \multicolumn{1}{c|}{44.1±0.1}                & \multicolumn{1}{c|}{71.5±0.9}        & \textbf{75.6±0.6}               & \multicolumn{1}{c|}{49.2±0.3}      & \multicolumn{1}{c|}{44.2±0.3}         & \multicolumn{1}{c|}{44.2±0.0}                & \multicolumn{1}{c|}{71.5±1.6}        & \textbf{75.5±0.6}               \\ \hline
fr                & \multicolumn{1}{c|}{51.0±0.2}      & \multicolumn{1}{c|}{48.9±0.1}         & \multicolumn{1}{c|}{48.9±0.1}                & \multicolumn{1}{c|}{72.8±0.3}        & \textbf{74.6±0.6}               & \multicolumn{1}{c|}{54.3±0.3}      & \multicolumn{1}{c|}{48.7±0.1}         & \multicolumn{1}{c|}{48.9±0.1}                & \multicolumn{1}{c|}{73.5±0.5}        & \textbf{75.9±0.3}               \\ \hline
ru                & \multicolumn{1}{c|}{48.3±0.7}      & \multicolumn{1}{c|}{46.3±0.2}         & \multicolumn{1}{c|}{45.7±0.1}                & \multicolumn{1}{c|}{73.1±0.2}        & \textbf{74.1±0.7}               & \multicolumn{1}{c|}{51.7±1.1}      & \multicolumn{1}{c|}{45.9±0.2}         & \multicolumn{1}{c|}{46.0±0.2}                & \multicolumn{1}{c|}{72.9±0.5}        & \textbf{74.9±0.2}               \\ 

\bottomrule\toprule
$\mathbf{r}$      & \multicolumn{5}{c|}{8}                                                                                                                                                                             & \multicolumn{5}{c}{16}                                                                                                                                                                            \\ \midrule
\textbf{Lang.} & \multicolumn{1}{c|}{\textbf{LoRA}} & \multicolumn{1}{c|}{\textbf{AdaLoRA}} & \multicolumn{1}{c|}{\textbf{BT-LoRA}} & \multicolumn{1}{c|}{\textbf{FedL2P}} & \textbf{\method{}} & \multicolumn{1}{c|}{\textbf{LoRA}} & \multicolumn{1}{c|}{\textbf{AdaLoRA}} & \multicolumn{1}{c|}{\textbf{BT-LoRA}} & \multicolumn{1}{c|}{\textbf{FedL2P}} & \textbf{\method{}} \\ \hline
bg                & \multicolumn{1}{c|}{55.3±1.6}      & \multicolumn{1}{c|}{45.8±0.0}         & \multicolumn{1}{c|}{45.9±0.2}                & \multicolumn{1}{c|}{64.7±1.3}        & \textbf{73.9±0.3}               & \multicolumn{1}{c|}{63.9±1.5}      & \multicolumn{1}{c|}{45.8±0.2}         & \multicolumn{1}{c|}{46.3±0.1}                & \multicolumn{1}{c|}{61.5±2.9}        & \textbf{73.9±0.1}               \\ \hline
hi                & \multicolumn{1}{c|}{54.1±0.1}      & \multicolumn{1}{c|}{44.4±0.0}         & \multicolumn{1}{c|}{44.3±0.1}                & \multicolumn{1}{c|}{66.1±2.2}        & \textbf{70.7±0.1}               & \multicolumn{1}{c|}{64.2±0.0}      & \multicolumn{1}{c|}{44.4±0.0}         & \multicolumn{1}{c|}{44.9±0.2}                & \multicolumn{1}{c|}{62.3±3.4}        & \textbf{70.8±0.2}               \\ \hline
es                & \multicolumn{1}{c|}{59.9±0.5}      & \multicolumn{1}{c|}{48.5±0.1}         & \multicolumn{1}{c|}{49.1±0.5}                & \multicolumn{1}{c|}{72.1±1.8}        & \textbf{75.9±0.2}               & \multicolumn{1}{c|}{70.4±0.2}      & \multicolumn{1}{c|}{48.7±0.2}         & \multicolumn{1}{c|}{49.7±0.2}                & \multicolumn{1}{c|}{68.7±3.6}        & \textbf{75.9±0.1}               \\ \hline
el                & \multicolumn{1}{c|}{58.2±1.3}      & \multicolumn{1}{c|}{49.0±0.2}         & \multicolumn{1}{c|}{49.3±0.1}                & \multicolumn{1}{c|}{67.6±1.4}        & \textbf{73.8±0.2}               & \multicolumn{1}{c|}{67.1±1.3}      & \multicolumn{1}{c|}{48.9±0.2}         & \multicolumn{1}{c|}{49.6±0.0}                & \multicolumn{1}{c|}{65.1±1.9}        & \textbf{73.7±0.7}               \\ \hline
vi                & \multicolumn{1}{c|}{58.1±0.2}      & \multicolumn{1}{c|}{48.1±0.1}         & \multicolumn{1}{c|}{48.2±0.0}                & \multicolumn{1}{c|}{68.8±1.9}        & \textbf{74.3±0.5}               & \multicolumn{1}{c|}{68.5±0.2}      & \multicolumn{1}{c|}{48.1±0.2}         & \multicolumn{1}{c|}{48.6±0.2}                & \multicolumn{1}{c|}{65.1±3.1}        & \textbf{74.9±0.2}               \\ \hline
tr                & \multicolumn{1}{c|}{56.5±0.2}      & \multicolumn{1}{c|}{46.9±0.1}         & \multicolumn{1}{c|}{47.1±0.1}                & \multicolumn{1}{c|}{68.5±1.9}        & \textbf{75.6±0.2}               & \multicolumn{1}{c|}{66.1±0.7}      & \multicolumn{1}{c|}{47.0±0.3}         & \multicolumn{1}{c|}{47.5±0.2}                & \multicolumn{1}{c|}{65.1±3.2}        & \textbf{75.3±0.2}               \\ \hline
de                & \multicolumn{1}{c|}{59.7±0.1}      & \multicolumn{1}{c|}{48.3±0.1}         & \multicolumn{1}{c|}{48.6±0.2}                & \multicolumn{1}{c|}{69.4±2.0}        & \textbf{74.9±0.1}               & \multicolumn{1}{c|}{68.1±0.2}      & \multicolumn{1}{c|}{48.3±0.1}         & \multicolumn{1}{c|}{49.0±0.0}                & \multicolumn{1}{c|}{65.8±3.6}        & \textbf{74.4±0.2}               \\ \hline
ur                & \multicolumn{1}{c|}{58.5±1.3}      & \multicolumn{1}{c|}{48.0±0.2}         & \multicolumn{1}{c|}{47.8±0.0}                & \multicolumn{1}{c|}{70.9±2.2}        & \textbf{74.3±0.2}               & \multicolumn{1}{c|}{69.7±0.6}      & \multicolumn{1}{c|}{47.7±0.1}         & \multicolumn{1}{c|}{48.0±0.0}                & \multicolumn{1}{c|}{67.7±2.9}        & \textbf{74.1±0.2}               \\ \hline
en                & \multicolumn{1}{c|}{63.4±0.7}      & \multicolumn{1}{c|}{50.8±0.2}         & \multicolumn{1}{c|}{51.0±0.3}                & \multicolumn{1}{c|}{73.5±1.2}        & \textbf{76.5±0.1}               & \multicolumn{1}{c|}{72.5±0.9}      & \multicolumn{1}{c|}{50.9±0.2}         & \multicolumn{1}{c|}{51.5±0.2}                & \multicolumn{1}{c|}{71.3±2.1}        & \textbf{76.5±0.2}               \\ \hline
zh                & \multicolumn{1}{c|}{57.1±0.3}      & \multicolumn{1}{c|}{48.1±0.1}         & \multicolumn{1}{c|}{48.6±0.2}                & \multicolumn{1}{c|}{65.9±1.9}        & \textbf{72.5±0.5}               & \multicolumn{1}{c|}{64.7±1.2}      & \multicolumn{1}{c|}{48.2±0.2}         & \multicolumn{1}{c|}{49.0±0.2}                & \multicolumn{1}{c|}{62.7±2.8}        & \textbf{72.4±0.3}               \\ \hline
th                & \multicolumn{1}{c|}{56.1±0.2}      & \multicolumn{1}{c|}{43.7±0.1}         & \multicolumn{1}{c|}{44.1±0.1}                & \multicolumn{1}{c|}{65.7±2.0}        & \textbf{70.7±0.1}               & \multicolumn{1}{c|}{65.5±1.1}      & \multicolumn{1}{c|}{43.5±0.1}         & \multicolumn{1}{c|}{44.7±0.1}                & \multicolumn{1}{c|}{62.3±3.3}        & \textbf{70.8±0.3}               \\ \hline
sw                & \multicolumn{1}{c|}{56.1±0.5}      & \multicolumn{1}{c|}{42.4±0.2}         & \multicolumn{1}{c|}{42.9±0.1}                & \multicolumn{1}{c|}{70.7±3.0}        & \textbf{71.7±0.3}               & \multicolumn{1}{c|}{69.9±0.9}      & \multicolumn{1}{c|}{42.5±0.1}         & \multicolumn{1}{c|}{43.1±0.2}                & \multicolumn{1}{c|}{67.2±3.9}        & \textbf{71.1±0.6}               \\ \hline
ar                & \multicolumn{1}{c|}{55.9±0.1}      & \multicolumn{1}{c|}{44.2±0.2}         & \multicolumn{1}{c|}{44.4±0.0}                & \multicolumn{1}{c|}{67.7±1.9}        & \textbf{76.9±0.2}               & \multicolumn{1}{c|}{67.2±0.2}      & \multicolumn{1}{c|}{44.2±0.0}         & \multicolumn{1}{c|}{44.3±0.2}                & \multicolumn{1}{c|}{63.9±3.8}        & \textbf{76.2±0.4}               \\ \hline
fr                & \multicolumn{1}{c|}{59.2±0.3}      & \multicolumn{1}{c|}{48.9±0.1}         & \multicolumn{1}{c|}{49.2±0.4}                & \multicolumn{1}{c|}{70.5±1.9}        & \textbf{75.9±0.2}               & \multicolumn{1}{c|}{68.7±0.1}      & \multicolumn{1}{c|}{48.9±0.1}         & \multicolumn{1}{c|}{49.1±0.5}                & \multicolumn{1}{c|}{68.1±2.9}        & \textbf{75.9±0.1}               \\ \hline
ru                & \multicolumn{1}{c|}{58.2±2.0}      & \multicolumn{1}{c|}{45.9±0.1}         & \multicolumn{1}{c|}{46.2±0.2}                & \multicolumn{1}{c|}{70.5±1.9}        & \textbf{75.0±0.6}               & \multicolumn{1}{c|}{68.0±0.8}      & \multicolumn{1}{c|}{45.8±0.0}         & \multicolumn{1}{c|}{46.7±0.3}                & \multicolumn{1}{c|}{67.2±3.4}        & \textbf{75.1±0.1}               \\ 
\bottomrule
\end{tabular}

\caption{Mean±SD Accuracy of each language across 3 different seeds for clients in the \seen{} pool of our XNLI setup. The pretrained model is trained using \textbf{Standard FL with full fine-tuning} and the resulting \basemodel{} is personalized to each client given a baseline approach.}
\label{tab:xnli_seen}
\end{table*}

\begin{table*}[t]
\centering

\begin{tabular}{c|ccccc|ccccc}
\toprule
$\mathbf{r}$   & \multicolumn{5}{c|}{2}                                                                                                                                                                               & \multicolumn{5}{c}{4}                                                                                                                                                                                   \\ \midrule
\textbf{Lang.} & \multicolumn{1}{c|}{\textbf{LoRA}} & \multicolumn{1}{c|}{\textbf{AdaLoRA}} & \multicolumn{1}{c|}{\textbf{BT-LoRA}} & \multicolumn{1}{c|}{\textbf{FedL2P}}   & \textbf{\method{}} & \multicolumn{1}{c|}{\textbf{LoRA}}     & \multicolumn{1}{c|}{\textbf{AdaLoRA}} & \multicolumn{1}{c|}{\textbf{BT-LoRA}} & \multicolumn{1}{c|}{\textbf{FedL2P}}   & \textbf{\method{}} \\ \hline
bg             & \multicolumn{1}{c|}{49.3±0.1}      & \multicolumn{1}{c|}{49.3±0.1}         & \multicolumn{1}{c|}{49.1±0.1}                & \multicolumn{1}{c|}{\textbf{54.4±0.4}} & 54.0±0.6                        & \multicolumn{1}{c|}{49.9±0.1}          & \multicolumn{1}{c|}{49.3±0.1}         & \multicolumn{1}{c|}{49.3±0.2}                & \multicolumn{1}{c|}{54.7±0.4}          & \textbf{55.3±0.3}               \\ \hline
hi             & \multicolumn{1}{c|}{45.9±0.2}      & \multicolumn{1}{c|}{45.3±0.1}         & \multicolumn{1}{c|}{45.1±0.2}                & \multicolumn{1}{c|}{49.9±0.2}          & \textbf{51.1±0.3}               & \multicolumn{1}{c|}{47.1±0.1}          & \multicolumn{1}{c|}{45.2±0.2}         & \multicolumn{1}{c|}{45.1±0.1}                & \multicolumn{1}{c|}{\textbf{50.9±0.7}} & 50.6±0.9                        \\ \hline
es             & \multicolumn{1}{c|}{51.1±0.1}      & \multicolumn{1}{c|}{50.3±0.2}         & \multicolumn{1}{c|}{50.2±0.0}                & \multicolumn{1}{c|}{56.8±0.3}          & \textbf{57.7±0.1}               & \multicolumn{1}{c|}{52.3±0.3}          & \multicolumn{1}{c|}{50.1±0.2}         & \multicolumn{1}{c|}{50.0±0.2}                & \multicolumn{1}{c|}{57.3±0.3}          & \textbf{58.2±0.2}               \\ \hline
el             & \multicolumn{1}{c|}{48.5±0.2}      & \multicolumn{1}{c|}{47.8±0.2}         & \multicolumn{1}{c|}{48.0±0.2}                & \multicolumn{1}{c|}{\textbf{51.4±0.3}} & 51.1±0.7                        & \multicolumn{1}{c|}{49.7±0.2}          & \multicolumn{1}{c|}{47.9±0.2}         & \multicolumn{1}{c|}{48.2±0.2}                & \multicolumn{1}{c|}{\textbf{51.5±0.3}} & 51.3±0.4                        \\ \hline
vi             & \multicolumn{1}{c|}{48.5±0.2}      & \multicolumn{1}{c|}{47.3±0.1}         & \multicolumn{1}{c|}{47.1±0.2}                & \multicolumn{1}{c|}{54.0±0.7}          & \textbf{54.5±0.2}               & \multicolumn{1}{c|}{49.9±0.2}          & \multicolumn{1}{c|}{47.3±0.1}         & \multicolumn{1}{c|}{47.1±0.1}                & \multicolumn{1}{c|}{\textbf{54.7±1.1}} & 54.6±0.3                        \\ \hline
tr             & \multicolumn{1}{c|}{45.3±0.2}      & \multicolumn{1}{c|}{44.6±0.0}         & \multicolumn{1}{c|}{44.7±0.1}                & \multicolumn{1}{c|}{48.3±0.5}          & \textbf{49.7±0.1}               & \multicolumn{1}{c|}{46.5±0.3}          & \multicolumn{1}{c|}{44.6±0.0}         & \multicolumn{1}{c|}{44.7±0.1}                & \multicolumn{1}{c|}{47.7±0.2}          & \textbf{48.8±0.3}               \\ \hline
de             & \multicolumn{1}{c|}{49.8±0.2}      & \multicolumn{1}{c|}{49.3±0.3}         & \multicolumn{1}{c|}{49.1±0.2}                & \multicolumn{1}{c|}{\textbf{54.0±0.6}} & 53.3±0.7                        & \multicolumn{1}{c|}{50.3±0.1}          & \multicolumn{1}{c|}{49.3±0.2}         & \multicolumn{1}{c|}{49.4±0.0}                & \multicolumn{1}{c|}{\textbf{54.2±0.4}} & 52.7±0.6                        \\ \hline
ur             & \multicolumn{1}{c|}{46.9±0.3}      & \multicolumn{1}{c|}{46.6±0.2}         & \multicolumn{1}{c|}{46.5±0.2}                & \multicolumn{1}{c|}{50.1±0.4}          & \textbf{53.1±1.0}               & \multicolumn{1}{c|}{48.2±0.2}          & \multicolumn{1}{c|}{46.5±0.1}         & \multicolumn{1}{c|}{46.4±0.2}                & \multicolumn{1}{c|}{50.9±0.2}          & \textbf{53.5±0.5}               \\ \hline
en             & \multicolumn{1}{c|}{55.4±0.3}      & \multicolumn{1}{c|}{54.9±0.2}         & \multicolumn{1}{c|}{54.7±0.1}                & \multicolumn{1}{c|}{\textbf{59.9±0.4}} & 59.6±0.3                        & \multicolumn{1}{c|}{56.5±0.6}          & \multicolumn{1}{c|}{54.7±0.2}         & \multicolumn{1}{c|}{54.9±0.1}                & \multicolumn{1}{c|}{\textbf{59.8±0.6}} & 59.1±0.7                        \\ \hline
zh             & \multicolumn{1}{c|}{49.3±0.2}      & \multicolumn{1}{c|}{48.7±0.2}         & \multicolumn{1}{c|}{48.7±0.2}                & \multicolumn{1}{c|}{\textbf{52.5±0.2}} & 51.5±0.4                        & \multicolumn{1}{c|}{49.8±0.2}          & \multicolumn{1}{c|}{48.6±0.2}         & \multicolumn{1}{c|}{48.9±0.1}                & \multicolumn{1}{c|}{\textbf{52.7±0.1}} & 51.5±0.3                        \\ \hline
th             & \multicolumn{1}{c|}{44.4±0.0}      & \multicolumn{1}{c|}{44.7±0.1}         & \multicolumn{1}{c|}{44.8±0.2}                & \multicolumn{1}{c|}{48.0±0.3}          & \textbf{48.9±0.2}               & \multicolumn{1}{c|}{45.0±0.4}          & \multicolumn{1}{c|}{44.6±0.2}         & \multicolumn{1}{c|}{44.6±0.0}                & \multicolumn{1}{c|}{47.9±0.5}          & \textbf{48.5±0.4}               \\ \hline
sw             & \multicolumn{1}{c|}{41.2±0.2}      & \multicolumn{1}{c|}{40.1±0.1}         & \multicolumn{1}{c|}{40.1±0.1}                & \multicolumn{1}{c|}{44.5±0.3}          & \textbf{46.5±0.7}               & \multicolumn{1}{c|}{41.4±0.0}          & \multicolumn{1}{c|}{40.1±0.2}         & \multicolumn{1}{c|}{40.1±0.1}                & \multicolumn{1}{c|}{44.8±0.0}          & \textbf{47.2±0.6}               \\ \hline
ar             & \multicolumn{1}{c|}{46.9±0.1}      & \multicolumn{1}{c|}{46.7±0.1}         & \multicolumn{1}{c|}{46.6±0.2}                & \multicolumn{1}{c|}{\textbf{50.3±0.4}} & 49.0±0.4                        & \multicolumn{1}{c|}{47.9±0.1}          & \multicolumn{1}{c|}{46.5±0.1}         & \multicolumn{1}{c|}{46.6±0.0}                & \multicolumn{1}{c|}{\textbf{50.2±0.3}}          & 49.0±0.2               \\ \hline
fr             & \multicolumn{1}{c|}{48.3±0.1}      & \multicolumn{1}{c|}{48.7±0.1}         & \multicolumn{1}{c|}{48.7±0.1}                & \multicolumn{1}{c|}{53.0±0.3}          & \textbf{56.3±0.7}               & \multicolumn{1}{c|}{48.7±0.1}          & \multicolumn{1}{c|}{48.8±0.0}         & \multicolumn{1}{c|}{48.5±0.1}                & \multicolumn{1}{c|}{53.5±0.5}          & \textbf{56.7±0.7}               \\ \hline
ru             & \multicolumn{1}{c|}{45.1±0.1}      & \multicolumn{1}{c|}{44.5±0.1}         & \multicolumn{1}{c|}{44.3±0.1}                & \multicolumn{1}{c|}{49.1±0.7}          & \textbf{52.5±0.4}               & \multicolumn{1}{c|}{45.8±0.2}          & \multicolumn{1}{c|}{44.5±0.3}         & \multicolumn{1}{c|}{44.4±0.0}                & \multicolumn{1}{c|}{49.5±0.9}          & \textbf{52.1±0.4}               \\ 

\bottomrule \toprule
$\mathbf{r}$   & \multicolumn{5}{c|}{8}                                                                                                                                                                               & \multicolumn{5}{c}{16}                                                                                                                                                                                  \\ \midrule
\textbf{Lang.} & \multicolumn{1}{c|}{\textbf{LoRA}} & \multicolumn{1}{c|}{\textbf{AdaLoRA}} & \multicolumn{1}{c|}{\textbf{BT-LoRA}} & \multicolumn{1}{c|}{\textbf{FedL2P}}   & \textbf{\method{}} & \multicolumn{1}{c|}{\textbf{LoRA}}     & \multicolumn{1}{c|}{\textbf{AdaLoRA}} & \multicolumn{1}{c|}{\textbf{BT-LoRA}} & \multicolumn{1}{c|}{\textbf{FedL2P}}   & \textbf{\method{}} \\ \hline
bg             & \multicolumn{1}{c|}{51.0±0.3}      & \multicolumn{1}{c|}{49.2±0.2}         & \multicolumn{1}{c|}{49.3±0.1}                & \multicolumn{1}{c|}{\textbf{54.1±0.2}} & 53.4±0.0                        & \multicolumn{1}{c|}{52.7±0.2}          & \multicolumn{1}{c|}{49.1±0.1}         & \multicolumn{1}{c|}{49.3±0.2}                & \multicolumn{1}{c|}{52.5±1.2}          & \textbf{53.7±0.5}               \\ \hline
hi             & \multicolumn{1}{c|}{48.1±0.5}      & \multicolumn{1}{c|}{45.4±0.2}         & \multicolumn{1}{c|}{45.1±0.2}                & \multicolumn{1}{c|}{49.6±1.1}          & \textbf{50.6±0.3}               & \multicolumn{1}{c|}{48.8±0.0}          & \multicolumn{1}{c|}{45.2±0.0}         & \multicolumn{1}{c|}{45.3±0.2}                & \multicolumn{1}{c|}{49.0±0.3}          & \textbf{49.9±0.5}               \\ \hline
es             & \multicolumn{1}{c|}{54.0±0.3}      & \multicolumn{1}{c|}{50.1±0.2}         & \multicolumn{1}{c|}{50.3±0.1}                & \multicolumn{1}{c|}{57.5±0.8}          & \textbf{57.7±0.6}               & \multicolumn{1}{c|}{55.5±0.2}          & \multicolumn{1}{c|}{50.2±0.2}         & \multicolumn{1}{c|}{50.2±0.0}                & \multicolumn{1}{c|}{56.4±1.4}          & \textbf{57.1±0.5}               \\ \hline
el             & \multicolumn{1}{c|}{50.9±0.3}      & \multicolumn{1}{c|}{48.1±0.1}         & \multicolumn{1}{c|}{47.9±0.1}                & \multicolumn{1}{c|}{50.9±0.5}          & 50.5±0.3                        & \multicolumn{1}{c|}{\textbf{51.1±1.1}} & \multicolumn{1}{c|}{47.9±0.2}         & \multicolumn{1}{c|}{47.9±0.1}                & \multicolumn{1}{c|}{50.5±1.0}          & 49.9±0.2                        \\ \hline
vi             & \multicolumn{1}{c|}{51.2±0.3}      & \multicolumn{1}{c|}{47.1±0.1}         & \multicolumn{1}{c|}{47.2±0.2}                & \multicolumn{1}{c|}{53.2±0.3}          & \textbf{55.1±0.4}               & \multicolumn{1}{c|}{53.5±0.4}          & \multicolumn{1}{c|}{46.9±0.2}         & \multicolumn{1}{c|}{47.5±0.2}                & \multicolumn{1}{c|}{52.5±0.7}          & \textbf{54.7±0.9}               \\ \hline
tr             & \multicolumn{1}{c|}{46.9±0.2}      & \multicolumn{1}{c|}{44.7±0.1}         & \multicolumn{1}{c|}{44.7±0.1}                & \multicolumn{1}{c|}{47.6±0.4}          & \textbf{48.0±0.8}               & \multicolumn{1}{c|}{\textbf{48.9±0.8}} & \multicolumn{1}{c|}{44.5±0.1}         & \multicolumn{1}{c|}{44.8±0.2}                & \multicolumn{1}{c|}{47.8±0.3}          & 47.7±0.4                        \\ \hline
de             & \multicolumn{1}{c|}{52.1±1.1}      & \multicolumn{1}{c|}{49.1±0.1}         & \multicolumn{1}{c|}{49.3±0.2}                & \multicolumn{1}{c|}{\textbf{54.2±0.2}} & 52.5±0.3                        & \multicolumn{1}{c|}{54.4±0.9}          & \multicolumn{1}{c|}{49.3±0.2}         & \multicolumn{1}{c|}{49.4±0.3}                & \multicolumn{1}{c|}{\textbf{53.9±0.4}} & 52.3±0.4                        \\ \hline
ur             & \multicolumn{1}{c|}{48.3±0.2}      & \multicolumn{1}{c|}{46.3±0.1}         & \multicolumn{1}{c|}{46.3±0.2}                & \multicolumn{1}{c|}{49.9±0.1}          & \textbf{51.9±0.2}               & \multicolumn{1}{c|}{49.4±1.0}          & \multicolumn{1}{c|}{46.5±0.1}         & \multicolumn{1}{c|}{46.5±0.1}                & \multicolumn{1}{c|}{48.8±0.8}          & \textbf{52.3±0.5}               \\ \hline
en             & \multicolumn{1}{c|}{58.3±0.2}      & \multicolumn{1}{c|}{54.9±0.2}         & \multicolumn{1}{c|}{54.9±0.1}                & \multicolumn{1}{c|}{\textbf{59.5±0.6}} & 57.0±1.0                        & \multicolumn{1}{c|}{58.9±1.1}          & \multicolumn{1}{c|}{54.8±0.2}         & \multicolumn{1}{c|}{55.3±0.1}                & \multicolumn{1}{c|}{\textbf{59.1±0.4}} & 57.9±0.7                        \\ \hline
zh             & \multicolumn{1}{c|}{50.1±0.2}      & \multicolumn{1}{c|}{48.5±0.3}         & \multicolumn{1}{c|}{48.5±0.1}                & \multicolumn{1}{c|}{\textbf{52.3±0.9}} & 50.7±0.1                        & \multicolumn{1}{c|}{\textbf{51.8±0.5}} & \multicolumn{1}{c|}{48.7±0.1}         & \multicolumn{1}{c|}{48.9±0.3}                & \multicolumn{1}{c|}{51.7±0.8}          & 50.3±0.7                        \\ \hline
th             & \multicolumn{1}{c|}{45.8±0.2}      & \multicolumn{1}{c|}{44.6±0.0}         & \multicolumn{1}{c|}{44.5±0.1}                & \multicolumn{1}{c|}{47.5±0.2}          & \textbf{47.6±0.0}               & \multicolumn{1}{c|}{46.9±0.1}          & \multicolumn{1}{c|}{44.7±0.2}         & \multicolumn{1}{c|}{44.5±0.1}                & \multicolumn{1}{c|}{47.1±0.8}          & \textbf{47.6±0.3}               \\ \hline
sw             & \multicolumn{1}{c|}{42.4±0.0}      & \multicolumn{1}{c|}{40.1±0.1}         & \multicolumn{1}{c|}{40.3±0.1}                & \multicolumn{1}{c|}{43.6±0.7}          & \textbf{46.7±0.5}               & \multicolumn{1}{c|}{43.6±1.0}          & \multicolumn{1}{c|}{40.1±0.1}         & \multicolumn{1}{c|}{40.2±0.3}                & \multicolumn{1}{c|}{43.0±1.1}          & \textbf{46.2±0.2}               \\ \hline
ar             & \multicolumn{1}{c|}{48.1±1.1}      & \multicolumn{1}{c|}{46.7±0.1}         & \multicolumn{1}{c|}{46.7±0.1}                & \multicolumn{1}{c|}{\textbf{50.3±0.5}} & 48.5±0.7                        & \multicolumn{1}{c|}{49.1±1.1}          & \multicolumn{1}{c|}{46.7±0.2}         & \multicolumn{1}{c|}{46.5±0.1}                & \multicolumn{1}{c|}{\textbf{49.3±0.8}} & 49.0±0.5                        \\ \hline
fr             & \multicolumn{1}{c|}{50.9±0.1}      & \multicolumn{1}{c|}{48.7±0.2}         & \multicolumn{1}{c|}{48.5±0.1}                & \multicolumn{1}{c|}{52.6±0.6}          & \textbf{55.0±0.3}               & \multicolumn{1}{c|}{52.1±0.8}          & \multicolumn{1}{c|}{48.7±0.1}         & \multicolumn{1}{c|}{48.7±0.2}                & \multicolumn{1}{c|}{52.1±1.2}          & \textbf{54.6±0.7}               \\ \hline
ru             & \multicolumn{1}{c|}{46.8±0.6}      & \multicolumn{1}{c|}{44.3±0.2}         & \multicolumn{1}{c|}{44.5±0.4}                & \multicolumn{1}{c|}{48.8±0.6}          & \textbf{50.7±0.8}               & \multicolumn{1}{c|}{48.9±2.0}          & \multicolumn{1}{c|}{44.4±0.3}         & \multicolumn{1}{c|}{44.7±0.2}                & \multicolumn{1}{c|}{48.3±0.7}          & \textbf{49.7±0.6}               \\ 
\bottomrule
\end{tabular}

\caption{Mean±SD Accuracy of each language across 3 different seeds for clients in the \unseen{} pool of our XNLI setup. The pretrained model is trained using \textbf{Standard FL with full fine-tuning} and the resulting \basemodel{} is personalized to each client given a baseline approach.}
\label{tab:xnli_unseen}
\end{table*}
\begin{table*}[t]
\centering

\begin{tabular}{c|ccccc}
\toprule
$\mathbf{r}$      & \multicolumn{5}{c}{2}                                                                                                                                                                                                                                                                                   \\ \midrule
\textbf{Lang.} & \multicolumn{1}{c|}{\textbf{LoRA}}                            & \multicolumn{1}{c|}{\textbf{AdaLoRA}}                         & \multicolumn{1}{c|}{\textbf{BT-LoRA}}                  & \multicolumn{1}{c|}{\textbf{FedL2P}}                          & \textbf{\method{}}          \\ \hline
te                & \multicolumn{1}{c|}{0.12/0.05/0.05}                           & \multicolumn{1}{c|}{0.12/0.06/0.06}                           & \multicolumn{1}{c|}{0.12/0.06/0.06}                           & \multicolumn{1}{c|}{\textbf{0.15/0.07/0.07}}                  & 0.13/0.06/0.06                           \\ \hline
ar                & \multicolumn{1}{c|}{0.18/0.02/\textbf{0.02}}                           & \multicolumn{1}{c|}{0.18/0.02/\textbf{0.02}}                           & \multicolumn{1}{c|}{0.18/0.02/\textbf{0.02}}                           & \multicolumn{1}{c|}{0.20/0.02/\textbf{0.02}}                           & \textbf{0.21/0.03/0.02}                  \\ \hline
es                & \multicolumn{1}{c|}{0.32/0.35/0.33}                           & \multicolumn{1}{c|}{0.33/0.36/0.33}                           & \multicolumn{1}{c|}{0.29/0.33/0.31}                           & \multicolumn{1}{c|}{0.33/0.35/0.32}                           & \textbf{0.34/0.37/0.34}                  \\ \hline
en                & \multicolumn{1}{c|}{0.25/0.31/0.25}                           & \multicolumn{1}{c|}{0.25/0.32/0.25}                           & \multicolumn{1}{c|}{0.23/0.29/0.24}                           & \multicolumn{1}{c|}{0.26/0.33/0.27}                           & \textbf{0.27/0.35/0.28}                  \\ \hline
fr                & \multicolumn{1}{c|}{0.20/0.25/0.20}                           & \multicolumn{1}{c|}{0.20/0.24/0.19}                           & \multicolumn{1}{c|}{0.19/0.24/0.20}                           & \multicolumn{1}{c|}{\textbf{0.23}/0.25/0.20}                           & \textbf{0.23/0.28/0.22}                  \\ \hline
zh                & \multicolumn{1}{c|}{0.06/\textbf{0.09/0.09}} & \multicolumn{1}{c|}{\textbf{0.07}/0.08/0.08} & \multicolumn{1}{c|}{0.06/0.08/0.08}                           & \multicolumn{1}{c|}{0.05/0.08/0.08}                           & 0.05/0.08/0.08                           \\ \hline
pt                & \multicolumn{1}{c|}{0.26/0.33/0.29}                           & \multicolumn{1}{c|}{0.27/0.33/0.30}                           & \multicolumn{1}{c|}{0.24/0.30/0.26}                           & \multicolumn{1}{c|}{\textbf{0.28}/0.31/0.28}                           & \textbf{0.28/0.34/0.31}                  \\ \hline
\bottomrule\toprule
$\mathbf{r}$      & \multicolumn{5}{c}{4}                                                                                                                                                                                                                                                                                   \\ \midrule
\textbf{Lang.} & \multicolumn{1}{c|}{\textbf{LoRA}}                            & \multicolumn{1}{c|}{\textbf{AdaLoRA}}                         & \multicolumn{1}{c|}{\textbf{BT-LoRA}}                  & \multicolumn{1}{c|}{\textbf{FedL2P}}                          & \textbf{\method{}}          \\ \hline
te       & \multicolumn{1}{c|}{0.12/0.06/0.05}                  & \multicolumn{1}{c|}{0.12/0.06/0.06}                  & \multicolumn{1}{c|}{0.12/0.05/0.05}                  & \multicolumn{1}{c|}{0.13/0.06/0.06}                  & \textbf{0.14/0.07/0.07}                  \\ \hline
ar                & \multicolumn{1}{c|}{0.19/0.02/0.02}                           & \multicolumn{1}{c|}{0.19/0.02/0.02}                           & \multicolumn{1}{c|}{0.19/0.02/0.02}                           & \multicolumn{1}{c|}{0.18/0.02/0.02}                           & \textbf{0.22/0.03/0.03}                  \\ \hline
es                & \multicolumn{1}{c|}{0.33/0.36/0.33}                           & \multicolumn{1}{c|}{0.33/0.35/0.33}                           & \multicolumn{1}{c|}{0.30/0.34/0.32}                           & \multicolumn{1}{c|}{0.33/0.36/0.33}                           & \textbf{0.34/0.37/0.34}                  \\ \hline
en                & \multicolumn{1}{c|}{0.25/0.32/0.26}                           & \multicolumn{1}{c|}{0.24/0.31/0.25}                           & \multicolumn{1}{c|}{0.24/0.30/0.24}                           & \multicolumn{1}{c|}{0.25/0.31/0.25}                           & \textbf{0.28/0.35/0.29}                  \\ \hline
fr                & \multicolumn{1}{c|}{0.20/0.25/0.20}                           & \multicolumn{1}{c|}{0.19/0.23/0.19}                           & \multicolumn{1}{c|}{0.20/0.24/0.19}                           & \multicolumn{1}{c|}{0.20/0.25/0.20}                           & \textbf{0.25/0.28/0.22}                  \\ \hline
zh                & \multicolumn{1}{c|}{0.06/\textbf{0.09/0.09}} & \multicolumn{1}{c|}{0.06/\textbf{0.09/0.09}}                           & \multicolumn{1}{c|}{\textbf{0.07}/0.08/0.08} & \multicolumn{1}{c|}{0.05/0.08/0.08}                           & 0.06/0.08/0.08                  \\ \hline
pt                & \multicolumn{1}{c|}{0.27/0.33/0.30}                           & \multicolumn{1}{c|}{0.26/0.33/0.29}                           & \multicolumn{1}{c|}{0.25/0.31/0.27}                           & \multicolumn{1}{c|}{0.26/0.32/0.28}                           & \textbf{0.29/0.34/0.31}                  \\ 

\bottomrule \toprule
$\mathbf{r}$      & \multicolumn{5}{c}{8}                                                                                                                                                                                                                                                                                   \\ \midrule
\textbf{Lang.} & \multicolumn{1}{c|}{\textbf{LoRA}}                            & \multicolumn{1}{c|}{\textbf{AdaLoRA}}                         & \multicolumn{1}{c|}{\textbf{BT-LoRA}}                  & \multicolumn{1}{c|}{\textbf{FedL2P}}                          & \textbf{\method{}}          \\ \hline
te                & \multicolumn{1}{c|}{0.12/0.05/0.05}                           & \multicolumn{1}{c|}{0.12/0.06/0.06}                           & \multicolumn{1}{c|}{0.12/0.06/0.06}                           & \multicolumn{1}{c|}{\textbf{0.13}/\textbf{0.07/0.07}} & \textbf{0.13}/\textbf{0.07}/0.06 \\ \hline
ar                & \multicolumn{1}{c|}{0.21/0.02/\textbf{0.02}}                           & \multicolumn{1}{c|}{0.18/0.02/\textbf{0.02}}                           & \multicolumn{1}{c|}{0.20/0.02/\textbf{0.02}}                           & \multicolumn{1}{c|}{0.19/0.02/\textbf{0.02}}                           & \textbf{0.23/0.03/0.02}                  \\ \hline
es                & \multicolumn{1}{c|}{0.34/0.37/0.34}                           & \multicolumn{1}{c|}{0.32/0.35/0.32}                           & \multicolumn{1}{c|}{0.32/0.35/0.33}                           & \multicolumn{1}{c|}{0.33/0.35/0.33}                           & \textbf{0.35/0.38/0.35}                  \\ \hline
en                & \multicolumn{1}{c|}{0.27/0.34/0.28}                           & \multicolumn{1}{c|}{0.25/0.31/0.25}                           & \multicolumn{1}{c|}{0.24/0.31/0.25}                           & \multicolumn{1}{c|}{0.27/0.33/0.27}                           & \textbf{0.29/0.37/0.30}                  \\ \hline
fr                & \multicolumn{1}{c|}{0.22/0.26/0.20}                           & \multicolumn{1}{c|}{0.18/0.23/0.18}                           & \multicolumn{1}{c|}{0.21/0.25/0.21}                           & \multicolumn{1}{c|}{0.21/0.26/0.21}                           & \textbf{0.26/0.28/0.23}                  \\ \hline
zh                & \multicolumn{1}{c|}{\textbf{0.06}/\textbf{0.09}/\textbf{0.09}} & \multicolumn{1}{c|}{\textbf{0.06}/0.08/0.08} & \multicolumn{1}{c|}{\textbf{0.06}/0.08/0.08} & \multicolumn{1}{c|}{0.05/0.08/0.08}                           & 0.05/0.08/0.08                           \\ \hline
pt                & \multicolumn{1}{c|}{0.28/0.34/\textbf{0.31}}                           & \multicolumn{1}{c|}{0.26/0.32/0.29}                           & \multicolumn{1}{c|}{0.26/0.32/0.29}                           & \multicolumn{1}{c|}{0.28/0.33/0.29}                           & \textbf{0.29/0.35/0.31}                  \\ 

\bottomrule \toprule
$\mathbf{r}$      & \multicolumn{5}{c}{16}                                                                                                                                                                                                                                                                                  \\ \midrule
\textbf{Lang.} & \multicolumn{1}{c|}{\textbf{LoRA}}                            & \multicolumn{1}{c|}{\textbf{AdaLoRA}}                         & \multicolumn{1}{c|}{\textbf{BT-LoRA}}                  & \multicolumn{1}{c|}{\textbf{FedL2P}}                          & \textbf{\method{}}          \\ \hline
te                & \multicolumn{1}{c|}{0.12/0.06/0.05}                           & \multicolumn{1}{c|}{0.13/0.06/0.06}                           & \multicolumn{1}{c|}{0.14/0.07/0.07}                           & \multicolumn{1}{c|}{\textbf{0.16/0.08/0.08}}                  & 0.13/0.07/0.06                           \\ \hline
ar                & \multicolumn{1}{c|}{0.21/0.02/0.02}                           & \multicolumn{1}{c|}{0.17/0.02/0.02}                           & \multicolumn{1}{c|}{0.21/0.02/0.02}                           & \multicolumn{1}{c|}{0.20/0.02/0.02}                           & \textbf{0.24/0.03/0.03}                  \\ \hline
es                & \multicolumn{1}{c|}{\textbf{0.34}/0.36/0.34}                           & \multicolumn{1}{c|}{0.30/0.33/0.30}                           & \multicolumn{1}{c|}{\textbf{0.34}/0.36/0.34}                           & \multicolumn{1}{c|}{0.33/0.35/0.33}                           & \textbf{0.34/0.38/0.35}                  \\ \hline
en                & \multicolumn{1}{c|}{\textbf{0.28}/0.34/0.28}                           & \multicolumn{1}{c|}{0.24/0.30/0.24}                           & \multicolumn{1}{c|}{0.26/0.32/0.26}                           & \multicolumn{1}{c|}{\textbf{0.28}/0.33/0.27}                           & \textbf{0.28/0.35/0.29}                  \\ \hline
fr                & \multicolumn{1}{c|}{\textbf{0.25/0.29/0.23}}                  & \multicolumn{1}{c|}{0.18/0.23/0.19}                           & \multicolumn{1}{c|}{0.22/0.26/0.21}                           & \multicolumn{1}{c|}{0.23/0.23/0.19}                           & \textbf{0.25}/0.28/0.22                           \\ \hline
zh                & \multicolumn{1}{c|}{\textbf{0.06}/\textbf{0.08/0.08}} & \multicolumn{1}{c|}{\textbf{0.06}/\textbf{0.08/0.08}} & \multicolumn{1}{c|}{0.05/\textbf{0.08/0.08}}                           & \multicolumn{1}{c|}{0.05/\textbf{0.08/0.08}}                           & 0.05/\textbf{0.08/0.08 }                          \\ \hline
pt                & \multicolumn{1}{c|}{0.29/0.34/0.31}                           & \multicolumn{1}{c|}{0.25/0.31/0.28}                           & \multicolumn{1}{c|}{0.29/0.34/0.31}                           & \multicolumn{1}{c|}{0.28/0.32/0.29}                           & \textbf{0.30/0.35/0.32}                  \\ 
\bottomrule
\end{tabular}


\caption{Average METEOR/ROUGE-1/ROUGE-L of each language for \seen{} clients of our Fed-Aya setup. The pretrained MobileLLaMA-1.4B model is trained using \textbf{Standard FL with LoRA} following FedLLM-Bench~\cite{fedllm-bench} and the resulting \basemodel{} is personalized to each client given a baseline approach.}
\label{tab:mobilellama_fedaya_seen}
\end{table*}

\begin{table*}[t]
\centering

\begin{tabular}{c|ccccc}
\toprule
$\mathbf{r}$      & \multicolumn{5}{c}{2}                                                                                                                                                                                                                                                                                   \\ \midrule
\textbf{Lang.} & \multicolumn{1}{c|}{\textbf{LoRA}}                            & \multicolumn{1}{c|}{\textbf{AdaLoRA}}                         & \multicolumn{1}{c|}{\textbf{BT-LoRA}}                  & \multicolumn{1}{c|}{\textbf{FedL2P}}                          & \textbf{\method{}}          \\ \hline
te                & \multicolumn{1}{c|}{0.05/0.00/0.00}                           & \multicolumn{1}{c|}{0.03/0.00/0.00}                           & \multicolumn{1}{c|}{0.03/0.00/0.00}                           & \multicolumn{1}{c|}{\textbf{0.12}/0.00/0.00} & 0.11/\textbf{0.01/0.01} \\ \hline
ar                & \multicolumn{1}{c|}{0.10/\textbf{0.05/0.05}} & \multicolumn{1}{c|}{0.10/\textbf{0.05/0.05}}                           & \multicolumn{1}{c|}{0.09/\textbf{0.05/0.05}}                           & \multicolumn{1}{c|}{\textbf{0.14}/0.02/0.02} & 0.11/\textbf{0.05/0.05}                           \\ \hline
es                & \multicolumn{1}{c|}{0.35/0.45/0.40}                           & \multicolumn{1}{c|}{0.32/0.42/0.38}                           & \multicolumn{1}{c|}{0.36/0.41/0.37}                           & \multicolumn{1}{c|}{\textbf{0.39/0.46}/\textbf{0.42}} & 0.38/0.45/\textbf{0.42} \\ \hline
en                & \multicolumn{1}{c|}{0.24/0.27/0.24}                           & \multicolumn{1}{c|}{0.22/0.26/0.22}                           & \multicolumn{1}{c|}{0.18/0.20/0.17}                           & \multicolumn{1}{c|}{0.27/0.30/0.27}                           & \textbf{0.30/0.34/0.30}                  \\ \hline
fr                & \multicolumn{1}{c|}{\textbf{0.05}/0.00/0.00}                           & \multicolumn{1}{c|}{\textbf{0.05}/\textbf{0.05/0.05}} & \multicolumn{1}{c|}{0.04/0.00/0.00}                           & \multicolumn{1}{c|}{0.00/0.00/0.00}                           & \textbf{0.05}/0.00/0.00                           \\ \hline
zh                & \multicolumn{1}{c|}{0.14/\textbf{0.01/0.01}}                           & \multicolumn{1}{c|}{\textbf{0.16}/\textbf{0.01/0.01}} & \multicolumn{1}{c|}{\textbf{0.16}/\textbf{0.01/0.01}}                           & \multicolumn{1}{c|}{0.09/\textbf{0.01/0.01}} & 0.07/\textbf{0.01/0.01 }                          \\ \hline
pt                & \multicolumn{1}{c|}{0.19/0.22/0.21}                           & \multicolumn{1}{c|}{0.19/0.23/0.21}                           & \multicolumn{1}{c|}{0.15/0.16/0.15}                           & \multicolumn{1}{c|}{\textbf{0.20}/0.22/0.21} & \textbf{0.20}/\textbf{0.25/0.24} \\ \hline
ru                & \multicolumn{1}{c|}{0.09/\textbf{0.06/0.06}}                           & \multicolumn{1}{c|}{\textbf{0.10}/\textbf{0.06/0.06}} & \multicolumn{1}{c|}{0.09/\textbf{0.06/0.06}}                           & \multicolumn{1}{c|}{0.09/\textbf{0.06/0.06}}                           & \textbf{0.10}/\textbf{0.06/0.06}                           \\ 

\bottomrule \toprule
$\mathbf{r}$      & \multicolumn{5}{c}{4}                                                                                                                                                                                                                                                                                   \\ \midrule
\textbf{Lang.} & \multicolumn{1}{c|}{\textbf{LoRA}}                            & \multicolumn{1}{c|}{\textbf{AdaLoRA}}                         & \multicolumn{1}{c|}{\textbf{BT-LoRA}}                  & \multicolumn{1}{c|}{\textbf{FedL2P}}                          & \textbf{\method{}}          \\ \hline
te                & \multicolumn{1}{c|}{0.07/0.01/0.01}                           & \multicolumn{1}{c|}{0.03/0.00/0.00}                           & \multicolumn{1}{c|}{0.05/0.01/0.01}                           & \multicolumn{1}{c|}{0.06/0.01/0.01}                           & \textbf{0.11/0.02/0.02}                  \\ \hline
ar                & \multicolumn{1}{c|}{0.10/\textbf{0.06/0.06}}                           & \multicolumn{1}{c|}{0.09/\textbf{0.06}/\textbf{0.06}} & \multicolumn{1}{c|}{0.09/0.05/0.05}                           & \multicolumn{1}{c|}{0.10/\textbf{0.06}/\textbf{0.06}} & \textbf{0.15}/0.03/0.03 \\ \hline
es                & \multicolumn{1}{c|}{0.32/0.41/0.37}                           & \multicolumn{1}{c|}{0.33/0.43/0.39}                           & \multicolumn{1}{c|}{0.34/0.41/0.36}                           & \multicolumn{1}{c|}{0.31/0.41/0.38}                           & \textbf{0.37/0.46/0.43}                  \\ \hline
en                & \multicolumn{1}{c|}{0.23/0.27/0.24}                           & \multicolumn{1}{c|}{0.18/0.21/0.18}                           & \multicolumn{1}{c|}{0.17/0.20/0.16}                           & \multicolumn{1}{c|}{0.25/0.29/0.26}                           & \textbf{0.29/0.33/0.28}                  \\ \hline
fr                & \multicolumn{1}{c|}{0.05/0.05/0.05}                           & \multicolumn{1}{c|}{0.05/0.00/0.00}                           & \multicolumn{1}{c|}{\textbf{0.12}/0.06/0.06}                           & \multicolumn{1}{c|}{\textbf{0.12/0.07/0.07}}                  & 0.05/0.00/0.00                           \\ \hline
zh                & \multicolumn{1}{c|}{0.15/\textbf{0.01/0.01}}                           & \multicolumn{1}{c|}{\textbf{0.16}/\textbf{0.01}/0.00} & \multicolumn{1}{c|}{\textbf{0.16}/\textbf{0.01/0.01}} & \multicolumn{1}{c|}{\textbf{0.16}/0.00/0.00}                           & 0.07/\textbf{0.01/0.01}                           \\ \hline
pt                & \multicolumn{1}{c|}{\textbf{0.19}/0.23/0.21}                           & \multicolumn{1}{c|}{0.18/0.21/0.20}                           & \multicolumn{1}{c|}{0.15/0.16/0.15}                           & \multicolumn{1}{c|}{0.17/0.19/0.18}                           & \textbf{0.19/0.25/0.24}                  \\ \hline
ru                & \multicolumn{1}{c|}{0.10/\textbf{0.06/0.06}}                           & \multicolumn{1}{c|}{0.12/\textbf{0.06/0.06}}                           & \multicolumn{1}{c|}{0.09/0.02/0.02}                           & \multicolumn{1}{c|}{0.08/\textbf{0.06/0.06}}                           & \textbf{0.13/0.06/0.06}                  \\ 
\bottomrule \toprule
$\mathbf{r}$      & \multicolumn{5}{c}{8}                                                                                                                                                                                                                                                                                   \\ \midrule
\textbf{Lang.} & \multicolumn{1}{c|}{\textbf{LoRA}}                            & \multicolumn{1}{c|}{\textbf{AdaLoRA}}                         & \multicolumn{1}{c|}{\textbf{BT-LoRA}}                  & \multicolumn{1}{c|}{\textbf{FedL2P}}                          & \textbf{\method{}}          \\ \hline
te                & \multicolumn{1}{c|}{0.10/0.00/0.00}                           & \multicolumn{1}{c|}{0.02/0.00/0.00}                           & \multicolumn{1}{c|}{0.05/0.00/0.00}                           & \multicolumn{1}{c|}{0.09/0.00/0.00}                           & \textbf{0.11/0.01/0.01}                  \\ \hline
ar                & \multicolumn{1}{c|}{0.12/0.05/0.05}                           & \multicolumn{1}{c|}{0.09/0.04/0.04}                           & \multicolumn{1}{c|}{0.10/\textbf{0.06/0.06}} & \multicolumn{1}{c|}{0.11/0.05/0.05}                           & \textbf{0.16}/0.01/0.01 \\ \hline
es                & \multicolumn{1}{c|}{0.34/0.44/0.40}                           & \multicolumn{1}{c|}{0.34/0.43/0.39}                           & \multicolumn{1}{c|}{0.35/0.43/0.38}                           & \multicolumn{1}{c|}{0.32/0.44/0.41}                           & \textbf{0.45/0.52/0.49}                  \\ \hline
en                & \multicolumn{1}{c|}{0.27/\textbf{0.31/0.27}} & \multicolumn{1}{c|}{0.18/0.21/0.18}                           & \multicolumn{1}{c|}{0.19/0.22/0.19}                           & \multicolumn{1}{c|}{0.24/0.28/0.25}                           & \textbf{0.28}/0.29/0.25 \\ \hline
fr                & \multicolumn{1}{c|}{0.05/0.00/0.00}                           & \multicolumn{1}{c|}{0.12/\textbf{0.08/0.08}} & \multicolumn{1}{c|}{\textbf{0.13}/\textbf{0.08/0.08}} & \multicolumn{1}{c|}{0.05/0.00/0.00}                           & 0.05/0.00/0.00                           \\ \hline
zh                & \multicolumn{1}{c|}{0.12/\textbf{0.01/0.01}}                           & \multicolumn{1}{c|}{\textbf{0.15/0.01}/0.00}                           & \multicolumn{1}{c|}{\textbf{0.15}/\textbf{0.01/0.01}} & \multicolumn{1}{c|}{0.14/\textbf{0.01/0.01}} & 0.11/\textbf{0.01/0.01}                           \\ \hline
pt                & \multicolumn{1}{c|}{0.19/0.23/0.22}                           & \multicolumn{1}{c|}{0.18/0.19/0.18}                           & \multicolumn{1}{c|}{0.14/0.16/0.15}                           & \multicolumn{1}{c|}{0.19/0.23/0.21}                           & \textbf{0.22/0.27/0.26}                  \\ \hline
ru                & \multicolumn{1}{c|}{0.08/\textbf{0.06/0.06}}                           & \multicolumn{1}{c|}{0.11/\textbf{0.06/0.06}}                           & \multicolumn{1}{c|}{0.09/\textbf{0.06/0.06}}                           & \multicolumn{1}{c|}{\textbf{0.12}/\textbf{0.06/0.06}} & 0.09/0.05/0.05                           \\ 
\bottomrule \toprule
$\mathbf{r}$      & \multicolumn{5}{c}{16}                                                                                                                                                                                                                                                                                  \\ \midrule
\textbf{Lang.} & \multicolumn{1}{c|}{\textbf{LoRA}}                            & \multicolumn{1}{c|}{\textbf{AdaLoRA}}                         & \multicolumn{1}{c|}{\textbf{BT-LoRA}}                  & \multicolumn{1}{c|}{\textbf{FedL2P}}                          & \textbf{\method{}}          \\ \hline
te                & \multicolumn{1}{c|}{0.12/0.00/0.00}                           & \multicolumn{1}{c|}{0.01/0.00/0.00}                           & \multicolumn{1}{c|}{0.07/0.01/0.01}                           & \multicolumn{1}{c|}{0.11/\textbf{0.07/0.07}} & \textbf{0.13}/0.02/0.01 \\ \hline
ar                & \multicolumn{1}{c|}{\textbf{0.14/0.06/0.06}}                  & \multicolumn{1}{c|}{0.08/0.04/0.04}                           & \multicolumn{1}{c|}{0.09/0.04/0.04}                           & \multicolumn{1}{c|}{0.12/0.03/0.03}                           & 0.12/0.02/0.02                           \\ \hline
es                & \multicolumn{1}{c|}{0.36/0.42/0.39}                           & \multicolumn{1}{c|}{0.35/0.42/0.39}                           & \multicolumn{1}{c|}{0.35/0.42/0.38}                           & \multicolumn{1}{c|}{0.39/0.45/0.41}                           & \textbf{0.40/0.49/0.45}                  \\ \hline
en                & \multicolumn{1}{c|}{0.29/\textbf{0.32/0.28}} & \multicolumn{1}{c|}{0.18/0.22/0.17}                           & \multicolumn{1}{c|}{0.21/0.24/0.21}                           & \multicolumn{1}{c|}{0.26/0.26/0.21}                           & \textbf{0.30}/0.31/0.27 \\ \hline
fr                & \multicolumn{1}{c|}{0.05/0.00/0.00}                           & \multicolumn{1}{c|}{0.14/0.13/0.13}                           & \multicolumn{1}{c|}{0.17/\textbf{0.19/0.19}} & \multicolumn{1}{c|}{0.13/0.10/0.10}                           & \textbf{0.29}/0.17/0.17 \\ \hline
zh                & \multicolumn{1}{c|}{0.10/\textbf{0.01/0.01}}                           & \multicolumn{1}{c|}{0.14/\textbf{0.01/0.01}} & \multicolumn{1}{c|}{\textbf{0.16}/\textbf{0.01/0.01}} & \multicolumn{1}{c|}{0.13/0.00/0.00}                           & 0.10/\textbf{0.01/0.01 }                          \\ \hline
pt                & \multicolumn{1}{c|}{0.19/0.24/0.23}                           & \multicolumn{1}{c|}{0.18/0.19/0.18}                           & \multicolumn{1}{c|}{0.19/0.24/0.22}                           & \multicolumn{1}{c|}{0.20/0.21/0.20}                           & \textbf{0.24/0.29/0.27}                  \\ \hline
ru                & \multicolumn{1}{c|}{\textbf{0.14/0.06/0.06}}                  & \multicolumn{1}{c|}{0.10/\textbf{0.06/0.06} }                          & \multicolumn{1}{c|}{0.10/\textbf{0.06/0.06}}                        & \multicolumn{1}{c|}{0.06/0.03/0.03}                           & 0.08/0.02/0.02                           \\ 
\bottomrule
\end{tabular}

\caption{Average METEOR/ROUGE-1/ROUGE-L of each language for \unseen{} clients of our Fed-Aya setup. The pretrained MobileLLaMA-1.4B model is trained using \textbf{Standard FL with LoRA} following FedLLM-Bench~\cite{fedllm-bench} and the resulting \basemodel{} is personalized to each client given a baseline approach.}
\label{tab:mobilellama_fedaya_unseen}
\end{table*}

\end{document}